\documentclass[journal]{IEEEtran}
\usepackage{soul}
\usepackage{paralist}

\usepackage{graphicx}
\usepackage{amsmath,amssymb,amsfonts}
\usepackage[ruled,linesnumbered]{algorithm2e}
\usepackage{algpseudocode}
\usepackage{amsmath}
\usepackage{subfigure}
\usepackage{booktabs}
\usepackage{tabu}
\usepackage{multirow}
\usepackage{color}
\usepackage[abs]{overpic}

\def\ie{\emph{i.e.,}}
\def\etc{\emph{etc.}}

\usepackage{comment}
\usepackage[pagebackref=true,breaklinks=true,colorlinks,citecolor=blue,linkcolor=blue,bookmarks=false,urlcolor=blue]{hyperref}

\begin{document}
%
\title{Self-Supervised Tracking via Target-Aware\\ Data Synthesis} 
%
%
%

\author{Xin~Li,
        Wenjie~Pei,
        Yaowei Wang,~\IEEEmembership{Member, IEEE},
        Zhenyu~He,~\IEEEmembership{Senior Member, IEEE},\\
        Huchuan~Lu,~\IEEEmembership{Senior Member, IEEE},
        and
        Ming-Hsuan~Yang,~\IEEEmembership{Fellow, ~IEEE}
\thanks{
This work was supported in part by the Key Research Project of Peng Cheng Laboratory (PCL2021A07), in part by the NSFC (62002241, U20B2052, U2013210, 62006060, and 62172126), and in part by Shenzhen Science and Technology Innovation Committee ( JCYJ20210324120202006 and JCYJ20220818102415032). }

\thanks{Xin~Li, Yaowei~Wang, and Huchuan~Lu are with Peng Cheng Laboratory, Shenzhen, China (e-mail: xinlihitsz@gmail.com; wangyw@pcl.ac.cn; lhchuan@dlut.edu.cn). Huchuan~Lu is also with the School of Information
and Communication Engineering, Dalian University of Technology,
Dalian, China.}
\thanks{Weijie~Pei and Zhenyu~He are with the School of Computer Science and Technology, Harbin Institute of Technology, Shenzhen, China (e-mail: wenjiecoder@outlook.com; zhenyuhe@hit.edu.cn). }
\thanks{Ming-Hsuan Yang is with the Department of Electrical Engineering and Computer Science, University of California at Merced, Merced, CA 95344 USA (e-mail: mhyang@ucmerced.edu).}
\thanks{(Xin Li and Wenjie Pei contributed equally to this work. Corresponding authors: Yaowei Wang and Huchuan Lu.)}

}

\maketitle

\begin{abstract}
While deep-learning based tracking methods have achieved substantial progress, they entail large-scale and high-quality annotated data for sufficient training.
To eliminate expensive and exhaustive annotation, we study self-supervised learning for visual tracking.
In this work, we develop the Crop-Transform-Paste operation, which is able to synthesize sufficient training data by simulating various appearance variations during tracking, including appearance variations of objects and background interference.
Since the target state is known in all synthesized data, existing deep trackers can be trained in routine ways using the synthesized data without human annotation.
%
The proposed target-aware data-synthesis method adapts existing tracking approaches within a self-supervised learning framework without algorithmic changes.
%
Thus, the proposed self-supervised learning mechanism can be seamlessly integrated into existing tracking frameworks to perform training.
%
Extensive experiments show that our method 1) achieves favorable performance against
supervised learning schemes under the cases with limited annotations; 2) helps deal with various tracking challenges such as object deformation, occlusion, or background clutter due to its manipulability; 3) performs favorably against state-of-the-art unsupervised tracking methods; 4) boosts the performance of various state-of-the-art supervised learning frameworks, including SiamRPN++, DiMP, and TransT.
\end{abstract}


\IEEEpeerreviewmaketitle

\section{Introduction}
\IEEEPARstart{G}{iven} a specified target object in the initial frame, visual tracking aims to estimate the target state (typically indicated by a bounding box) in the subsequent frames of a tracking sequence. 
It is a fundamental task in computer vision and has a wide range of applications, such as autonomous driving, human-computer interaction, and intelligent video analysis.
Recently, deep-learning-based trackers~\cite{SIAMESEFC,LSTMSiameseT, LARACF, CCRRDT} have made considerable progress in visual tracking due to the powerful capability of image feature learning.
Nevertheless, such methods demand large-scale and high-quality annotated data for sufficient training, which is an inherent limitation of deep convolutional networks applied in visual tracking. 
Thus these approaches do not perform well under challenging tracking scenarios when large-scale annotated data is not available.

\begin{figure}
    \centering
      \includegraphics[width=0.8\linewidth]{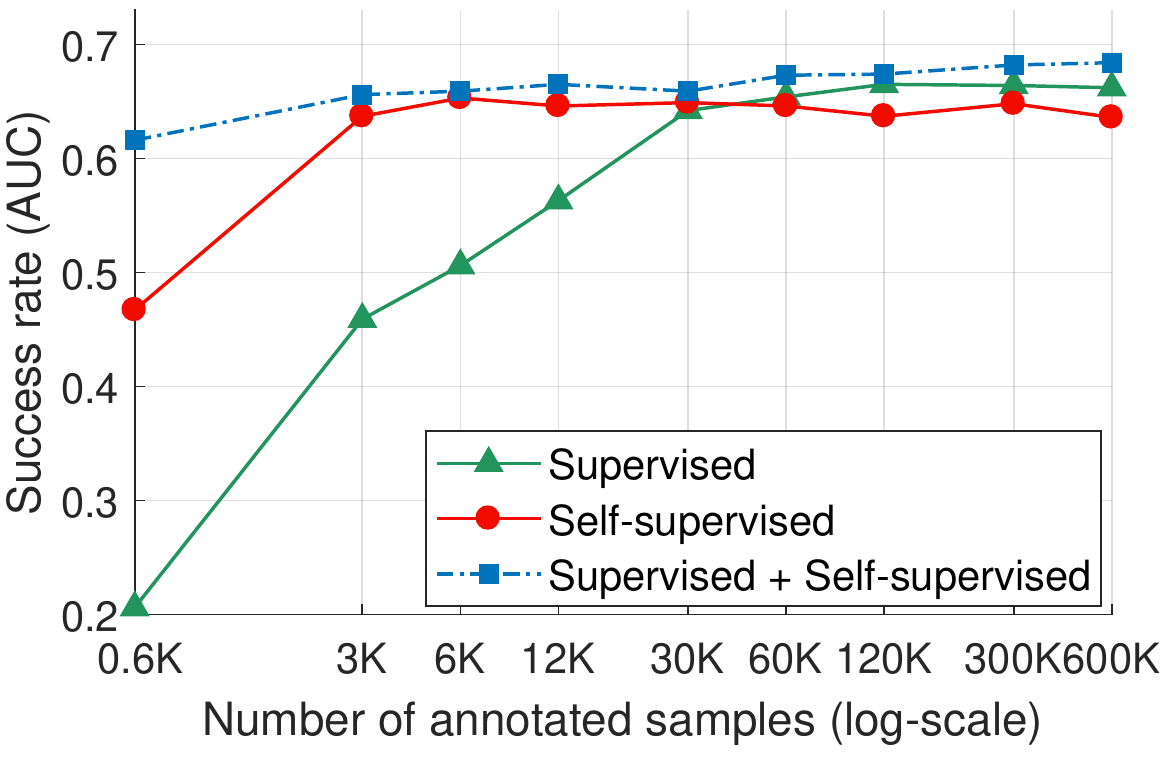}
    \caption{\textbf{Performance of SiamRPN++~\cite{SiamRPN++} on the OTB100 dataset trained with different learning schemes as a function of the annotated training data size.} Full annotations of training data are used in the supervised learning scheme while only one annotated frame per video (indicating the target object)  is used in the self-supervised tracking scheme. The proposed method based on supervised and self-supervised learning exploits both the annotated and the synthesized samples for training.}
    \label{fig:sl+su}
    \vspace{-2mm}
\end{figure}

In this paper, we explore the potential of self-supervised learning for visual tracking to alleviate the above-mentioned issues.
Instead of using manually annotated data, we synthesize training data by simulating various appearance variations of a target object.
Since the object state is known in all the synthesized data, we are able to train trackers in routine ways without human annotation. 
Such a learning way can be viewed as a form of self-supervised learning, which seeks to perform training without labeled data in a self-supervised manner.

Self-supervised learning has been extensively studied for visual representation learning via predefined pretext tasks~\cite{SSSDO,sslpir,SSDE}, which focus on visual representation by training an individual learning module following a contrastive learning framework~\cite{SimCLR,moco}.
However, it is challenging to effectively exploit self-supervised learning for visual tracking, since the  tracking task requires precise localization under scenarios with various target appearance changes and distractions from the background.
Thus, the proposed self-supervised learning mechanism should be designed specifically for visual tracking and can be integrated into existing tracking frameworks without model structure adaptation.

One essential task in visual tracking is to learn an effective target object representation that is robust to appearance variations as well as background interference.
Our core idea is to perform image transformation operations on the given target object for simulating potential target variations in tracking scenarios.
Specifically, we develop a simple Crop-Transform-Paste operation to synthesize data for performing self-supervised learning in tracking tasks.
It first crops the given target object patch from the initial frame, then performs image transformations on the cropped patch, and finally pastes the transformed patch onto a new frame with probable background interference.
The proposed Crop-Transform-Paste operation facilitates synthesizing sufficient data with diverse transformations as needed in tracking scenarios, which are readily used to train existing trackers.

Compared to existing supervised learning tracking schemes, the proposed self-supervised learning mechanism based on the Crop-Transform-Paste operation benefits from multiple advantages.
First, our method eliminates the need of expensive and exhaustive human annotation.
Second, our Crop-Transform-Paste operation accommodates different transformations to address the challenges in visual tracking, such as {\em cutout} for occlusion, {\em shear} for deformation, and {\em similar-patch paste} for background clutter.
In contrast, the samples involving tracking challenges from human-annotated data are always too scarce to train trackers sufficiently.
Third, we can synthesize a large amount of data from a limited size of initial (raw) tracking sequence data, which implies that our method requires much less raw data than the supervised learning way for sufficient training.
Finally, the self-supervised learning mechanism can be readily integrated with existing supervised learning frameworks to further boost tracking performance.
This is particularly useful in few-shot tracking scenarios where only a few annotated samples are provided.
Figure~\ref{fig:sl+su} shows the tracking performance of a tracker using the supervised learning manner, our self-supervised learning scheme, and the integrated way, respectively.
The proposed mechanism compares favorably against the supervised learning scheme when only a small size of training data is available and even achieves close performance when sufficient annotated data is provided.
In addition, integrating self-supervised learning into supervised learning boosts the performance distinctly.

We conduct two sets of experiments to evaluate the proposed method extensively. 
First, we perform ablation study to investigate the design of our method, including the effect of each individual component of the proposed Crop-Transform-Paste operation, the capability to handle various typical tracking challenges, and the effectiveness in improving data efficiency. 
Second, we evaluate our approach by comparing with the state-of-the-art methods under both unsupervised and supervised cases.
We demonstrate that our method are effective in both cases and boost the performance of various existing tracking frameworks, including SiamRPN++~\cite{SiamRPN++}, DiMP~\cite{DiMP}, and TransT~\cite{TransT} (based on Transformer).

%

\section{Related Work}
\vspace{1mm}
\noindent\textbf{Supervised tracking methods.}
More and more tracking methods~\cite{SiamRPN, DaSiamRPN,TADT,STMT, MTTIT} learn to track through the offline training on large-scale data with annotations.
The training data drawn from real videos are organized as target exemplar and test image pairs with bounding box annotations (such as the training data of TrackingNet~\cite{TrackingNet}, and GOT-10k~\cite{GOT10k}).
It has been verified by numerous exiting trackers~\cite{SiamRPN, DaSiamRPN, SiamRPN++} that simply increasing the scale of training datasets can improve the performance to some extent.
However, developing large-scale training data is expensive and the selection of videos is not easy since training data should contain sufficient various kinds of target variations for efficient training.
The challenges of expensive manual annotating and low data quality (in terms of sufficient variations and balanced distribution of the variations) are becoming the bottleneck of supervised tracking methods.

\vspace{1mm}
\noindent\textbf{Unsupervised tracking methods.}
A few approaches~\cite{UDT,selfSDCT, USOT} explore unsupervised learning schemes to train tracking models.
They exploit the temporal forward and backward trajectory-consistency to perform offline training in an unsupervised manner without using target box annotations.
Despite the promising results achieved by these methods, they do not work well when the temporal cycle-consistency does not hold in the cases where the target object is occluded or changes dramatically.
Instead of using the temporal consistency, we propose a target-aware data synthesis scheme to perform training on unlabeled data, enabling a self-supervised tracking framework.
Our approach develops a Crop-Transform-Paste mechanism to synthesize training data by simulating both the target appearance and background interference during tracking.
The proposed algorithm achieves close performance to that of the supervised training scheme, improves data-efficiency, and boosts the performance of supervised trackers.

\begin{figure*}[t]
    \centering
        \includegraphics[width=0.95\textwidth]{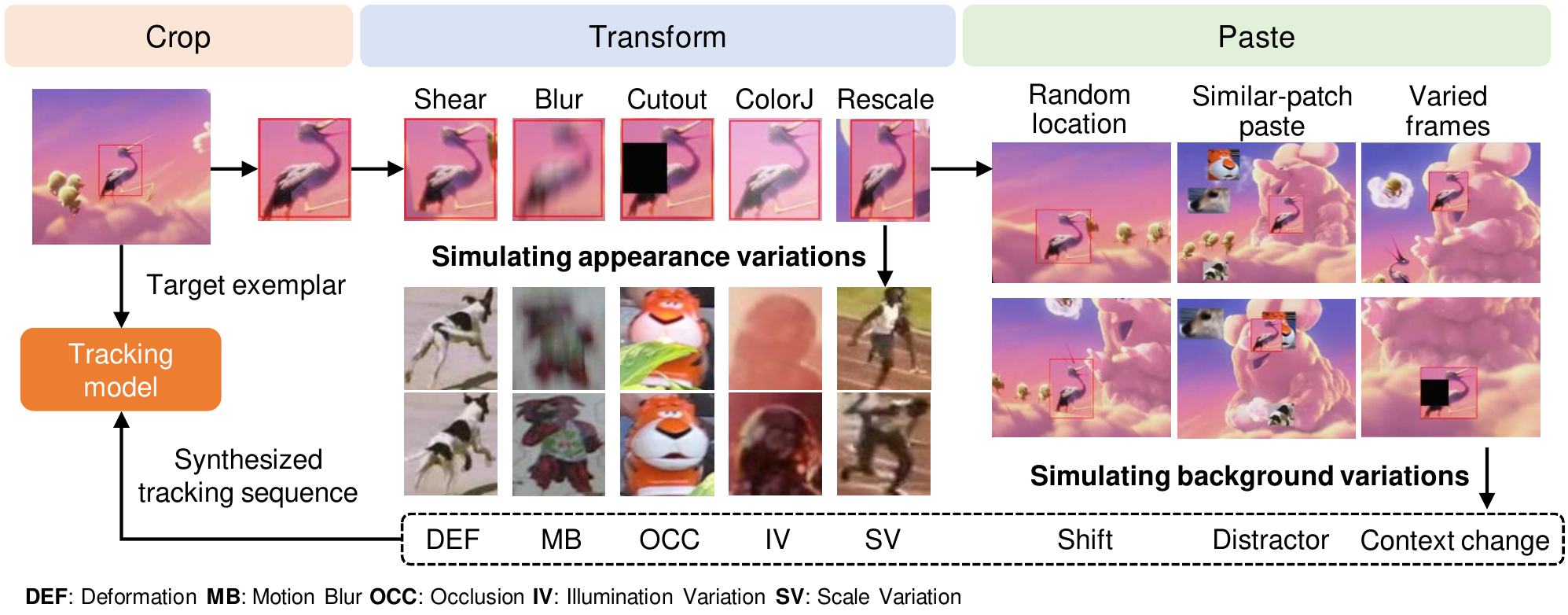}
    \caption{\textbf{Framework of the target-aware self-supervised learning mechanism.}
   The Crop-Transform-Paste operation is specifically developed to synthesize target-aware training data, which allows training existing trackers in a routine way without human annotations. It first crops a target patch from the initial frame, then performs the tracking-related transformations on the target patch to simulate diverse appearance variations of the target during tracking, and finally pastes the transformed target patches onto random frames with different locations to simulate the background interference around the target.
}\vspace{-3mm}
    \label{fig:framework}
\end{figure*}

\vspace{1mm}
\noindent\textbf{Self-supervised learning schemes.}
Numerous self-supervised learning methods~\cite{SimCLR,moco} have been developed to exploit large-scale data for effective representation schemes.
They are usually carried out by maximizing the similarity between two random augmentations of an image to learn invariance encodings across different data augmentation views.
The recent self-supervised learning methods~\cite{SimCLR,moco,BYOL} mainly focus on
making the training more efficient and effective, such as the methods using simpler frameworks~\cite{SimCLR}, less negative training samples~\cite{BYOL}, and smaller batch sizes~\cite{moco}.
In contrast, our approach focuses on how to eliminate expensive and exhaustive annotation for training tracking models.
We develop a Crop-Transform-Paste mechanism to synthesize training samples that simulate target appearance and background interference during tracking with several customized transformations, which enables self-supervised tracking. 

\vspace{1mm}
\noindent\textbf{Data augmentations.}
Data augmentation methods are widely used for training deep CNNs models to increase training sample diversity, which facilitates training efficiency and effectiveness.
General data augmentation methods including random crop~\cite{RandomCrop}, color jittering~\cite{ColorJ}, and blur have played an important role in achieving state-of-the-art performance on multiple visual tasks, such as classification~\cite{RandomCrop}, object detection~\cite{yolov1}, and segmentation~\cite{CopyPasteNew}.
In addition, some image mixing augmentations that mix the content of different images are proposed, such as, MixUp~\cite{mixup}, CutMix~\cite{CutMix}, and Copy-Paste~\cite{CopyPaste,CopyPasteNew}.
By combining the content from multiple images with appropriate adaptions, the mixing augmentations produce new data distributions and feature patterns.
Compared with the existing augmentation methods, the proposed Crop-Transform-Paste mechanism differs in the following aspects.
First, we use transformations to simulate tracking-specific target appearance and background interference for synthesizing the training data, whilst existing methods usually aim to learn robust features by introducing diverse transformations.
%
Second, we compose the tracking-related data augmentations with the copy-paste operation to generate image pairs that are able to simulate appearance variations during tracking for data generation, while general augmentation techniques focus on image transformations for learning deep representations. Thus different motivations lead to different appropriate techniques.
%


\section{Target-Aware Self-Supervision}
\label{sec:algorithm}

Our core idea is to synthesize sufficient target-aware training data that simulate real-world tracking scenarios for training tracking models without human annotations. 
Specifically, we develop the Crop-Transform-Paste operation to generate pseudo frames based on the given target exemplar with tracking-related data transformations. As all target states in the pseudo test frames are known, we can train trackers without human annotations in every test frame, enabling a form of self-supervised learning for visual tracking.
In the following, we first present the overall framework of our method and then elaborate on the Crop-Transform-Paste mechanism.

\subsection{Overall Framework}
Figure~\ref{fig:framework} shows the overall framework of the proposed target-aware self-supervised tracking mechanism. 
Given a tracking sequence with a specified target object, the proposed Crop-Transform-Paste operation first crops a target patch from the initial image, then performs the tracking-related transformations on the target patch to simulate diverse appearance variations of the target object during tracking, and finally pastes the transformed target patch onto different frames with different locations to simulate the background interference around the target.
It is designed specifically for tracking and focuses on data synthesis to construct sufficient target-aware training data that are consistent with the real-world tracking scenarios and can be readily used to train existing trackers without human annotations. 
Such a self-supervised learning mechanism can be seamlessly integrated into existing tracking frameworks without model structure adaptation.

\subsection{Crop-Transform-Paste}

Given a frame sequence $\mathcal{X}=\{x_i\}$ where $i$ is the frame number, and its reference frame $x_r$ with a bounding box $b_r$ specifying the target object to be tracked, we first get the target patch $z_t$ by cropping.
With the specified target patch, we generate a large number of target image patches $\mathcal{Z}=\{z_j\}$ with different states involving various appearance changes by performing a randomly selected subset $\Omega _j$ of the pre-defined tracking-related data transformations $\Omega$, which is formulated as:
\begin{equation}
    z_j,b_j = \Omega_j (x_r,b_r).
    \label{eq:transform}
\end{equation}

We then paste the generated target patches $\mathcal{Z}=\{z_j\}$ onto randomly selected image frames $\mathcal{X}=\{x_i\}$ from the same sequence at a randomly location and get every synthesized sample and its target state by
\begin{equation}
    x^{'}_j , b^{'}_j = \Psi_{j} (z_j,b_j,x_i),
    \label{eq:crop-transform-paste}
\end{equation}
where $\Psi(.)$ denotes the paste operation.
By performing the crop-transform-paste process with various (customized) sets of transformations and various degrees, we are able to generate massive training samples with sufficient variations only base on one target patch.
Below we present the details of each step.

\vspace{1mm}

\noindent\textbf{Crop: generation of the base target patch.}
The base target patch can be generated using different ways:
\begin{itemize}
    \item 
    Using given bounding boxes to crop target patches.
    Such option is consistent with general setting in the tracking task, which enables us to have a direct comparison with the routine supervised tracking methods.
    \item 
    Cropping random object patches detected by a pre-trained object detector. 
    We perform this option to eliminate the potential bias distribution of tracking objects annotated in the training dataset, and study the effect of tracking object selection.
    In addition, such option provides us a way that does not rely on the annotation of the tracking object in the initial frame.
    \item 
    Cropping a random patch from an image as the base target, which does not require annotation or object detectors. 
    Nevertheless, a random patch may be less effective in providing semantic features and meaningful object boundary information for effective training.
\end{itemize}
The effect of these ways for generating the base target patch is analyzed in Section~\ref{sec:study-CTP}.

\vspace{1mm}
\noindent\textbf{Transform: simulation of target appearance variations.}
A key property of our method is its manipulability, namely we can generate the synthesized data that simulate various kinds of challenging appearance variations by applying specifically designed transformations on the cropped target patch.
Figure~\ref{fig:framework} shows the used transformations for simulating corresponding appearance variations in tracking.
We adopt the transformations of \emph{Blur}, \emph{Cutout}, \emph{Rescale}, \emph{Color Jittering}, and \emph{Shear} to simulate target variations of motion blur, occlusion, scale change, illumination variation, and deformation in tracking scenarios. 
Since the blurs that occurred in tracking scenarios are usually caused by camera shaking or fast motion, we use a shaking blur transformation generated by a filter kernel with non-zero values only in the horizontal and vertical centerlines, instead of the Gaussian blur. 
For the cutout transformation, we fill the cut area of the target with 0s to simulate occlusion.
For all these transformations, we use wide ratio and degree ranges to generate sufficient training samples that cover target variations during tracking as much as possible.

%
Considering the compatibility between different transformations, we do not apply some transformations together, such as cutout and blur, which will make the target image too corrupted to be recognized.
It is worth noting that the augmentation of flip that is commonly adopted in the general self-supervised learning methods~\cite{SimCLR,moco} has a bad effect on training a tracking model (see results in Section~\ref{sec:effect-of-trans}). 
The evaluation of each transformation on the overall tracking performance can be found in Section~\ref{sec:effect-of-trans}. 
Furthermore, we validate the capability of these transformations to deal with the corresponding tracking challenges in Section~\ref{sec:challenge_ori}.

\vspace{1mm}
\noindent\textbf{Paste: simulation of background changes and interference.}
This operation pastes the transformed target patches onto other images (background) to synthesize desired images. 
We employ this operation to generate target shifts and context changes by selecting a random location to paste at and a random frame of image to paste onto, respectively. 
Thus, we are able to get the target localization and simulate background changes. 
Specifically, we consider two Paste options which are pasting the transformed target patches onto images from the same sequence (same-sequence paste) or a different sequence (different-sequence paste).
In addition, we select similar object patches from other videos by comparing the feature distance and paste them to the background for simulating background distractors in tracking scenarios.

To alleviate the artifacts of sharp boundaries caused by the pasting operation, we crop the target patch with pads and perform the pasting with a blending sum at the pad area.
The experimental evaluation and analysis on the selection of images for pasting onto can be found in Section~\ref{sec:study-CTP}.

\subsection{Discussions}

The proposed method adapts to synthesized data based on a specified target in the initial frame of a test video without human annotations.
%
The only requited annotation (manually or by an object detector) is to initialize the object location in the first frame, which is the common practice in online object tracking.
Different from existing unsupervised tracking methods which mainly aim to detect moving blobs, the proposed algorithm is based on self-supervision to accurately localize objects (e.g., object extents).
Existing self-supervised learning methods typically construct representation models by exploiting large-scale unlabeled image data for vision tasks, where the goals significantly different from that in this paper. 
Specifically, we highlight two key differences between the self-supervised learning methods for representation learning (abbreviated as SSL-RL) and our method:
\begin{compactitem}
    \item Different operation framework: A prominent way adopted by SSL-RL is to train an individual feature learning module following the contrastive learning framework. In contrast, our method focuses on synthesizing target-aware training data, which allows training existing trackers without human annotations. Thus, our proposed self-learning mechanism can be integrated into any existing tracking framework.
    \item Different data transformation techniques: SSL-RL performs data transformations to construct training pairs to train its feature learning module, while our method designs specific tracking-related transformations to simulate appearance variations during tracking.
    Owing to different objectives, different data transformation techniques are adopted. For instance, the flip augmentation, which is effective in self-supervised learning for representation learning, has an adverse effect on self-supervised learning for visual tracking.
\end{compactitem}

\section{Investigation of the Proposed Method}
In this section, we study the proposed algorithm thoroughly with four sets of experiments.
First, we conduct an ablation study to show the contribution of each component and the detailed designs.
Second, we show the effect of the challenge-oriented transformation strategy.
Third, we investigate the effect of each kind of transformation in improving tracking performance.
Finally, we compare the performance of the SiamRPN++ variants trained using different numbers of annotated samples.
In the following, we first present experimental settings and then introduce every set of experiments.

\subsection{Experimental Settings}

\noindent 
\textbf{Pipeline.}
Given a tracking sequence, our approach first generates the target exemplar patch based on the initial information. 
Then, we perform the tracking-related transformations on the target exemplar and paste the target exemplar onto other frames to generate the synthesized frames. 
The target states in the synthesized data can be computed based on the target exemplar and the specific transformations.
Finally, we randomly select two frames as the pair of template and search region to perform training, which enables training deep trackers in routine ways.

\noindent\textbf{Implementation details.} We investigate the proposed target-aware data synthesis method under the SiamRPN++~\cite{SiamRPN++} framework.
We perform training using the default settings as the baseline method except for the settings related to the synthesized training data.
The synthesized samples are generated based on the specified target patch cropped from the initial frame of a video, which is a basic setting in the visual tracking task.
It takes about 30 hours to get converged on a server with 4 V100 GPU cards.
To select the trained models for testing, we test the last 8 epochs from the 20 epochs and select the best performance as the final results, just like the strategy adopted by existing methods~\cite{SiamRPN++,SiamBAN}.
Note that we do not fine-tune the hyper-parameters, since there are hundreds of trained models to be evaluated, especially for the ablation study.
As pointed out by the authors~\cite{SiamRPN++}, the result scores obtained without fine-tuning maybe a little lower than the reported scores.
To ensure fair comparisons, we use the default testing hyper-parameters for all the evaluations and perform each evaluation on multiple testing datasets.
We implement the code in Python with PyTorch using the pysot~\cite{SiamRPN++} tookit.

\vspace{1mm}
\noindent\textbf{Training and testing datasets.}
For the ablation studies in Section~\ref{sec:study-CTP} and Section~\ref{sec:effect-of-trans}, only the ImageNet VID data is used for saving training loads.
For other experiments, we use the same training datasets as those used by the baseline model, \ie ~ImageNet VID~\cite{VID}, ImageNet DET~\cite{VID}, COCO~\cite{COCO}, and Youtube-BB~\cite{youtubeBB}.
Our approach only uses one annotated frame of a video for getting the base target patch, which is the same as the setting of the online tracking task.
After that, we use the base target patch as the reference image and generate a synthesized image sequence using the proposed Crop-Transform-Paste operation.
We use 3 datasets for evaluation including OTB100 (diverse variations)~\cite{OTB2015}, NFS (NFS30 version)~\cite{NFS}, and UAV123 (captured in a UAV view)~\cite{UAV123}.
The area under the curve (AUC) of the success plot (based on the overlap ratio of the predicted and ground-truth bounding boxes)~\cite{OTB2013} is used to report the results.

\begin{table}[hp]
\footnotesize
    \centering
    \caption{\textbf{Performance of different target patch generation strategies.} We use SiamRPN++ as the base model and perform training on the ImageNet VID dataset~\cite{VID} with labels generated by different strategies.}
    \vspace{1mm}
    \resizebox{0.95\linewidth}{!}{
    \begin{tabular}{c|cccc}
    \toprule
    Different scheme &  SS-random  &SS-detector & SS-annotated & Su.  \\
    \midrule
      OTB100 (AUC \%)   &54.8  &  61.9 &63.4   &64.2\\
      UAV123 (AUC \%)   &44.1  &  51.9 &52.1   &55.8\\
      NFS    (AUC \%)   &37.6  &  44.1  &44.8   &44.7\\
    \bottomrule
    \end{tabular}}
    \label{tab:patch-generation}
    \vspace{-2mm}
\end{table}

\subsection{Study on Detailed Designs}
\label{sec:study-CTP}
We first study the design of the proposed method including the target patch generation strategies, and the selection of the images to be pasted onto.
We then compare the proposed algorithm with its variants using only the copy-paste operation and only the transformations without copy-paste.

\vspace{1mm}
\noindent\textbf{Generation of base target patch.}
The proposed Crop-Transform-Paste mechanism first needs base target patches for generating synthesized data.
We test three ways to generate target patches, which are using randomly cropped patches as the target patch (SS-random), target patches generated using the YOLOv4~\cite{Mosaic} detector (SS-detector), and using the specified target patch in the initial frame of a video(SS-annotated).
We compare the results of these SiamRPN++ variants trained on the synthesized data with the above ways, and the variant trained using annotated data (Su).

Table~\ref{tab:patch-generation} shows the AUC scores of the SiamRPN++ variants on the OTB100, UAV123, and NFS datasets.
The comparison of the first three columns shows that using annotated patches or patches generated by a general detector achieves significantly better performance than that using random crop.
This is because the annotated and detected target patches contain more effective semantic features and more object boundary information, which is crucial for generating high-quality synthesized samples that simulate target appearance variations during tracking.
The last three columns show that using the synthesized data achieves a close performance against that obtained by using the annotated data, which demonstrates the effectiveness of the proposed data generation method.

\vspace{1mm}
\noindent\textbf{Selection of background images.}
\begin{table}[t]
\footnotesize
    \centering
    \caption{\textbf{Effect of background image selection.} The table compares the results of using different strategies for generating synthesized data which are pasting the target patch onto the image from the same sequence or a different sequence, and using manually annotated data.}
    \vspace{1mm}
    \begin{tabular}{ccccc}
    \toprule
   &\multicolumn{1}{r|}{Dataset}  &  OTB100 & UAV123 & NFS\\
  \multicolumn{1}{l}{Scheme}& \multicolumn{1}{r|}{} & AUC (\%)& AUC (\%)&AUC (\%)\\
     \midrule
     \multicolumn{2}{l|}{Different-sequence paste} & 55.0 & 45.0 & 36.2\\
    \multicolumn{2}{l|}{Same-sequence paste}   & 63.4 & 52.1 & 44.8\\
    \multicolumn{2}{l|}{Manual annotation}  & 64.2 &  55.8& 44.7\\
     \bottomrule
    \end{tabular}

    \label{tab:paste-comparison}
    \vspace{-5mm}
\end{table}
For the paste operation, we can select background images from the same video (where the target patch is cropped from) or different videos to paste onto.
Table~\ref{tab:paste-comparison} shows the results of the SiamRPN++ variants trained on the data generated using these two ways.
It shows that performing the Crop-Transform-Paste operation on the images from the same video achieves better performance than that on different videos and close performance to that obtained by the supervised learning scheme.
This is because the background from the same video usually contains similar negative samples and distractors, whose sample distribution fits better to real testing scenarios.

\subsection{Challenge-Oriented Transformation}
\label{sec:challenge_ori}

\begin{table*}[ht]
\footnotesize
    \centering
    \caption{\textbf{Performance of the transformation schemes for solving the related tracking challenges under the Self-Supervised (SS) and the integrated (SS+Su) schemes.} SV, OCC, MB, BC, and SPatchP denote Scale Variation, Occlusion, Motion Blur, Background clutter, and Similar-Patch Paste, respectively.
    The evaluation is conducted on the sequences tagged with the corresponding challenges from the OTB100 dataset.}
    \vspace{1mm}
    \resizebox{1.0\linewidth}{!}{
    \begin{tabular}{l| rrrrrrrr}
    \toprule
      & \multicolumn{4}{c|}{SS scheme} &\multicolumn{4}{c}{SS+Su scheme}\\
   \multicolumn{1}{r|}{AUC score (\%)}   &\multicolumn{1}{c}{Rescale-SV}& \multicolumn{1}{c}{Cutout-OCC}& \multicolumn{1}{c}{Blur-MB}& \multicolumn{1}{c|}{SPatchP-BC}&\multicolumn{1}{c}{Rescale-SV}& \multicolumn{1}{c}{Cutout-OCC}& \multicolumn{1}{c}{Blur-MB}& \multicolumn{1}{c}{SPatchP-BC}\\
     \midrule
     W/o related transformation  & 45.1          &57.7           &55.7  & 57.7  &64.1  &59.3   & 61.9 &64.1\\
     W/  related transformation  & (\textbf{+17.1}) 62.2 &(\textbf{+1.2}) 58.9  &(\textbf{+3.1}) 58.8  & (\textbf{+4.7}) 62.4  &(\textbf{+2.6}) 67.0  &(\textbf{+4.7}) 64.0   &(\textbf{+3.4}) 65.3 &(\textbf{+5.2}) 69.3\\
   \bottomrule
    \end{tabular}
   }
    \label{tab:ablation-challenge}

\end{table*}

One main advantage of our Crop-Transform-Paste operation is that we can customize transformation operations to synthesize effective data for dealing with specific tracking challenges.
To analyze the capability to handle different tracking challenges, we perform experiments to evaluate the effect of our proposed transformations on their targeted challenges.
Specifically, we consider four customized transformations for tackling four corresponding typical tracking challenges: 1) \textbf{rescale} for Scale Variation (SV); 2) \textbf{Cutout} for Occlusion (OCC); 3) \textbf{Blur} for Motion Blur (MB) and 4) \textbf{Similar-Patch Paste} (SPatchP) for Background Clutter (BC).

We train different variants of the SiamRPN++ model with and without each of the four transformations and keep other training settings fixed.
The evaluation of these models is conducted on the sequences tagged with the corresponding attribute from the OTB100 dataset.
In addition, we conduct the experiments under both the Self-Supervised (SS) learning scheme and the integrated scheme of Self-Supervised learning and Supervised learning (SS+Su).

\begin{figure}[t]
    \centering
    \subfigure[SS scheme.]{
        \begin{minipage}[b]{0.45\linewidth}
             \includegraphics[width=1\linewidth]{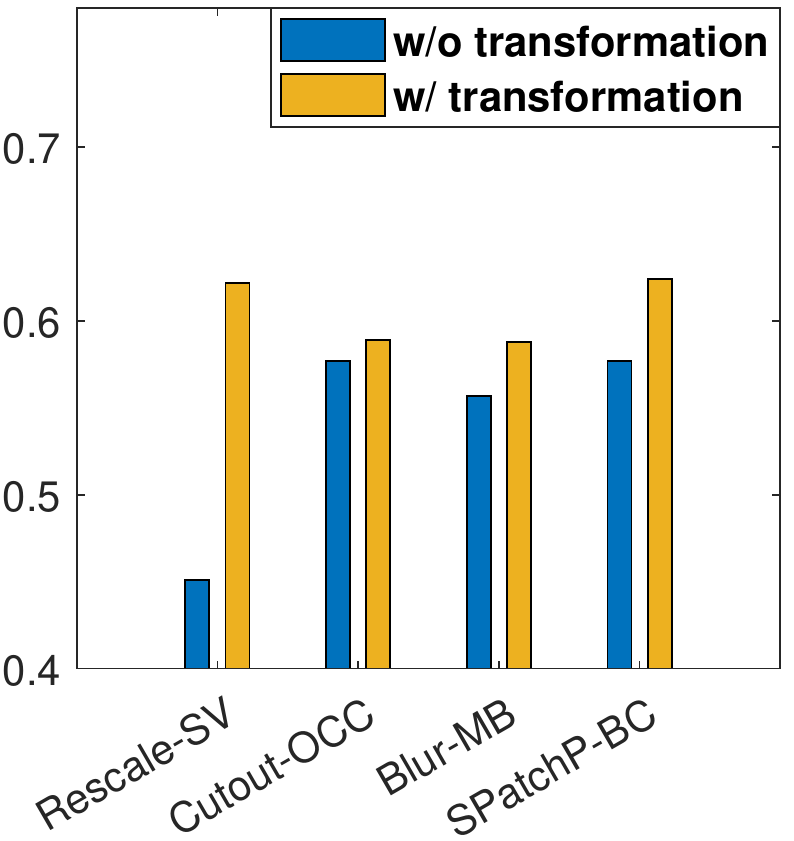}
    \end{minipage}
    }
    \subfigure[SS+Su scheme.]{
        \begin{minipage}[b]{0.45\linewidth}
             \includegraphics[width=1\linewidth]{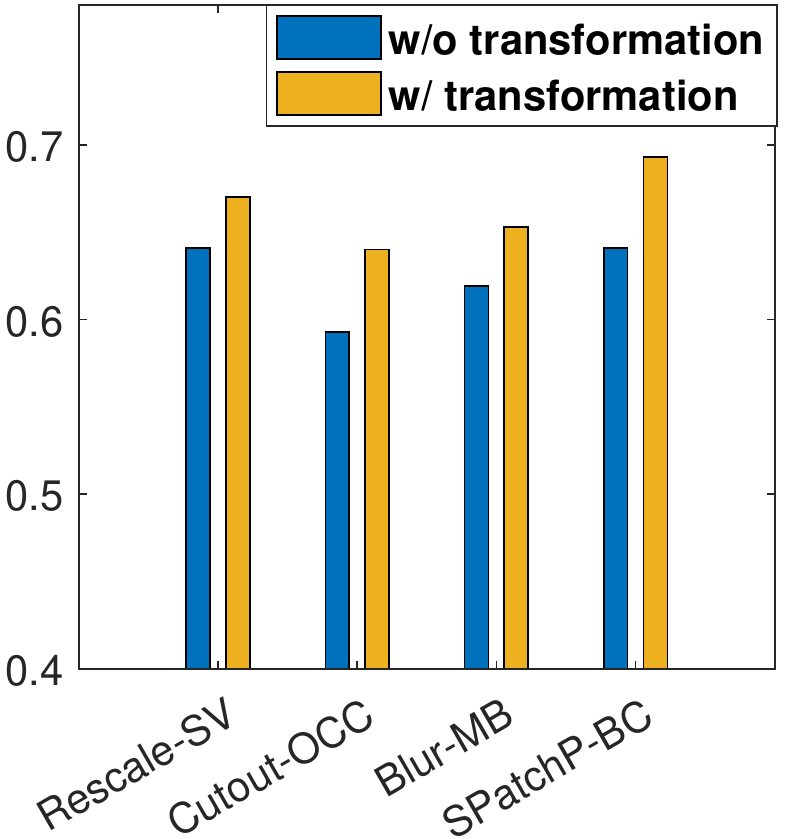}
        \end{minipage}
    }
    \caption{\textbf{Performance on the transformations for solving the corresponding tracking challenges in terms of the AUC score.} SV, OCC, MB, BC, and SPatchP denote Scale Variation, Occlusion, Motion Blur, Background clutter, and Similar-Patch Paste, respectively. The evaluation is conducted on the sequences tagged with the corresponding challenges from the OTB100 dataset. }
\label{fig:aug-attri}
\vspace{-5mm}
\end{figure}

Figure~\ref{fig:aug-attri} presents the performance in terms of the AUC score regarding each transformation on the corresponding test sequences under both schemes.
Table~\ref{tab:ablation-challenge} shows the detailed scores.
The transformations all achieve remarkable improvements on the sequences with the corresponding tracking challenges under both learning schemes.
The rescale, cutout, blur, and similar-patch paste transformations achieve absolute performance gains of 17.0\%, 1.2\%, 3.1\%, and 4.7\% under the Self-Supervised learning scheme, and absolute gains of 2.6\%, 4.7\%, 3.4\%, and 5.2\% under the integrated scheme of self-supervised learning and supervised learning in terms of the AUC score.
The gains under the SS scheme indicate that the synthesized training data generated with the rescale, cutout, blur, and similar-patch paste transformations are effective in providing appearance and background interference that are related to the challenging tracking scenario with scale variation, occlusion, motion blur, and background clutter, which facilitate the training of a tracking model to handle these challenges during tracking.
The gains under the SS+Su scheme show that the synthesized training data are also helpful for further improving the model trained using manually annotated data.
This indicates that the synthesized data provides supplementary information to the manually annotated data for learning a robust tracking model, which is attributed to the range of appearance variations simulated by the proposed method.

The evaluation results of these two schemes show 
that the cutout, blur, and similar-patch paste transformations even achieve larger gains under the integrated scheme of self-supervised learning and supervised learning on the test sequences with the challenging variations of Occlusion, Motion blur, and Background Clutter, while the rescale transformation obtain a relative smaller improvement.
This indicates that the synthesized data provides more supplementary information to the manually annotated training data in terms of the Occlusion, Motion blur, and Background Clutter variations compared to the Scale variation.
This is because the manually selected and annotated training samples may not contain sufficient challenging scenarios of Occlusion, Motion blur, and Background Clutter which are not commonly seen and only last for a relatively short time in real video.
In contrast, samples of these cases can be easily generated with the proposed Crop-Transform-Paste operation by setting the rescale ratio within a reasonable range and selecting a rescale ratio uniformly from this range for each synthesized sample.
The favorable performance indicates that the controllability of the variation range and distribution of the training samples is a specific merit of the proposed self-supervised learning scheme for training tracking models.

Overall, the performance improvements on all these test cases demonstrate the effectiveness of the proposed approach in tackling tracking challenges.

\subsection{Effect of Every Transformation Schedule}
\label{sec:effect-of-trans}
We analyze the effect of each transformation in generating synthesized data for visual tracking by comparing the performance of the tracking models trained with and without each of them.
For a concise and clear comparison, we only use the ImageNet VID training dataset (using 100 K pairs per epoch) to train the baseline tracking model (SiamRPN++), since larger scale training data will make the training process more sensitive to learning rates and more difficult to converge.
We use the same settings except for those related to the transformation schedule in all these experiments.
We set up two groups of experiments for evaluating the contribution of each transformation schedule.
The first group shows the performance gains by adding one transformation to the baseline version that only uses the shift transformation (denoted as Baseline1).
The other group presents the performance drops by removing one transformation from the baseline version that uses all the transformations (denoted as Baseline2).

\begin{table}[t]
\footnotesize
    \centering
    \caption{\textbf{Evaluation on different transformation operators.} The last column (Avg-Var) shows the Average  Variations compared (over three datasets) with their corresponding baseline in terms of the AUC score (\%), which measures the effect. The \textcolor{red}{red} and \textcolor{blue}{blue} fonts denote positive and negative effects.}
    \vspace{1mm}
    \resizebox{0.8\linewidth}{!}{
    \begin{tabular}{l| c c c c}
    \toprule
      Schedule& OTB100 &UAV123 &NFS& Avg-Var\\
     \midrule
     Baseline1          & 46.5& 37.6&28.8& -- \\ 
     + rescale            & 58.5&49.2 &39.4& \textcolor{red}{+11.4}\\
     + flip             & 42.6& 36.8&27.9& \textcolor{blue}{-1.8}\\
     + shear            & 46.7&37.7 &30.9& \textcolor{red}{+0.8}\\
     + cutout           & 45.8&39.6 &28.9& \textcolor{red}{+0.5}\\
     + blur             & 42.5&35.9 &30.2& \textcolor{blue}{-1.4} \\
     + SPatchP       & 50.3&40.1 &30.5& \textcolor{red}{+2.7}\\
     + CJ           & 42.8&34.8&28.3 & \textcolor{blue}{-2.3}\\
     \midrule
     Baseline2          &62.0 &55.7 &45.4& --\\
     - shift            &51.4 &47.5 &36.1&\textcolor{red}{-9.4}\\
     - rescale            & 45.9& 38.6&32.8&\textcolor{red}{-15.3}\\
     - flip             &63.4 &55.1 &45.2&\textcolor{blue}{+0.2}\\
     - shear            & 60.7&53.2 &44.3&\textcolor{red}{-1.7}\\
     - cutout           &61.2 & 53.8&44.7&\textcolor{red}{-1.5}\\
     - blur             & 60.2&50.2 &38.4&\textcolor{red}{-4.8}\\
     - SPatchP       &61.1& 53.1& 44.0&\textcolor{red}{-1.7}\\
     - CJ               &61.1 &  53.2 &45.1& \textcolor{red}{-1.3}\\
   \bottomrule
    \end{tabular}
    }
    \label{tab:ablation-trans}
    \vspace{-5mm}
\end{table}

Table~\ref{tab:ablation-trans} shows the AUC scores (\%) regarding to different transformation schedules on the OTB100, NFS, and UAV123 datasets.
These schedules include random locations (shift) with the range of $\pm$96 pixels, rescale with the range of 0.7-1.3, flip with the probability of 0.5, shear with the probability of 0.5, cutout with the probability of 0.15, shaking blur (blur) with the probability of 0.2, Similar-Patch Paste (SPatchP) with the probability of 0.8, and color jittering (CJ) with the probability of 0.4.
%
The detailed settings of variation ranges and probabilities are empirically set.

We have the following observations based on the effect of each transformation schedule from Table~\ref{tab:ablation-trans}.
First, the transformations of shift, rescale, shear, cutout, and similar-patch paste play positive roles in both groups, namely adding them to Baseline1 achieves better performance, meanwhile removing them from Baseline2 leads to performance drops.
This observation indicates that the variances provided by these transformations help a tracker to learn invariance encodings for recognizing a target object with variations, and they are effective when exploited together.
Among these transformations, rescale and shift play the two most important roles in improving tracking performance.
This is because scale and shift are the most commonly seen variations during tracking, which are crucial for training a tracker.

Second, the blur and color jitter transformations have negative effects when applied alone.
The negative effects may be caused by the overfitting issue to the blur or color jitter variations, since using them alone only provides limited appearance variations.
Instead, when applied together with other augmentations they contribute to performance gains.
This is because they provide similar variations in the training data that are helpful for learning to locate a target object with motion blur or illumination variations.

Third, the flip augmentation brings negative effects no matter applied alone or together with others.
This is because it ruins the boundary information of the target, which is prone to confuse the bounding box predictor during training.
In general, only the transformation schedules related to target variations in tracking scenarios contribute to performance gains under the proposed self-supervised data synthesis scheme.

\begin{figure}[t]

    \centering
    \subfigure[Performance on UAV123.]{
        \begin{minipage}[b]{0.4\linewidth}
             \includegraphics[width=1\linewidth]{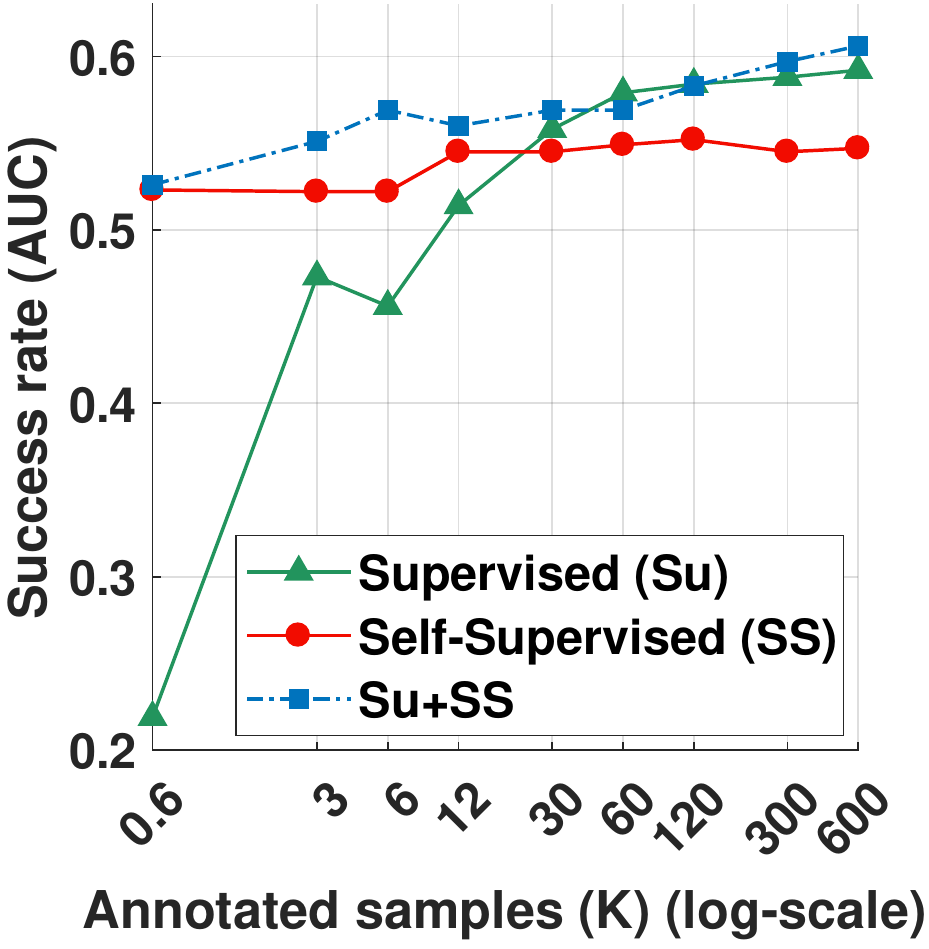}
    \end{minipage}
    }
    \subfigure[Performance on NFS.]{
        \begin{minipage}[b]{0.4\linewidth}
             \includegraphics[width=1\linewidth]{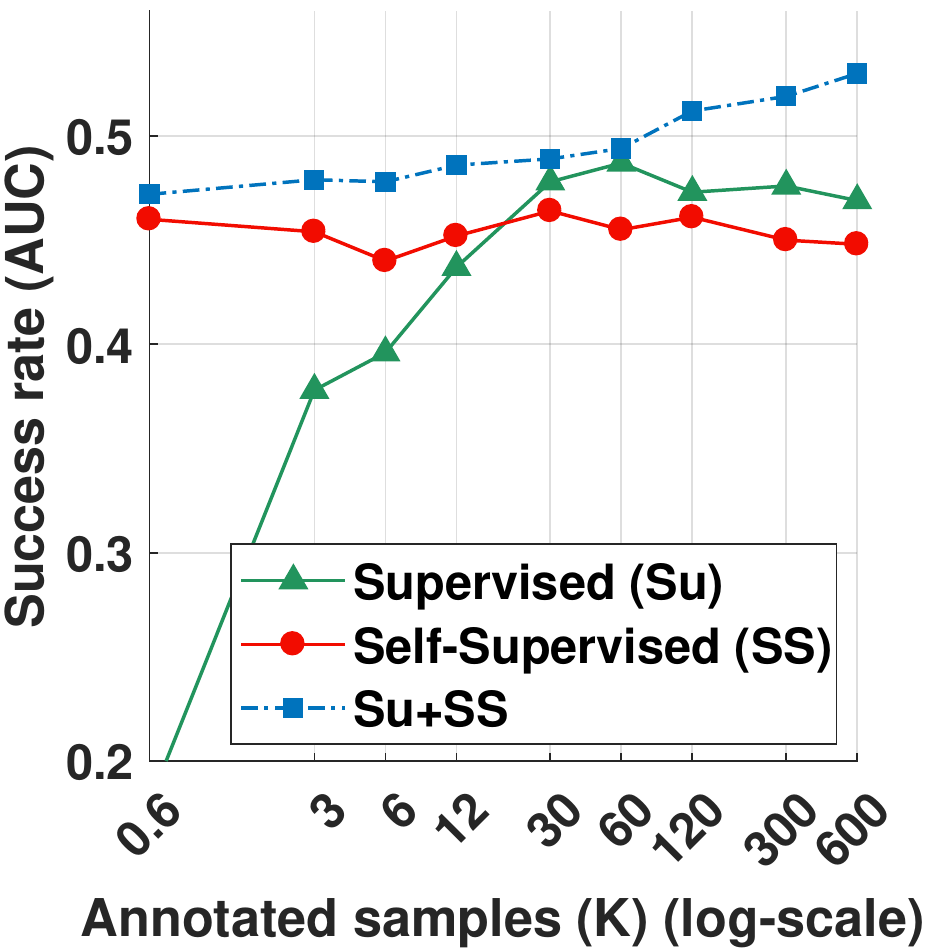}
        \end{minipage}
    }
    \caption{\textbf{Evaluation on data efficiency using different learning schemes for training the SiamRPN++ model.} It shows that the self-supervised learning scheme achieves significant improvements on both datasets especially when only a few annotated samples are available. }
\label{fig:sl+su-uav-nfs}
\vspace{-2mm}
\end{figure}

\subsection{Data Efficiency}
With the proposed Crop-Transform-Paste scheme, we are able to generate massive synthesized training samples with various appearance variations only based on a few annotations for indicating the target object.
To show how this helps data efficiency, we evaluate the SiamRPN++ model trained using different numbers of annotated samples with the Self-Supervised (SS), Supervised (Su), and combined (SS+Su) learning schemes, respectively.
Figure~\ref{fig:sl+su-uav-nfs} shows the AUC scores on the UAV123 and NFS datasets.
The results on the OTB100 dataset are shown in Figure~\ref{fig:sl+su}.

\begin{table}[ht]
\footnotesize
    \centering
    \caption{\textbf{Performance on data efficiency using different learning schemes for training the SiamRPN++ model in terms of the AUC score (\%).} It shows that the self-supervised learning scheme achieves significant improvements especially when only a few annotated samples are available. }
    \vspace{1mm}
    \resizebox{1.0\linewidth}{!}{
    \begin{tabular}{ll|cccccc|c }
    \toprule

    & & \multicolumn{6}{c}{ Number of annotated samples (K)}&\multicolumn{1}{c}{ Sample number (K)}\\
   Dataset &Scheme &0.6 & 3& 6   & 60 & 300  &600& 600    \\
     \midrule
   \multirow{3}{*}{OTB100}  & SS    &  46.7 & 63.7  &65.3   &64.6  & 64.8 &63.6&65.3 \\
                            & Su    &  20.6 & 45.9  &50.6   & 65.4  & 66.4 &66.2&66.5 \\
                            & Su+SS & 61.6  & 65.6  &65.9   &67.3  & 68.2 &68.4& (+1.7) 68.2 \\
      \midrule
    \multirow{3}{*}{UAV123} & SS    &  52.3& 52.2&52.2& 54.9& 54.5  &54.7&55.2\\
                            & Su    &  21.9&47.3 &45.6&57.9 &58.8   &59.2& 59.2\\
                            & Su+SS &  52.6&55.1 &56.9&56.9 &59.7   &60.6&(+0.5) 59.7\\
      \midrule
\multirow{3}{*}{NFS}    & SS    & 46.0  & 45.4&44.0 & 45.5 &45.0 &44.8& 46.4\\
                        & Su    &  18.0 &37.8 &39.6 &48.7  &47.6 &46.9& 48.7\\
                        & Su+SS &  47.2 &47.9 &47.8 &49.4&51.9 &53.0& (+3.2) 51.9\\
   \bottomrule
    \end{tabular}
   }
    \label{tab:data-efficiency}

\end{table}




\begin{figure*}[t]
    \centering
        \includegraphics[width=0.9\textwidth]{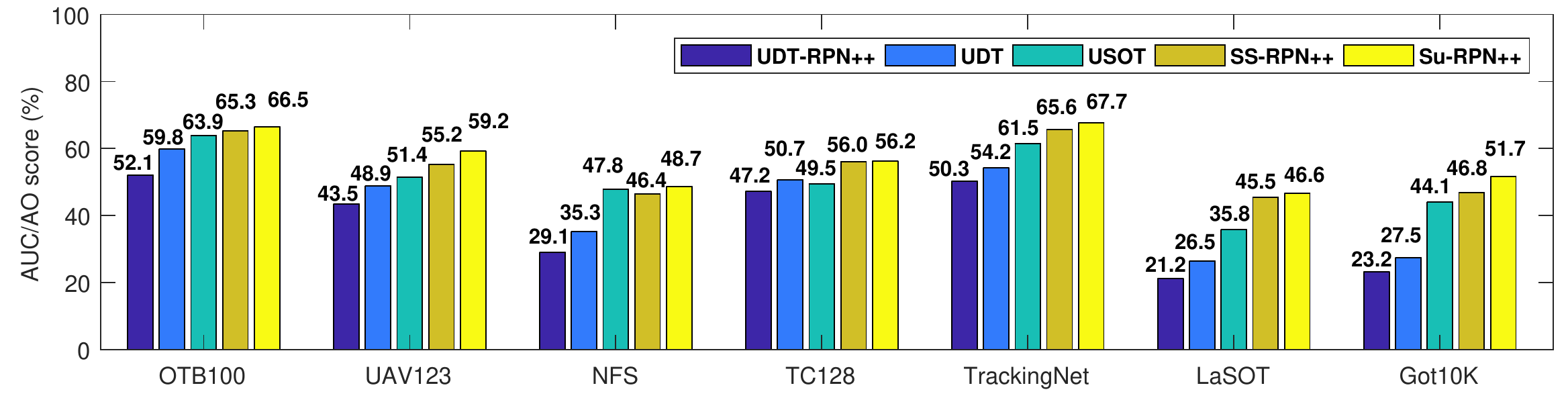}
    \caption{\textbf{Evaluation on the OTB100, UAV123, NFS, TC128 TrackingNet, LaSOT, and Got10k datasets against other methods.} SS-RPN++ and Su-RPN++ denote the SiamRPN++ models trained with the synthesized data and the manually annotated data, respectively. It shows that the proposed self-supervised scheme performs favorably against other methods.
}
    \label{fig:overall-comparison}
    \vspace{-3mm}
\end{figure*}

Table~\ref{tab:data-efficiency} shows the AUC scores of different SiamRPN++ variants trained with different schemes.
The model trained using the self-supervised learning scheme only needs 0.6K annotations to get converged and achieves the AUC scores of 52.3\% and 46.0\% on the UAV123 and NFS datasets, respectively.
In contrast, the model trained with the supervised learning scheme (using manually annotated samples) only obtains the AUC scores of 21.9\% and 18.0\% when using the same number of annotated samples for training.
The supervised learning scheme takes more than 60K annotated samples to converge.
The performance gaps regarding these two schemes demonstrate that the proposed self-supervised learning method improves the data efficiency effectively and is able to train a tracker effectively in few-shot tracking scenarios with limited annotations.
It is worth noting that the proposed self-supervised learning scheme achieves the AUC scores of 52.6\% and 47.2\% with 0.6K annotated samples and improves the final performance of the supervised learning scheme by +1.4\% and +4.3\% AUC scores on the UAV123 and NFS datasets, respectively.
The last column of Table~\ref{tab:data-efficiency} shows the results of the models that are trained using the same number of training samples (600K).
For the Su scheme, all the training samples are manually annotated, while for the SS+Su scheme, half of the samples are generated by the proposed method.
These improvements demonstrate that the proposed method can enhance the supervised learning scheme in both terms of data efficiency and tracking accuracy.

\begin{table*}[ht]
\footnotesize
\renewcommand\arraystretch{1.1} 
    \caption{\textbf{Evaluation on the integration of the proposed self-supervised learning scheme into the SiamRPN++, DiMP, and TransT methods on the OTB100, UAV123, NFS, TC128, and LaSOT datasets.} Su and SS denote the supervised learning and self-supervised learning schemes, respectively.
    Taking Su as the baseline, the scores in the parentheses show the absolute gains. }
\vspace{1mm}
\centering
\resizebox{0.9\linewidth}{!}{
\begin{tabular}{lrrrrrrrrrr}
\toprule
\multicolumn{1}{r}{Dataset} & \multicolumn{2}{c}{OTB100} & \multicolumn{2}{c}{UAV123} & \multicolumn{2}{c}{NFS} & \multicolumn{2}{c}{TC128} & \multicolumn{2}{c}{LaSOT}\\
\multicolumn{1}{l}{Model} & Pre(\%) & AUC (\%) &  Pre(\%) &AUC (\%)&  Pre(\%) &AUC (\%) &  Pre(\%) &AUC  (\%)&  AUC  (\%) & NPre  (\%)\\
\midrule
SiamRPN++  & 88.1& 66.5  & 77.6&59.2   & 57.0 &48.7  & 76.0& 56.2 & 47.6 & 56.1\\
SS+SiamRPN++  & (\textbf{+2.2}) 90.3  &(\textbf{+2.0}) 68.5 &(\textbf{+3.5}) 81.1 & (\textbf{+1.4}) 60.6   & (\textbf{5.9}) 62.9& (\textbf{+4.3}) 53.0  & (\textbf{+3.2}) 79.2&(\textbf{+2.0}) 58.2 & (\textbf{+2.3}) 49.9 & (\textbf{+3.5}) 59.6\\
\midrule
DiMP     & 88.2& 68.4  & 86.1&65.3   & 73.2&62.0  & 79.3& 58.8 & 56.8 & 64.8\\
SS+DiMP   &(\textbf{+1.6}) 89.8 &  (\textbf{+0.6}) 69.0  & (\textbf{+1.0}) 87.1 & (\textbf{+1.0}) 66.3   & (\textbf{+0.6}) 73.8 &(\textbf{+0.3}) 62.3  & (\textbf{+2.8}) 82.1&(\textbf{+1.5}) 60.3 &   (\textbf{+0.6}) 57.4 & (\textbf{+0.7}) 65.5\\
\midrule
Transt     & 88.6& 69.0  & 85.0&66.2   & 80.0&65.7  &78.0 & 58.2 & 64.5 & 73.8 \\
        
SS+Transt   &(\textbf{+1.7}) 90.3 &  (\textbf{+0.6}) 69.6  &(\textbf{+2.6}) 87.6 & (\textbf{+1.8}) 68.0   & (\textbf{+0.5}) 80.5 &(\textbf{+1.2}) 66.9  & (\textbf{+1.9}) 79.9&(\textbf{+1.2}) 59.4&   (\textbf{+1.8}) 66.3 & (\textbf{+1.6}) 75.6\\
\midrule
Average gains&  \textbf{+1.8} & \textbf{+1.1} & \textbf{+2.4} & \textbf{+1.4} &  \textbf{+2.3} & \textbf{+1.9} & \textbf{+2.6} & \textbf{+1.6} &\textbf{+1.6} &\textbf{+1.9} \\
\bottomrule
\end{tabular}
}
\label{tab:integration}
\end{table*}

\subsection{Limitation}
One limitation of the proposed method is that pasting a rectangular target image patch to a random frame may result in artifacts. We alleviate this issue by pasting the target patch to the frames from the same video. This could also be improved by using a mask annotation rather than a box annotation but it requires more detailed labeling.
\section{Experimental Comparison with Other Methods}
In this section, we evaluate the proposed algorithm by comparing with the state-of-the-art methods.
First, we compare the proposed algorithm with existing unsupervised trackers under unsupervised settings.
Then, we show the effectiveness in boosting the performance of existing supervised methods.

\subsection{Comparison against Unsupervised Methods}
\vspace{2mm}
\noindent\textbf{Detailed settings.} We train the SiamRPN++~\cite{SiamRPN++} model on synthesized data that are generated based on the specified target in the initial frame of a video, which is a basic setting in the visual tracking task.
Other training and testing settings are kept the same with those of the base model.
We compare the trained model using only the synthesized data with existing unsupervised methods including 
UDT~\cite{UDT}, UDT-RPN++ (UDT under the SiamRPN++ framework), USOT~\cite{USOT}, and Su-RPN++ (the SiamRPN++ model trained using annotated data) on the datasets of OTB100~\cite{OTB2015}, NFS (NFS30 version)~\cite{NFS}, TC128~\cite{TC128}, UAV123~\cite{UAV123}, LaSOT (280 long-term sequences)~\cite{LaSOT}, TrackingNet (511 sequences)~\cite{TrackingNet}, and Got10K~\cite{GOT10k}.
The AUC/AO score are used to report the results.

\begin{figure*}[!htb]
\begin{center}

		\begin{minipage}[c]{0.9\linewidth}
            \includegraphics[width=0.24\linewidth]{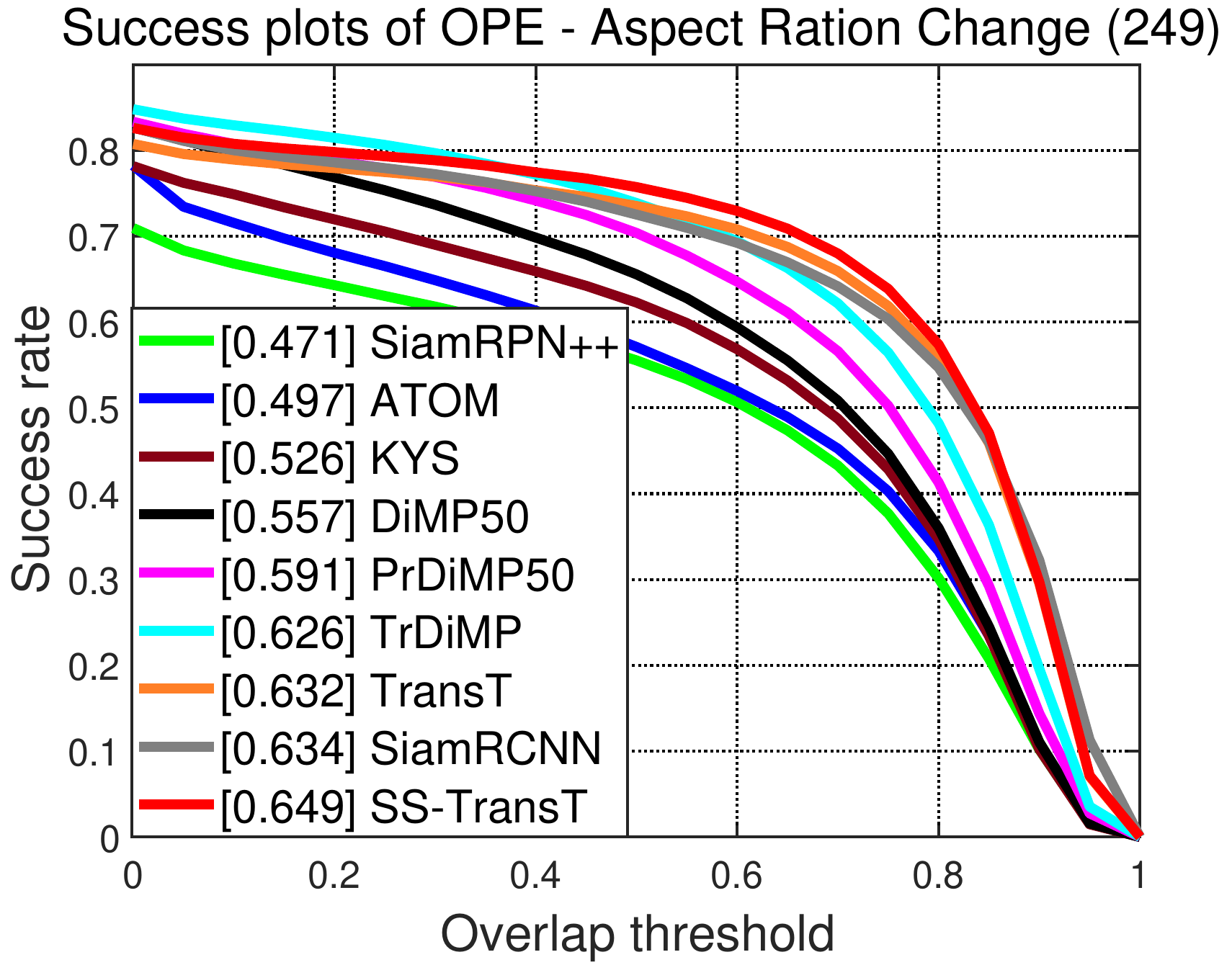}\vspace{0.5mm}
			\includegraphics[width=0.24\linewidth]{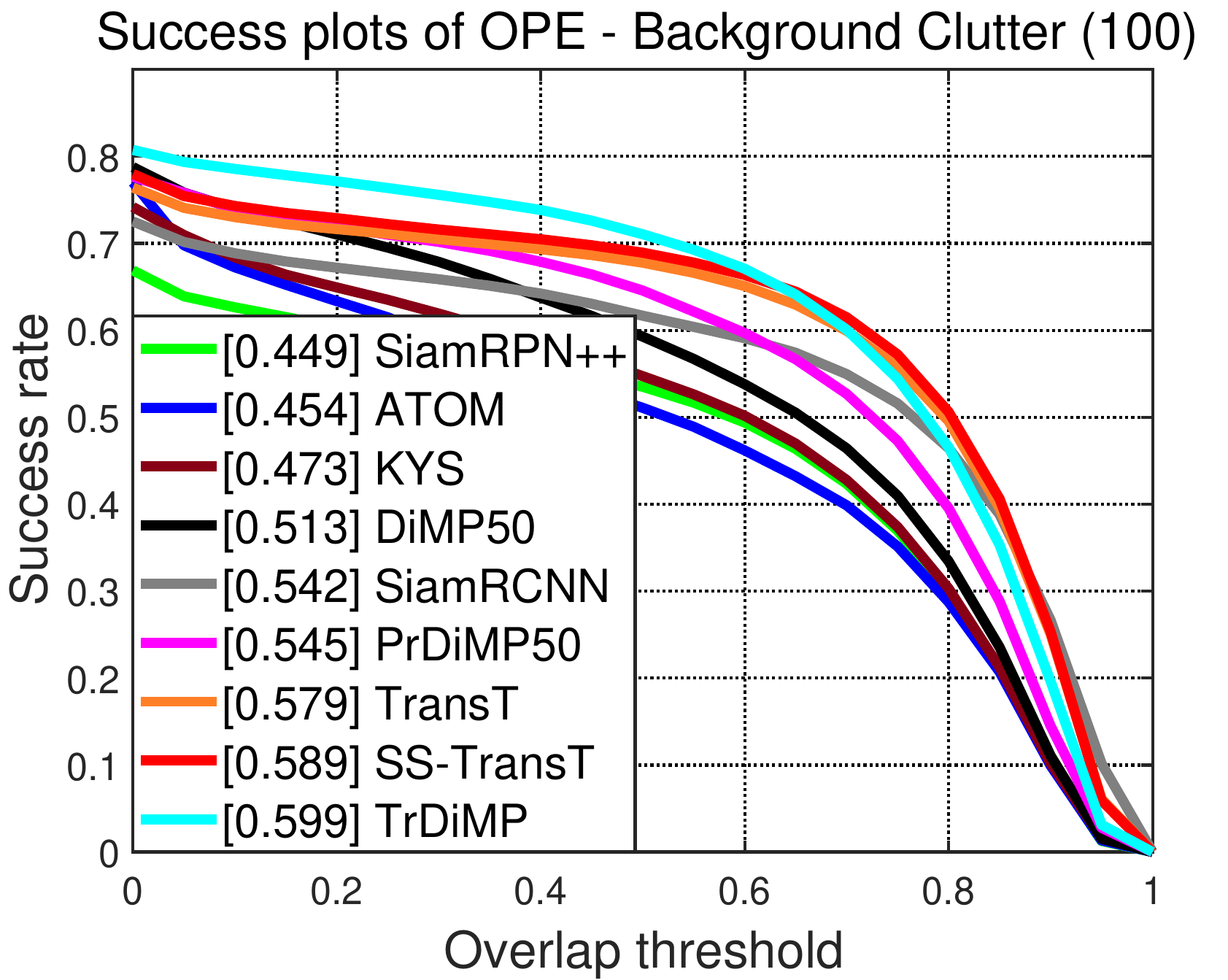}
			\includegraphics[width=0.24\linewidth]{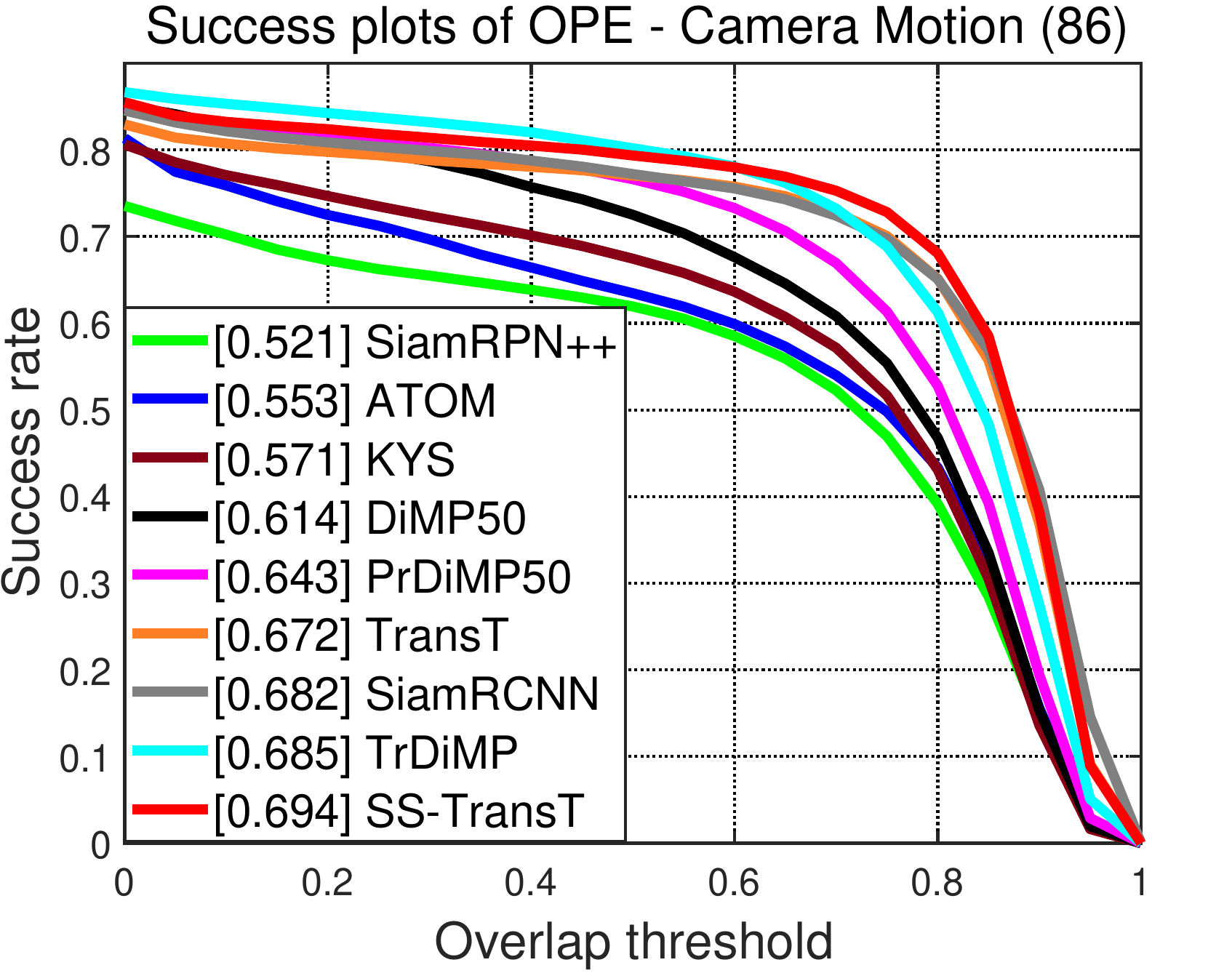}
			\includegraphics[width=0.24\linewidth]{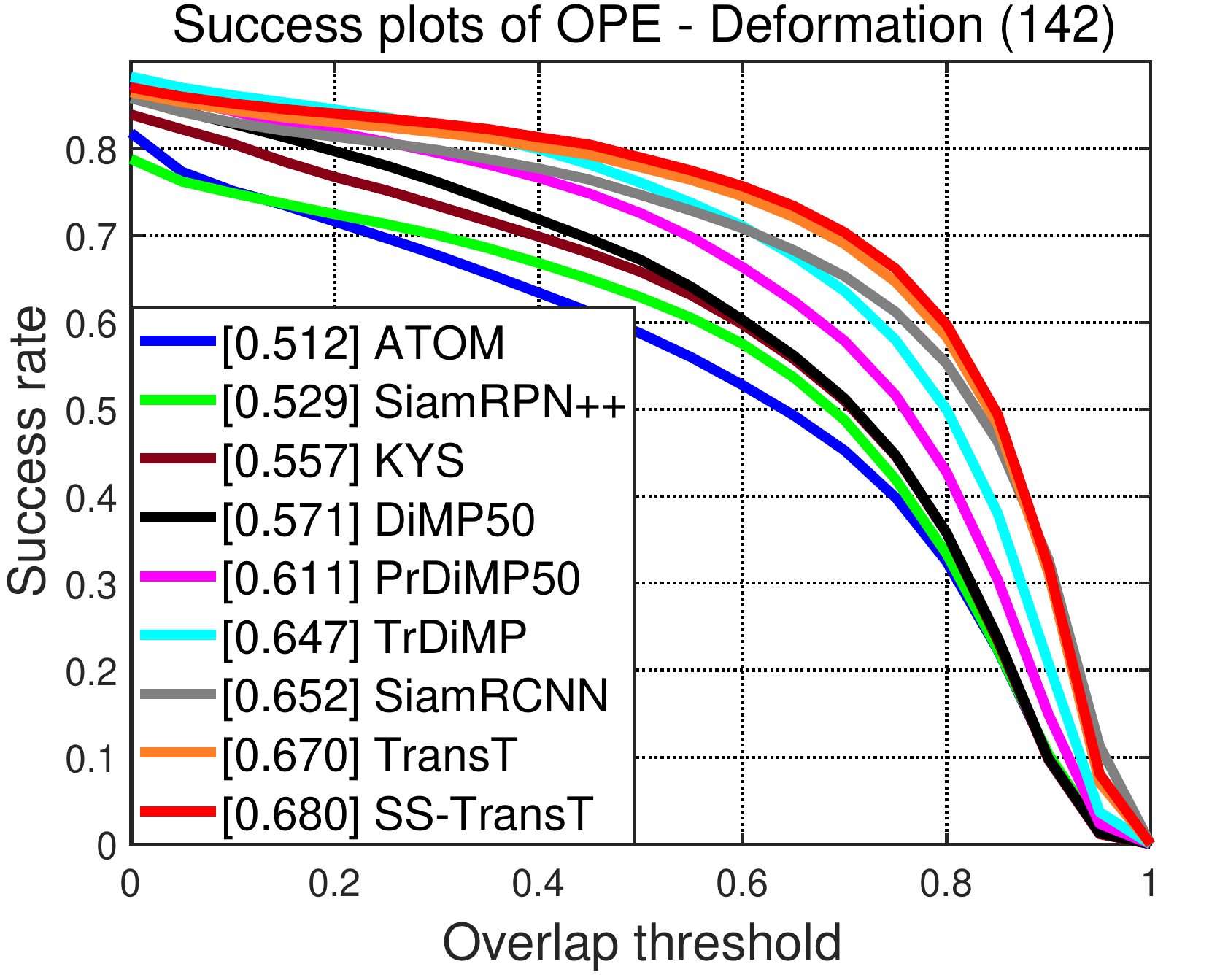}
		\end{minipage}
		\begin{minipage}[c]{0.9\linewidth}
            \includegraphics[width=0.24\linewidth]{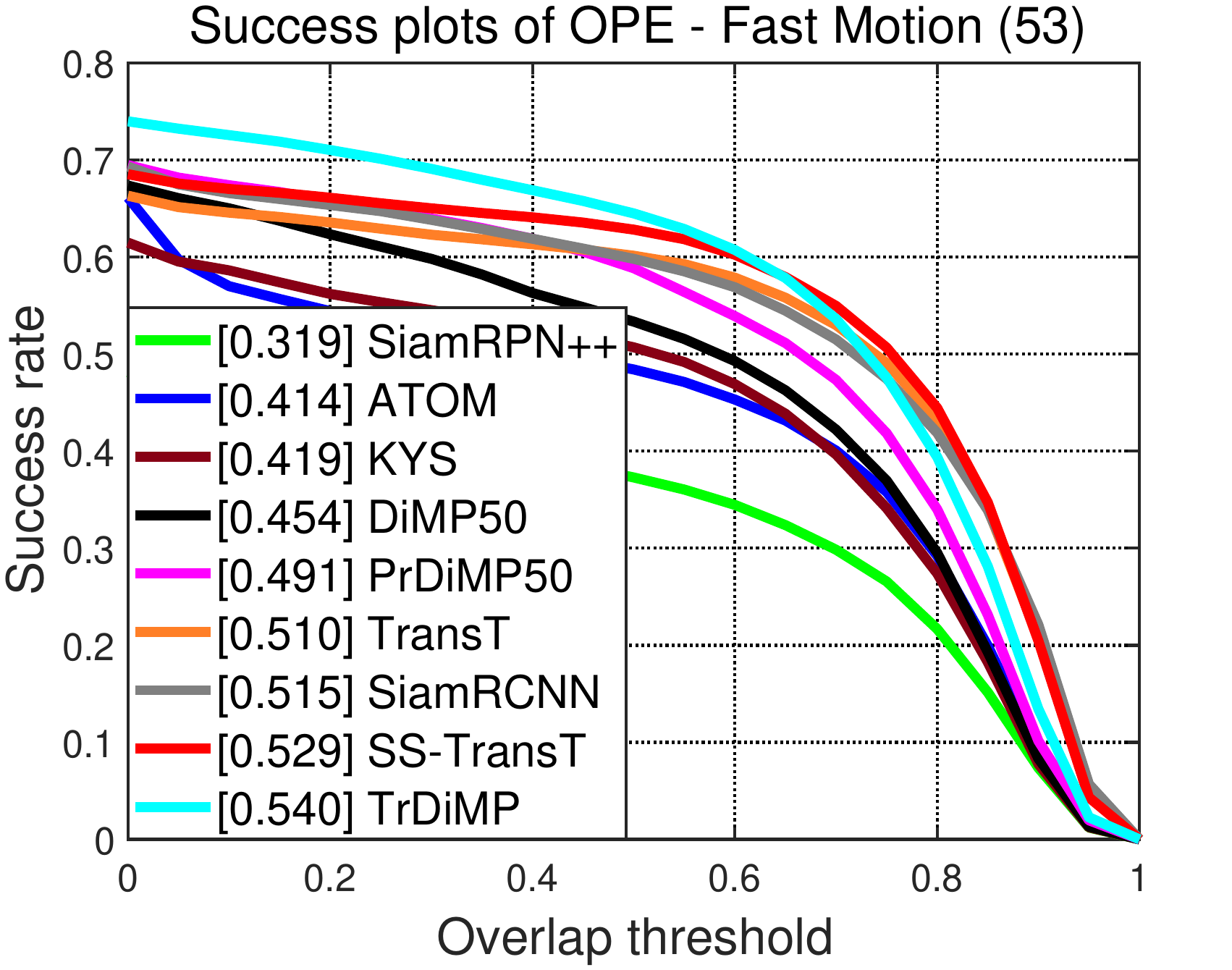}\vspace{0.5mm}
			\includegraphics[width=0.24\linewidth]{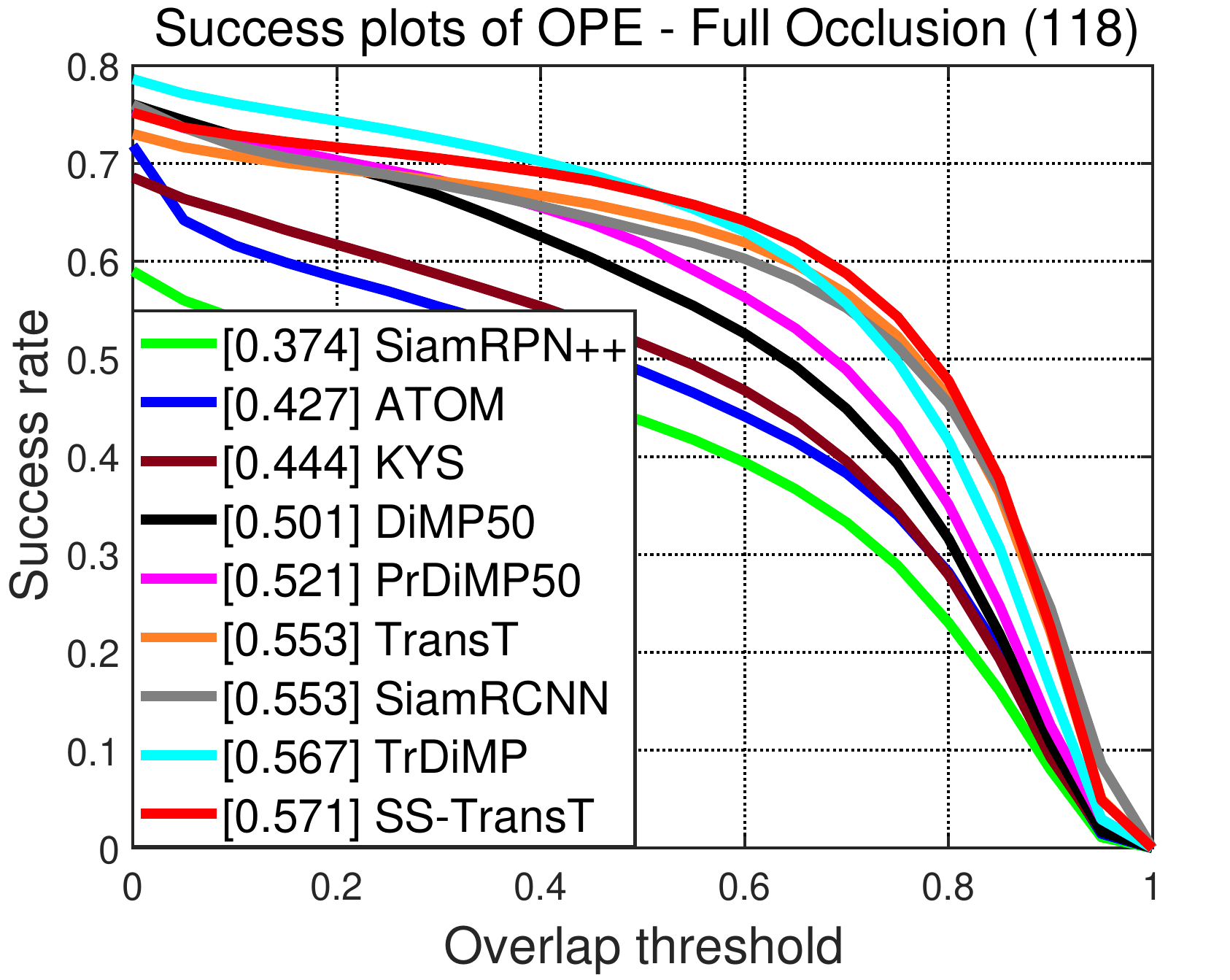}
			\includegraphics[width=0.24\linewidth]{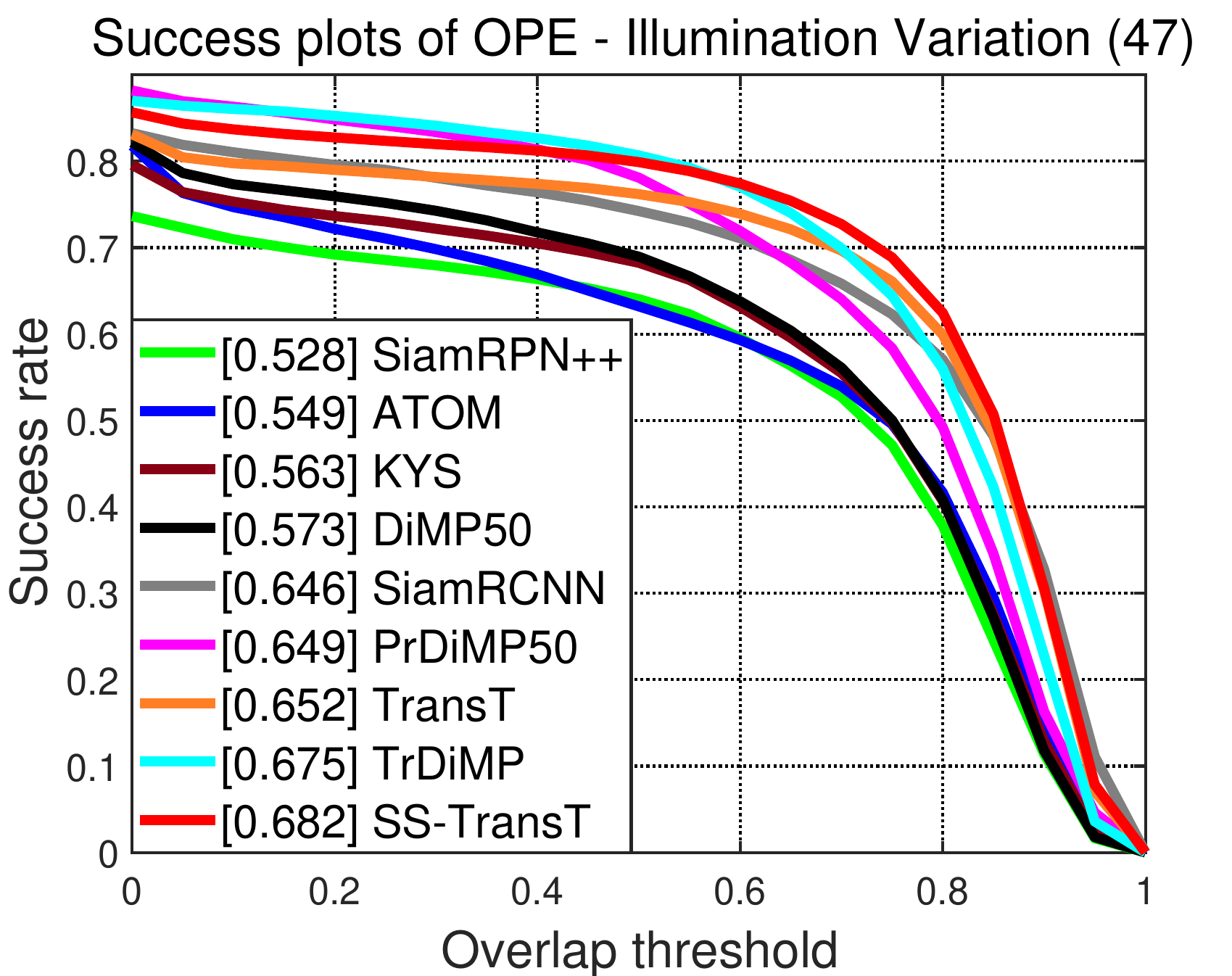}
			\includegraphics[width=0.24\linewidth]{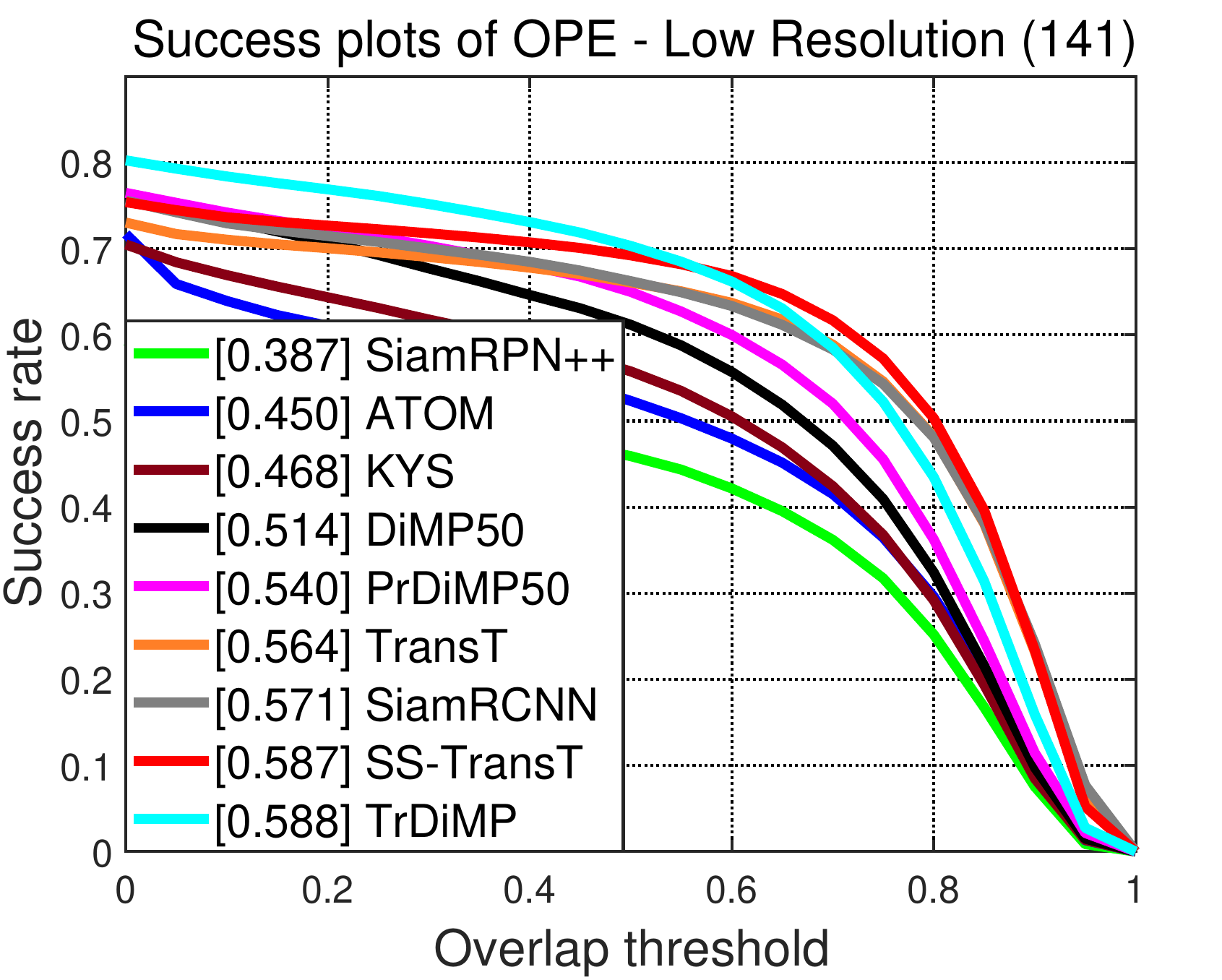}
		\end{minipage}
		\begin{minipage}[c]{0.9\linewidth}
            \includegraphics[width=0.24\linewidth]{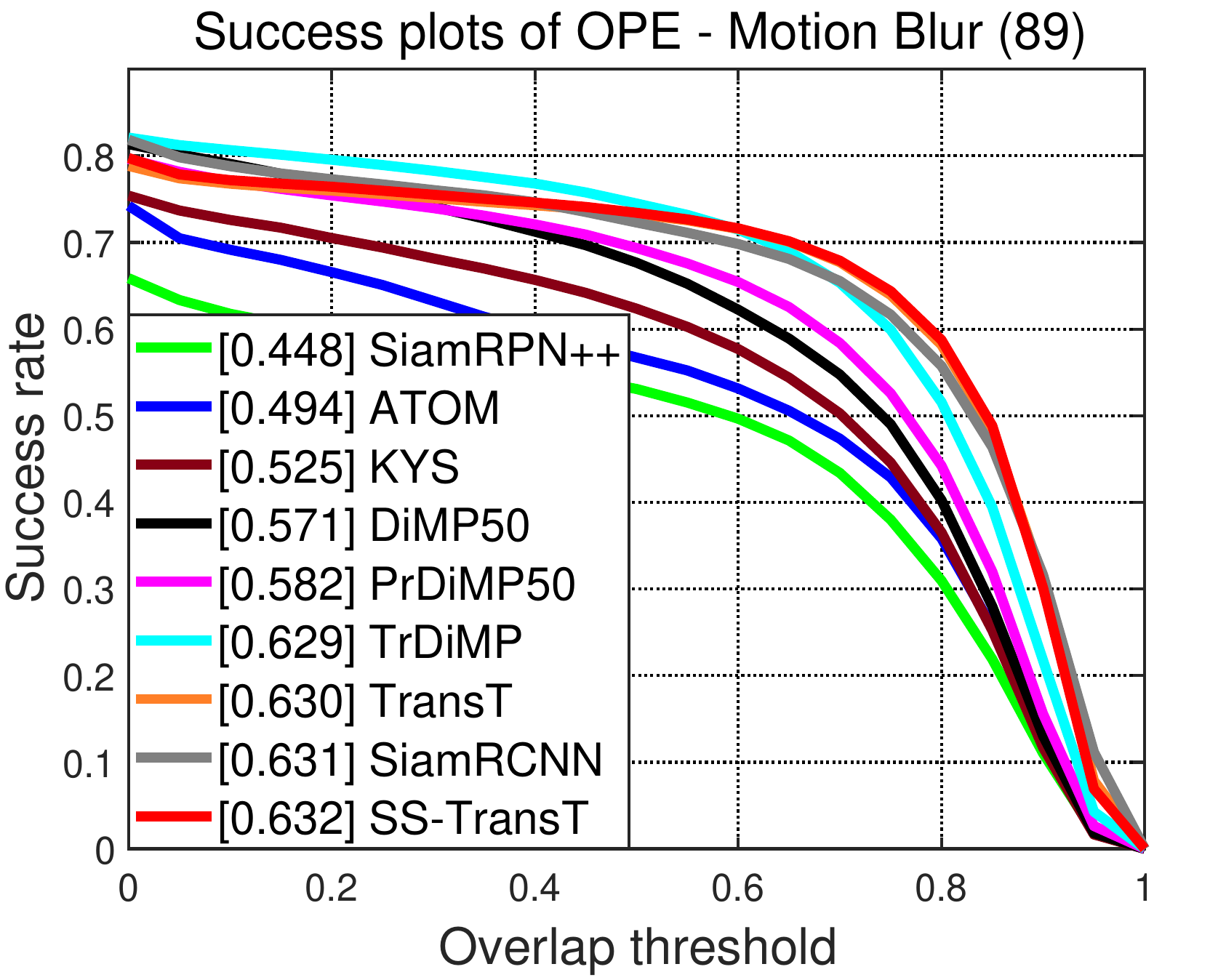}\vspace{0.5mm}
			\includegraphics[width=0.24\linewidth]{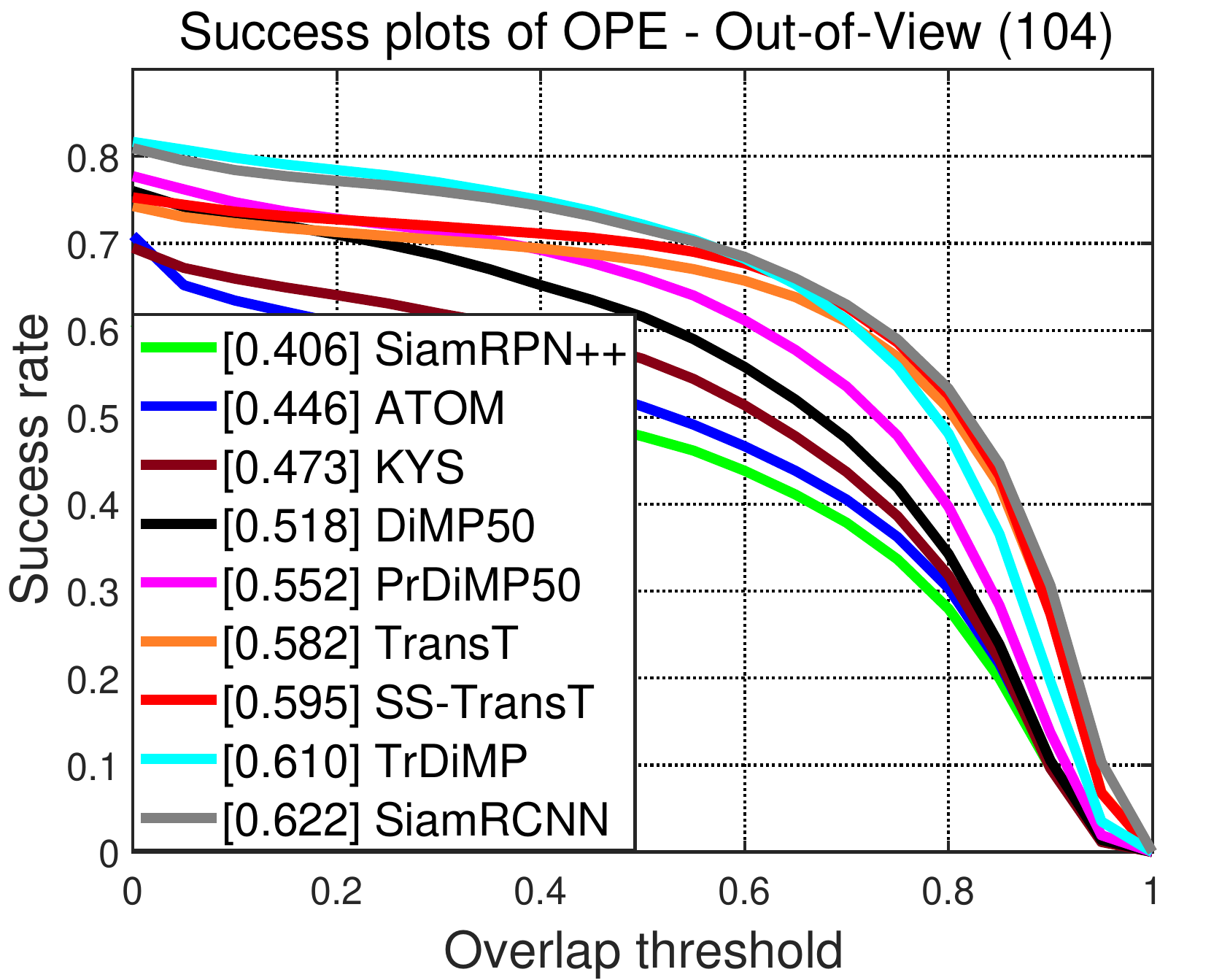}
			\includegraphics[width=0.24\linewidth]{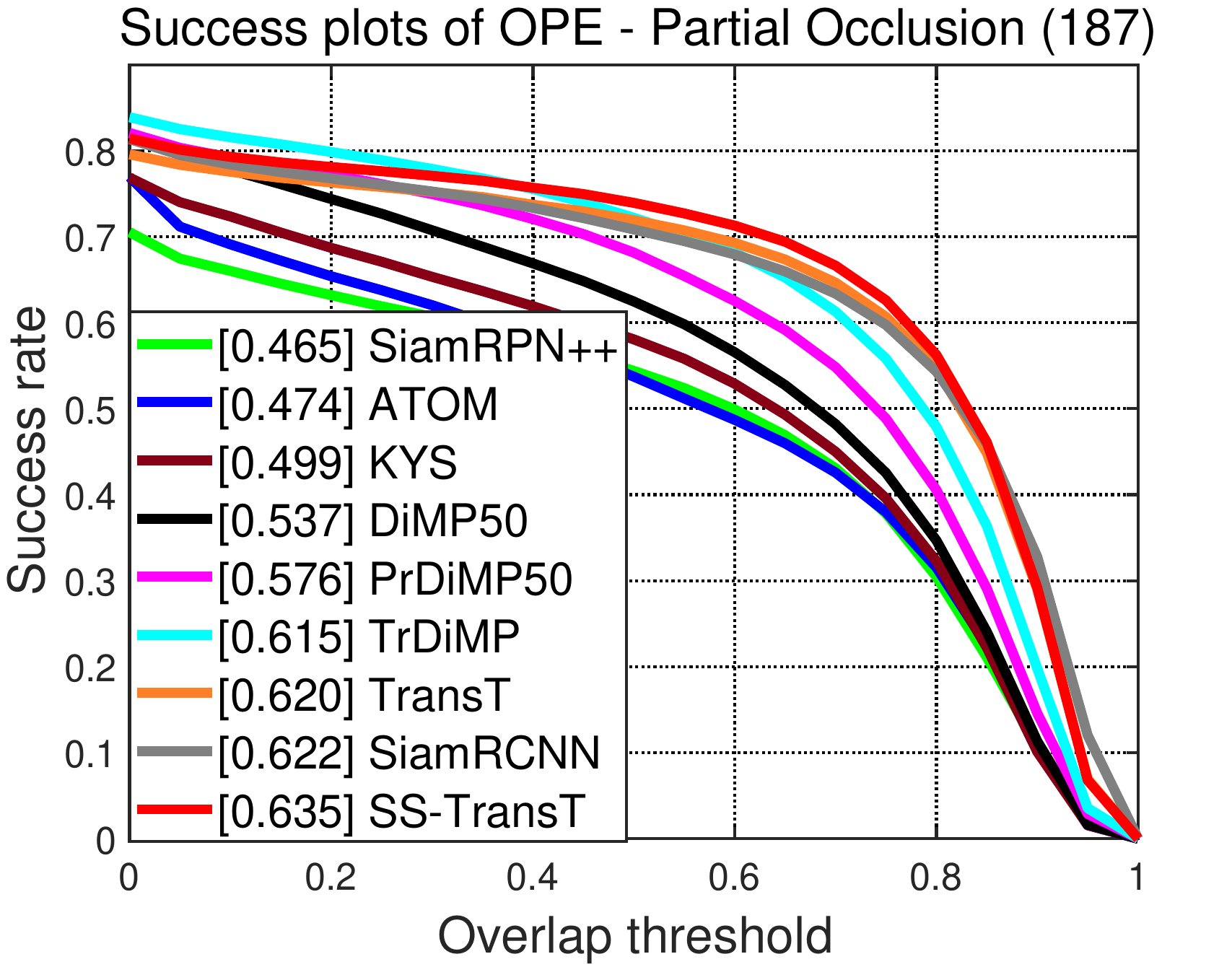}
			\includegraphics[width=0.24\linewidth]{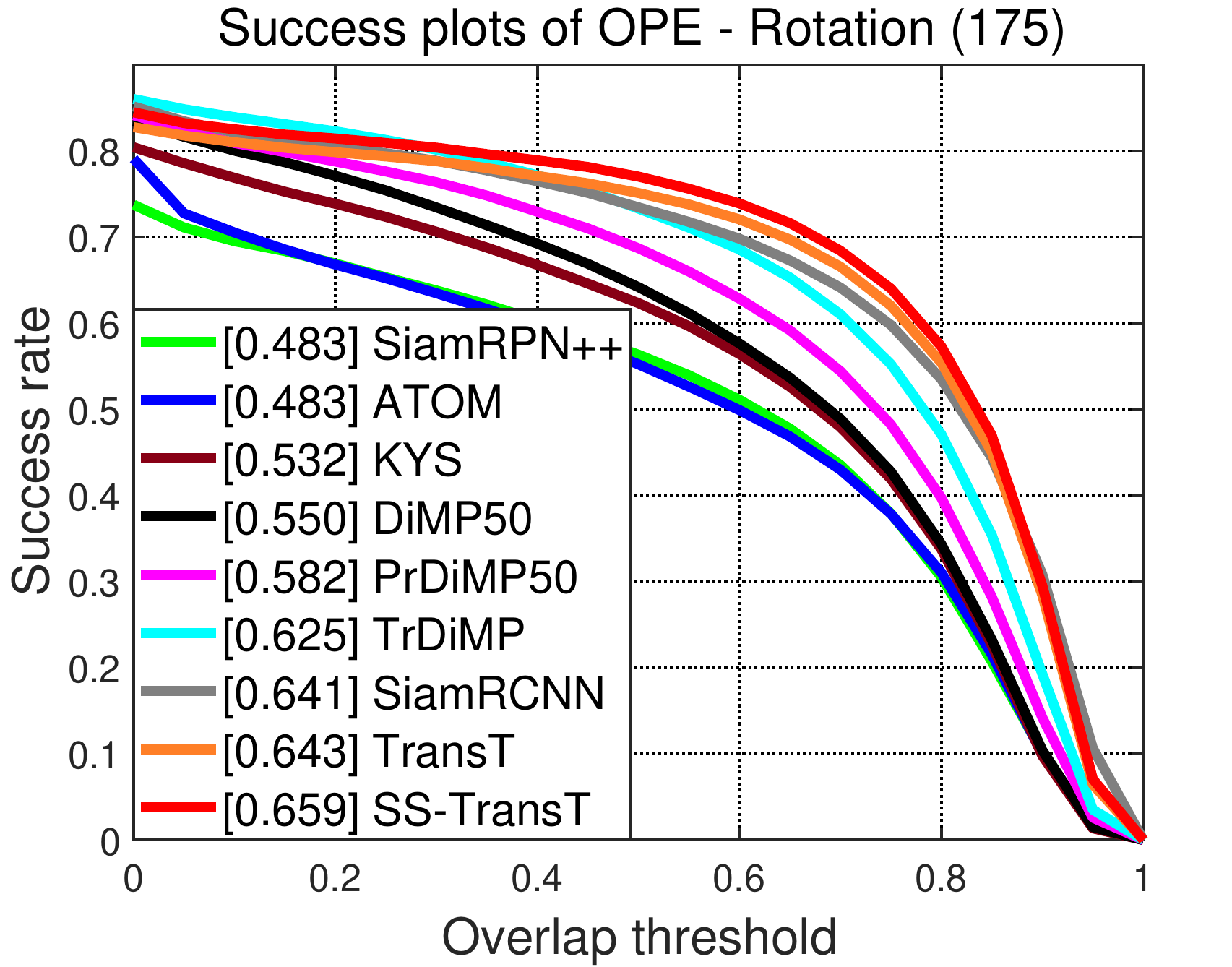}
		\end{minipage}
		\begin{minipage}[c]{0.9\linewidth}
            \includegraphics[width=0.24\linewidth]{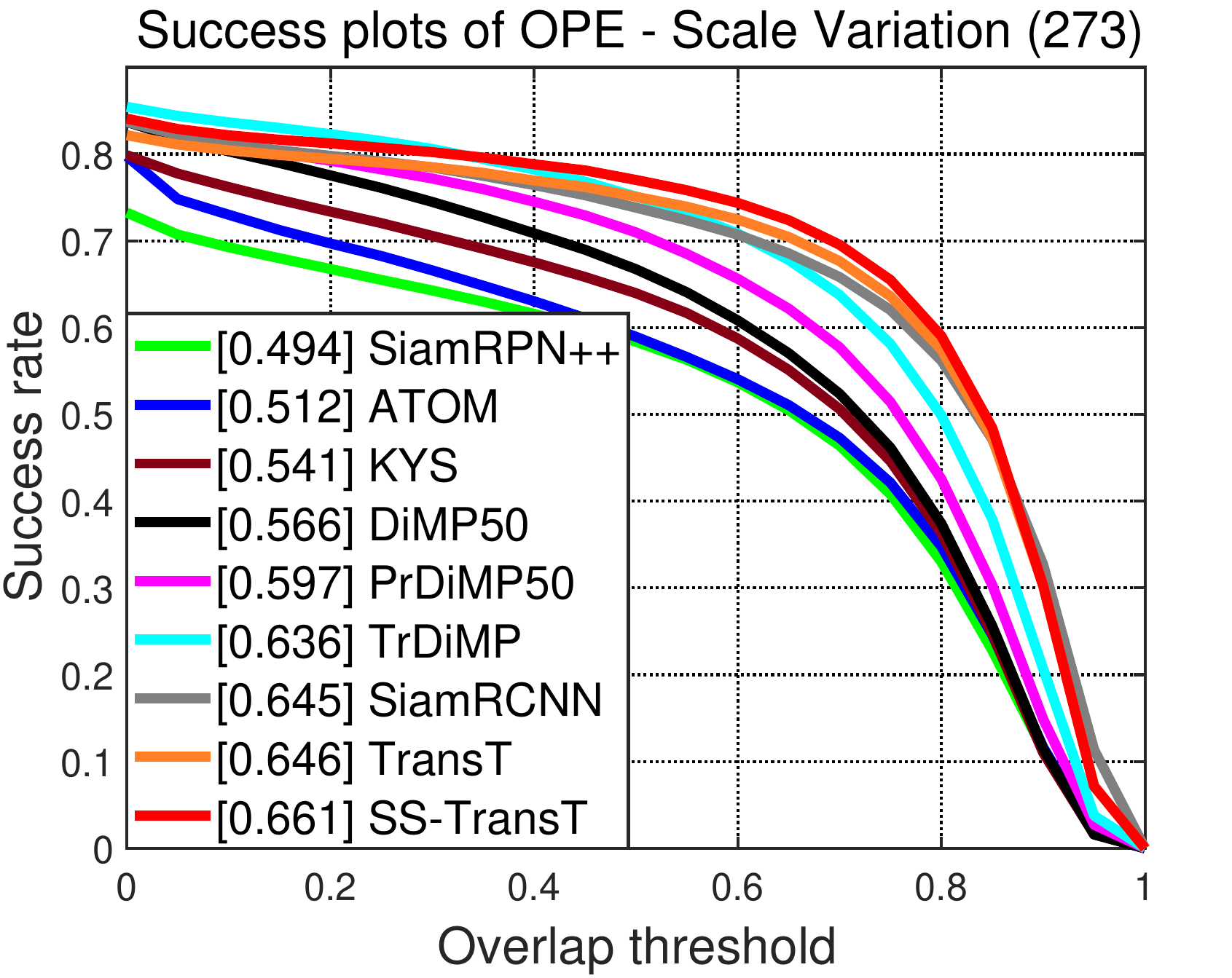}\vspace{0.5mm}
			\includegraphics[width=0.24\linewidth]{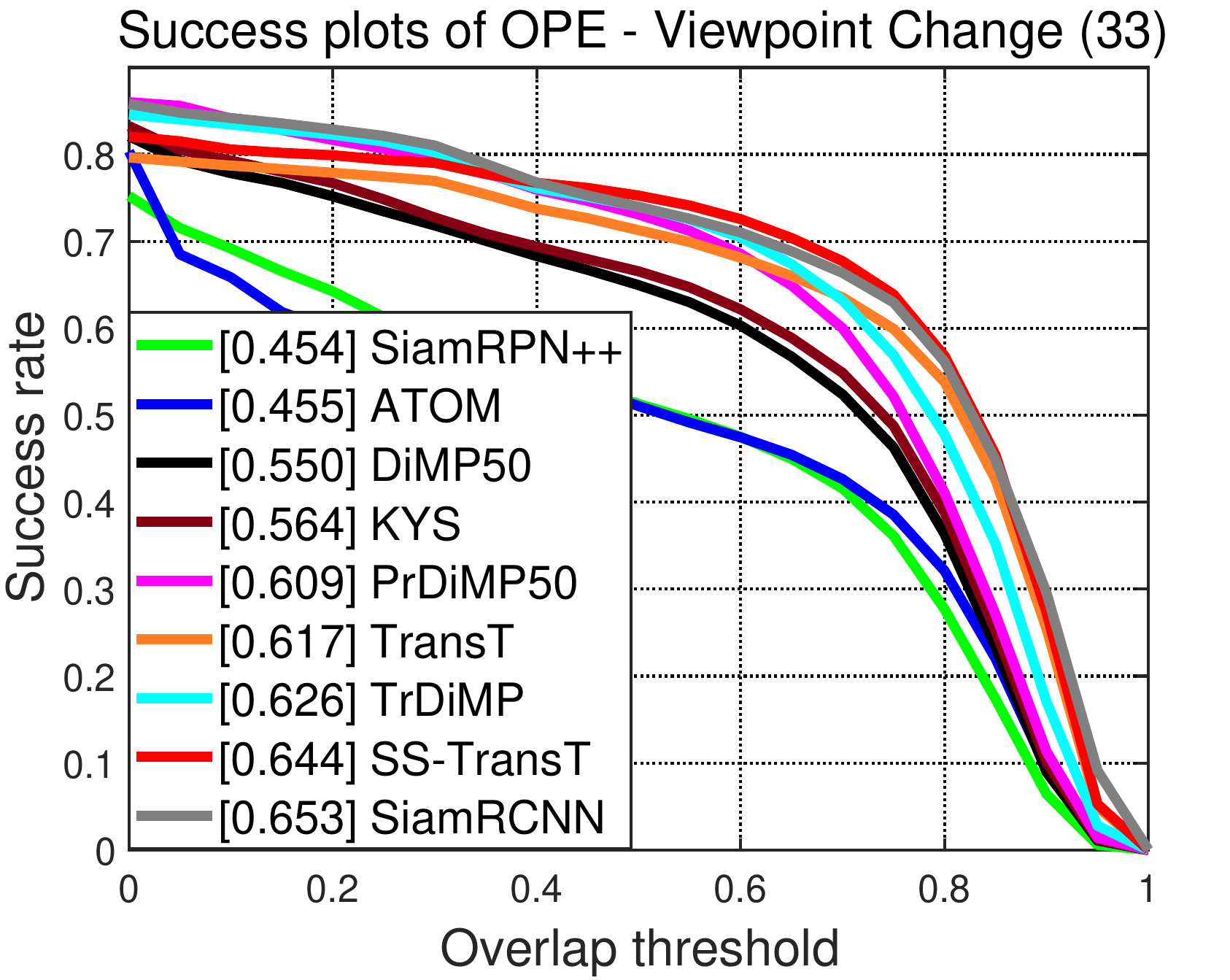}
			\includegraphics[width=0.24\linewidth]{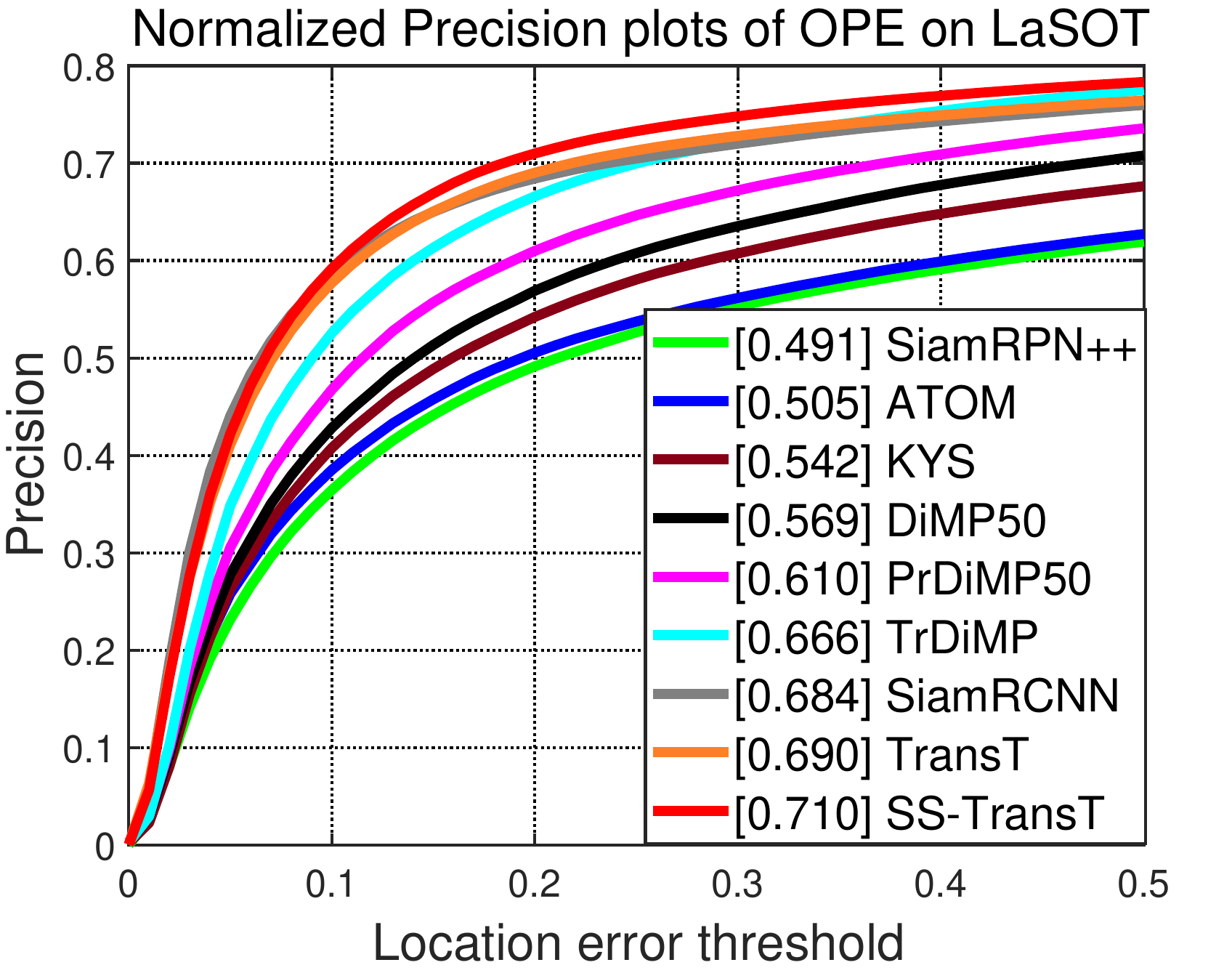}
			\includegraphics[width=0.24\linewidth]{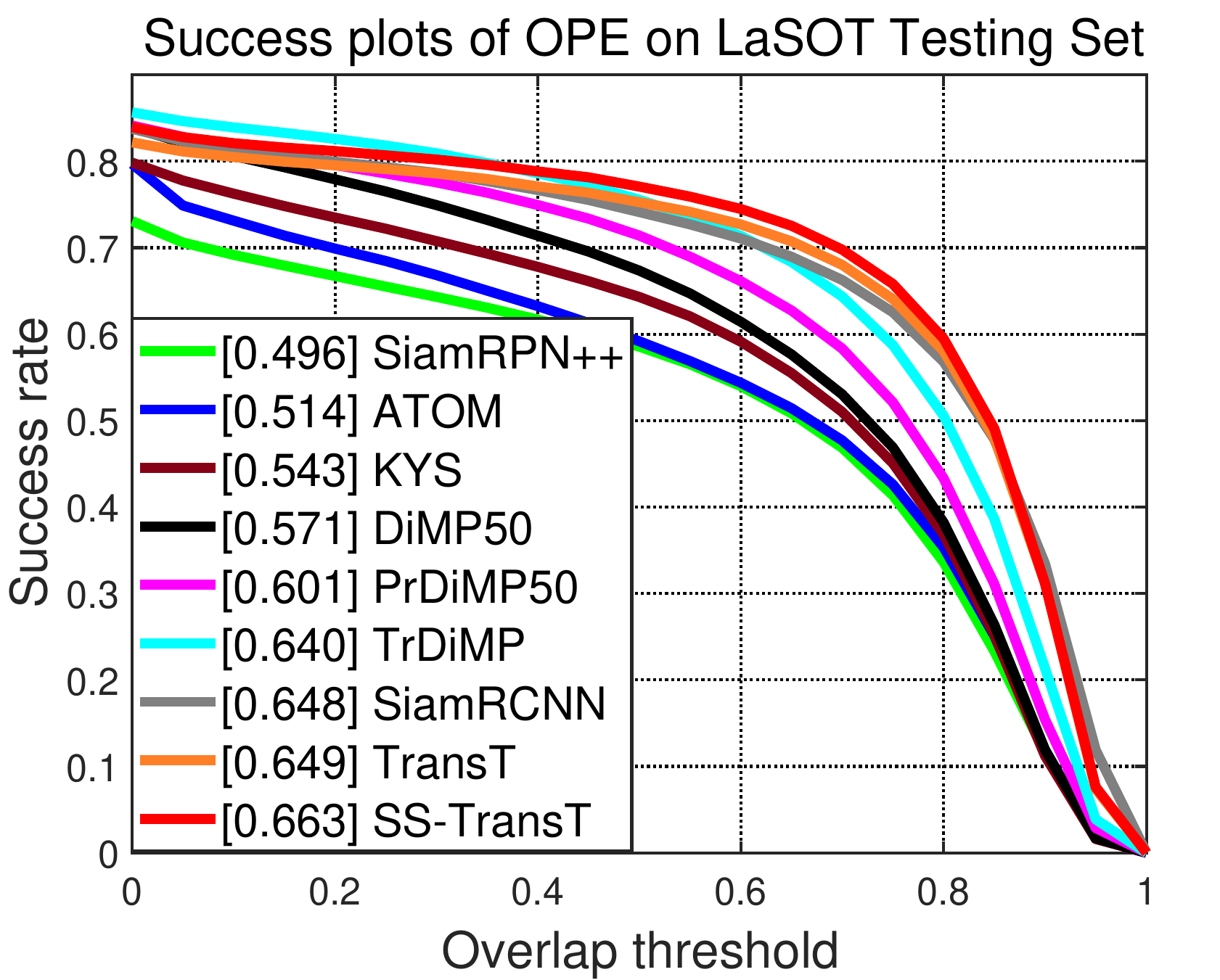}
		\end{minipage}	    

\end{center}
	\caption {\textbf{Success plots on the LaSOT dataset over each attribute.} The figure shows that by using the proposed crop-transform-paste, the proposed SS-TransT algorithm achieves favorable performance against other 8 state-of-the-art methods on all the 14 attributes. }
	\label{fig:plots-attri}
\end{figure*}

\begin{figure*}[!htb]
\begin{center}

		\begin{minipage}[c]{0.9\linewidth}
            \includegraphics[width=0.193\linewidth]{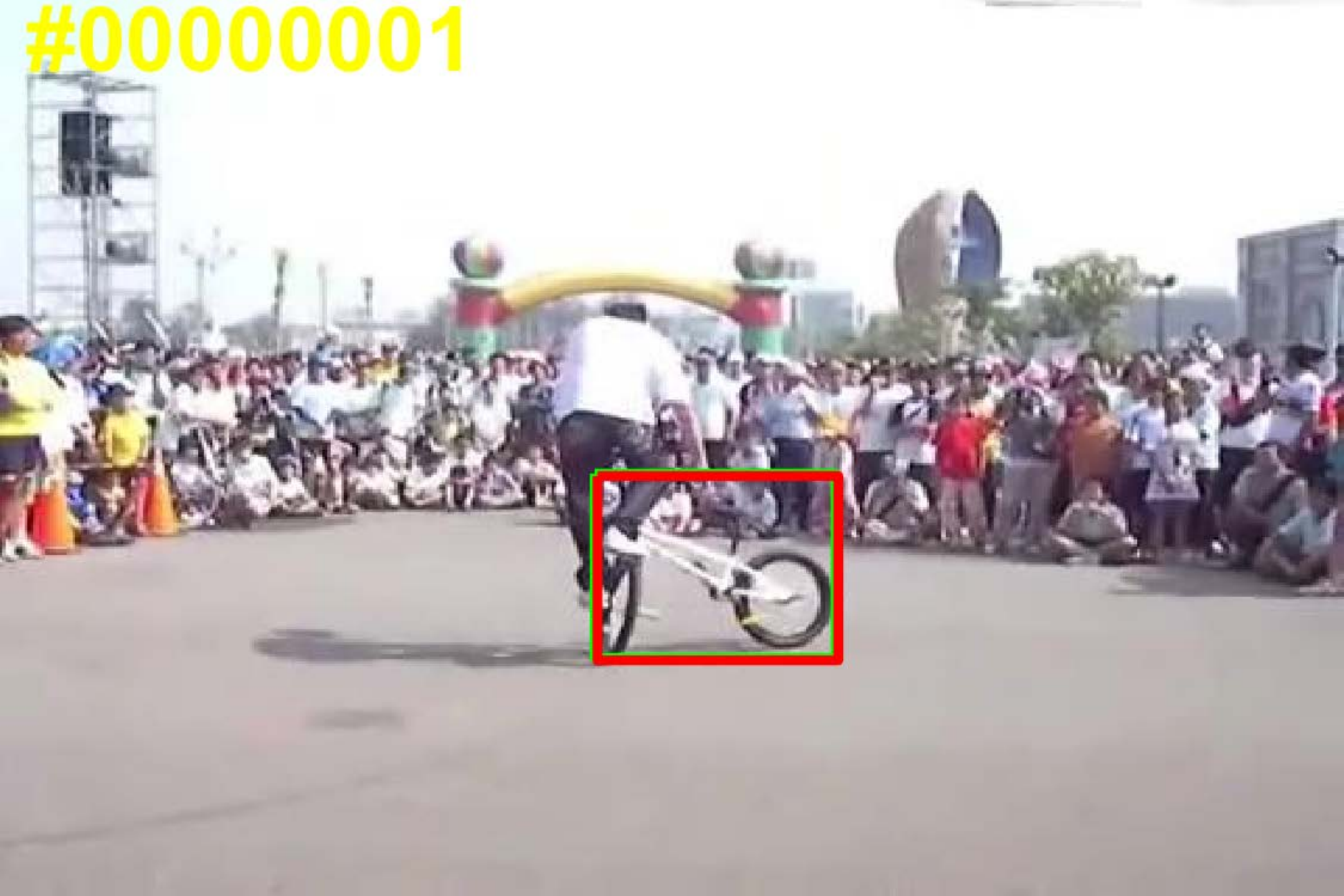}\vspace{1.5mm}
			\includegraphics[width=0.193\linewidth]{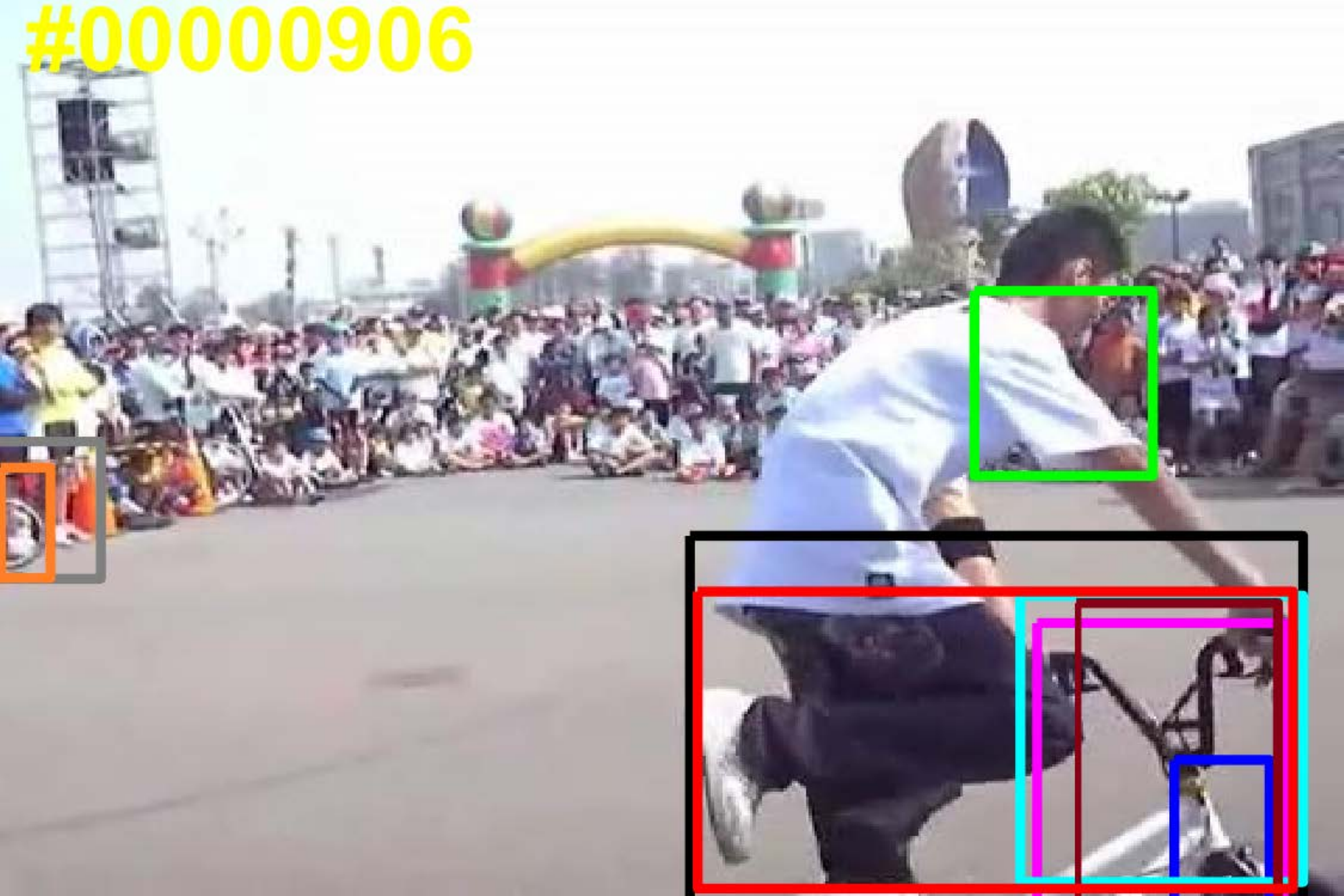}
			\includegraphics[width=0.193\linewidth]{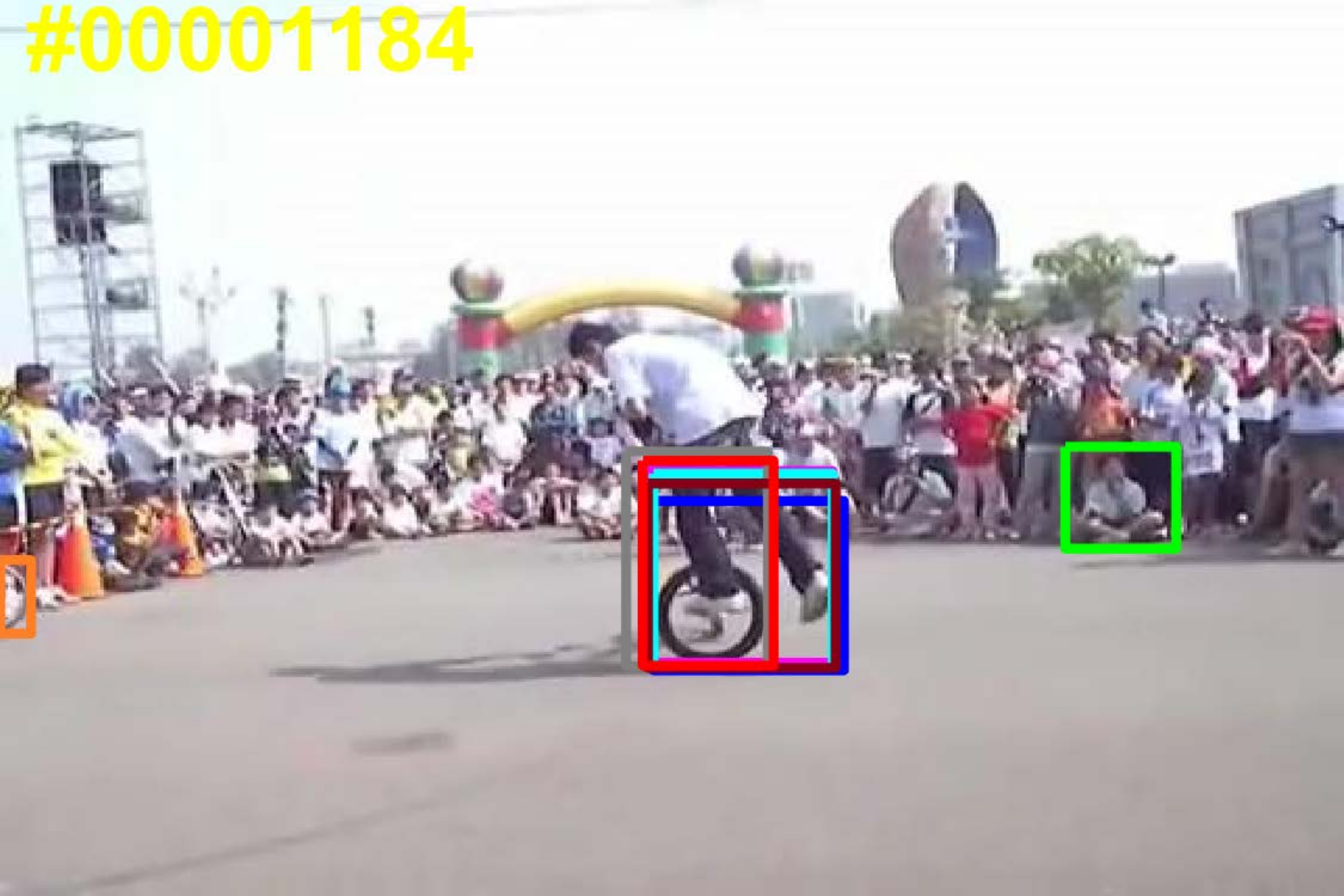}
			\includegraphics[width=0.193\linewidth]{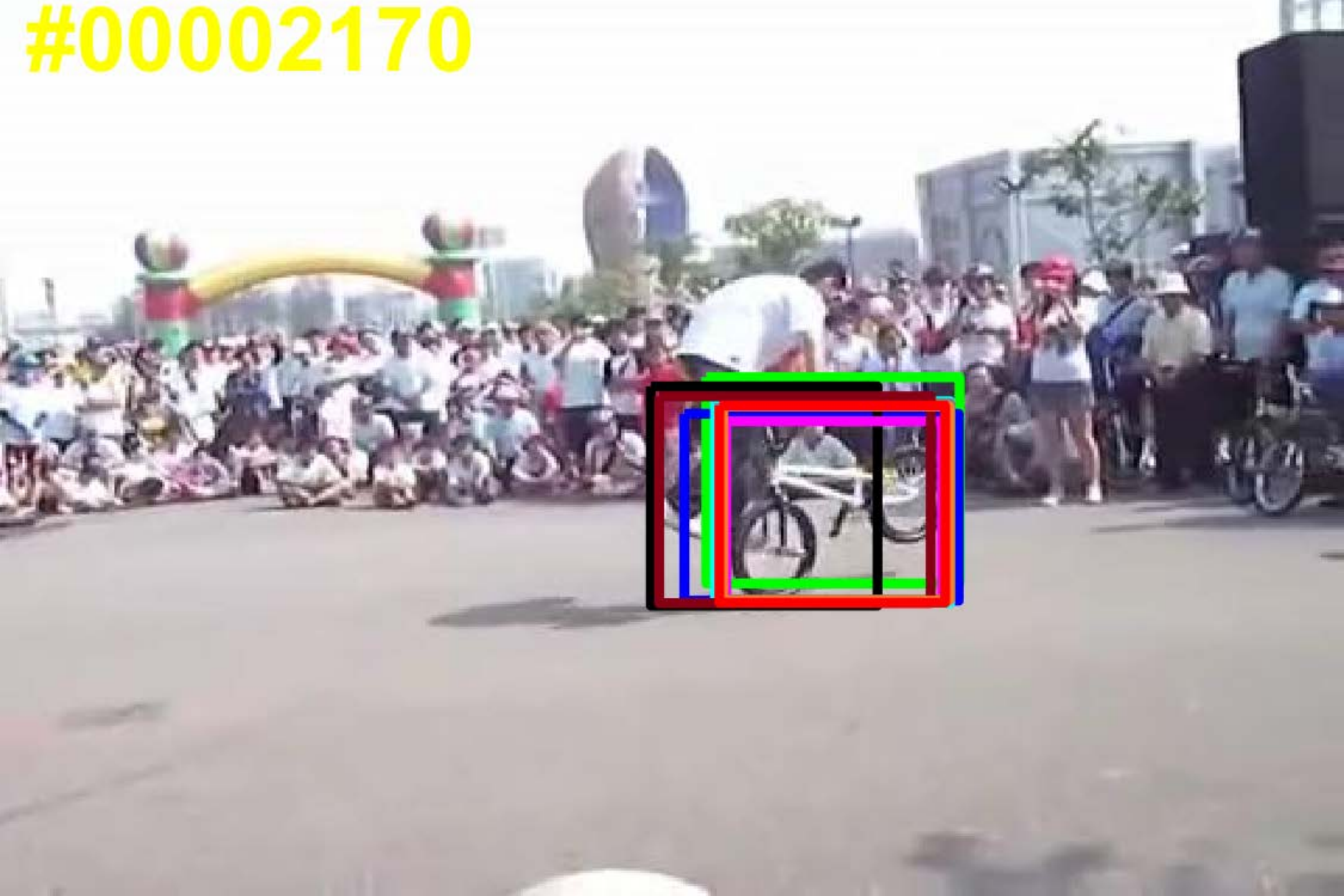}
			\includegraphics[width=0.193\linewidth]{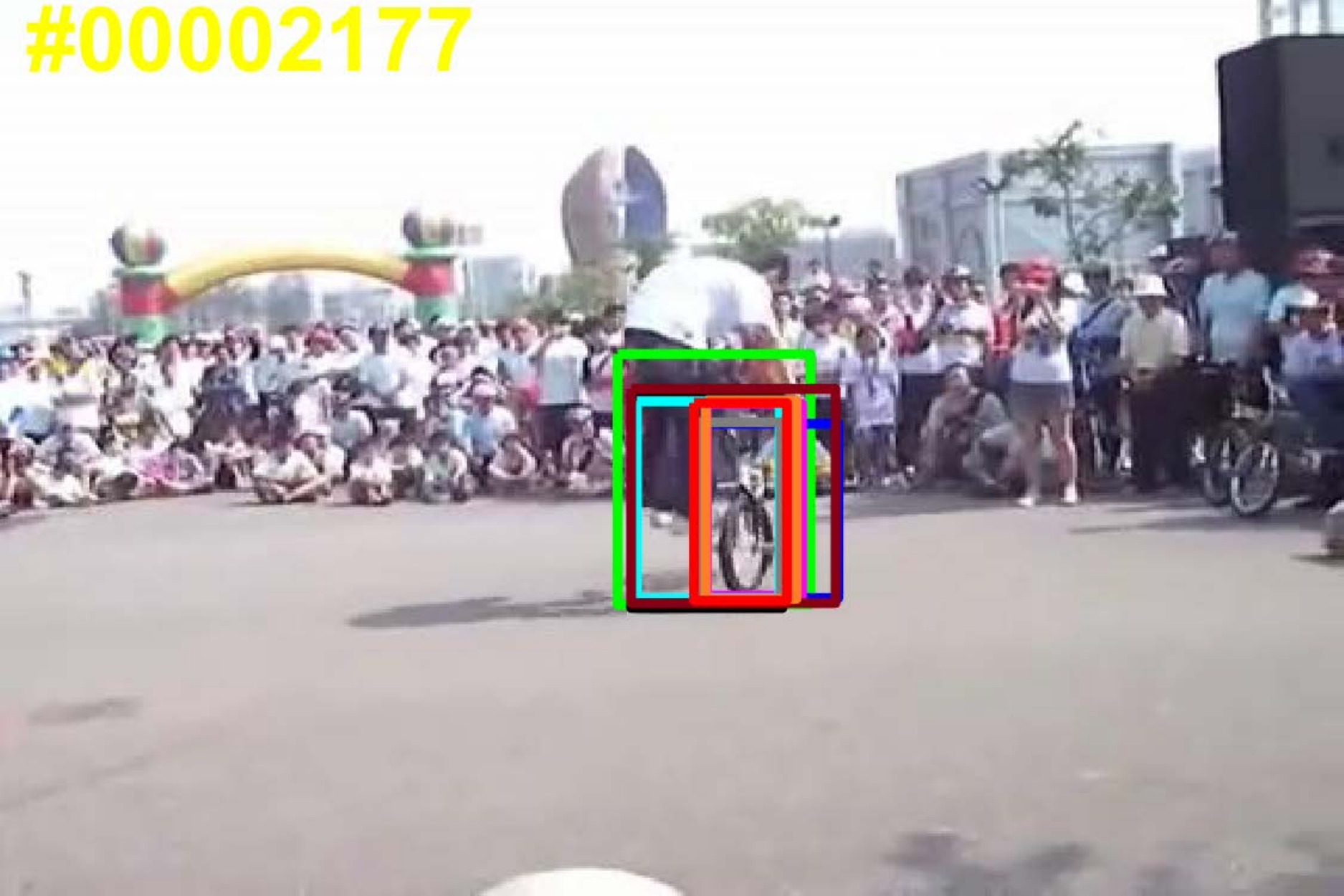}
		\end{minipage}
	    \begin{minipage}[c]{0.9\linewidth}
            \includegraphics[width=0.193\linewidth]{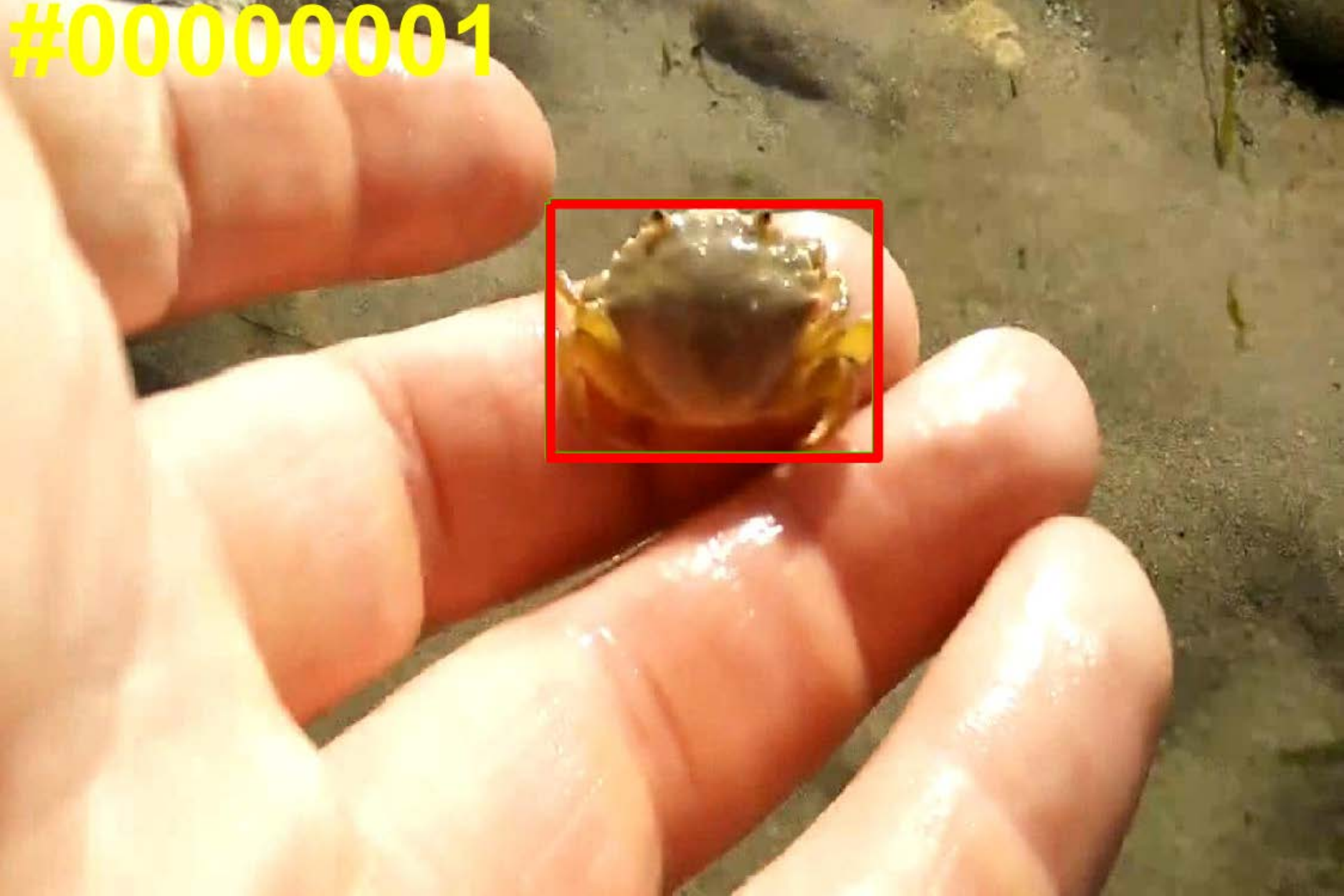}\vspace{1.5mm}
			\includegraphics[width=0.193\linewidth]{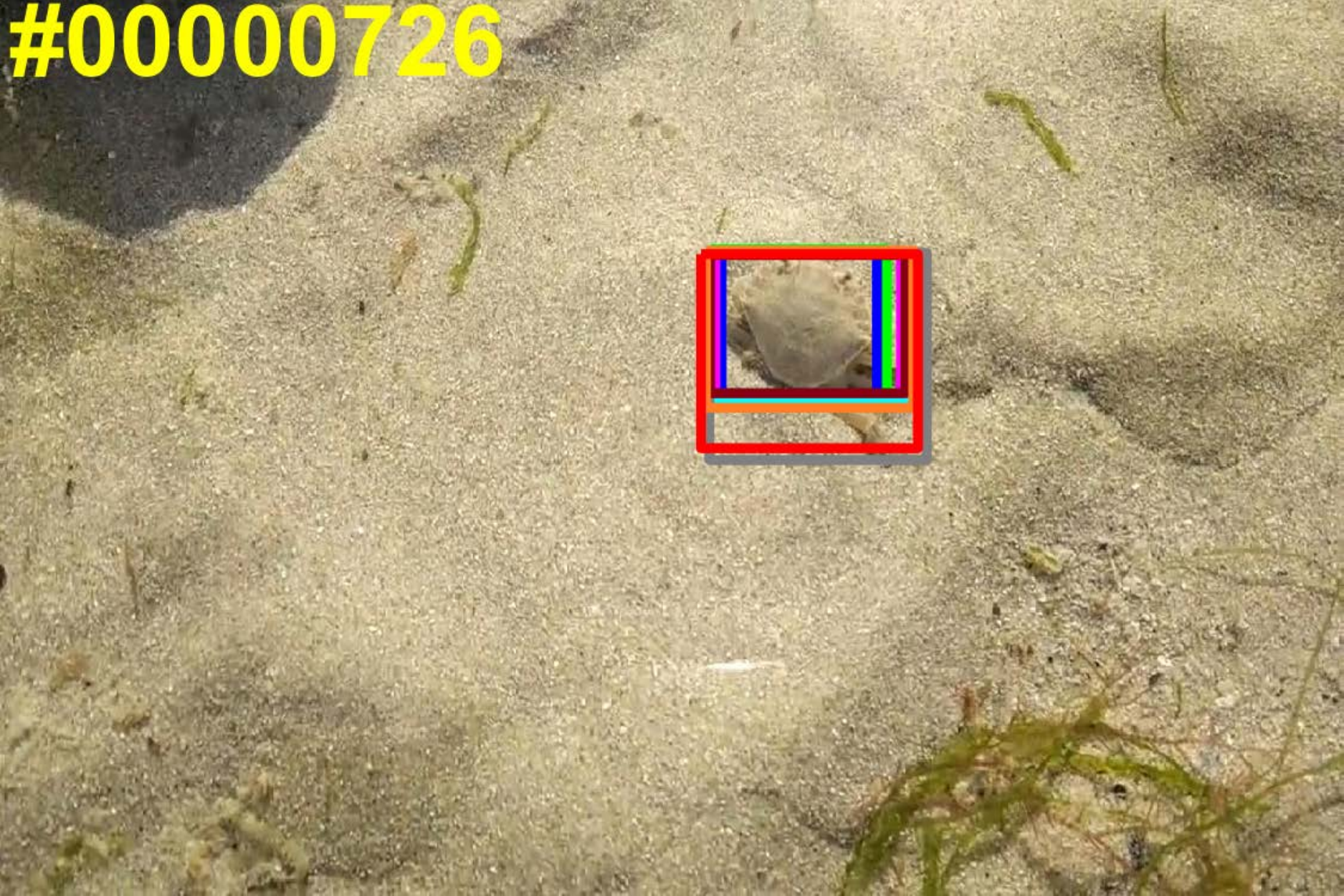}
			\includegraphics[width=0.193\linewidth]{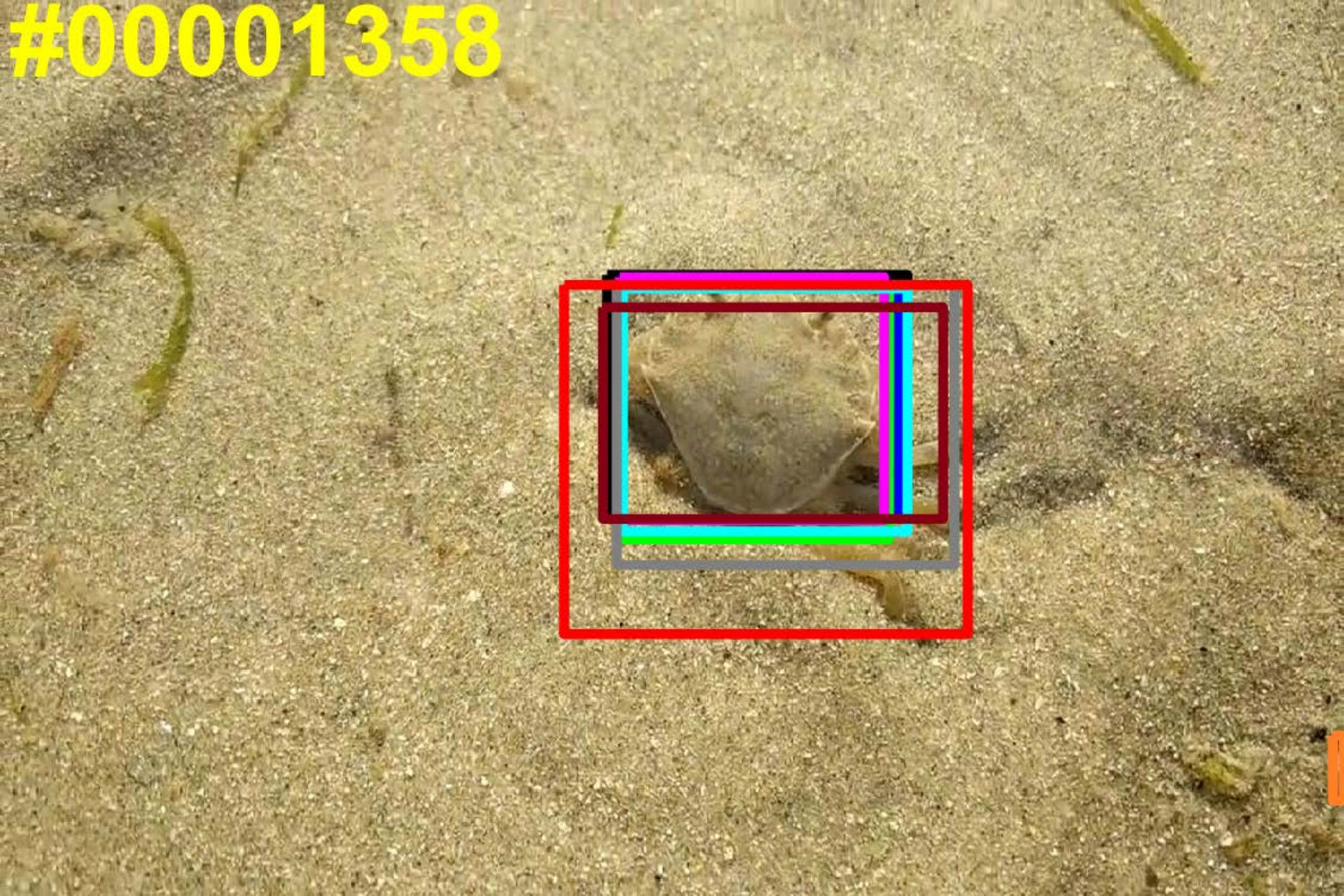}
			\includegraphics[width=0.193\linewidth]{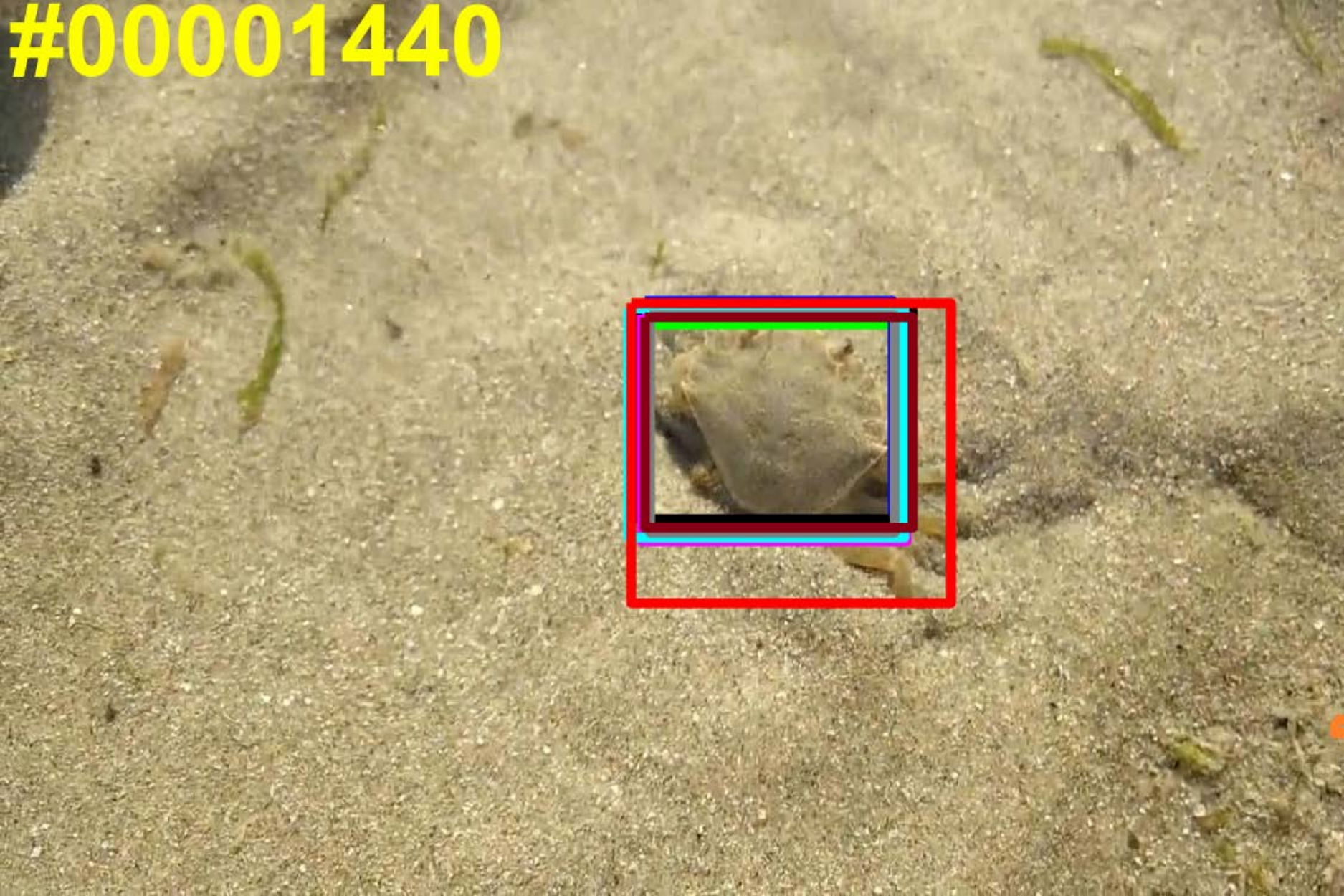}
			\includegraphics[width=0.193\linewidth]{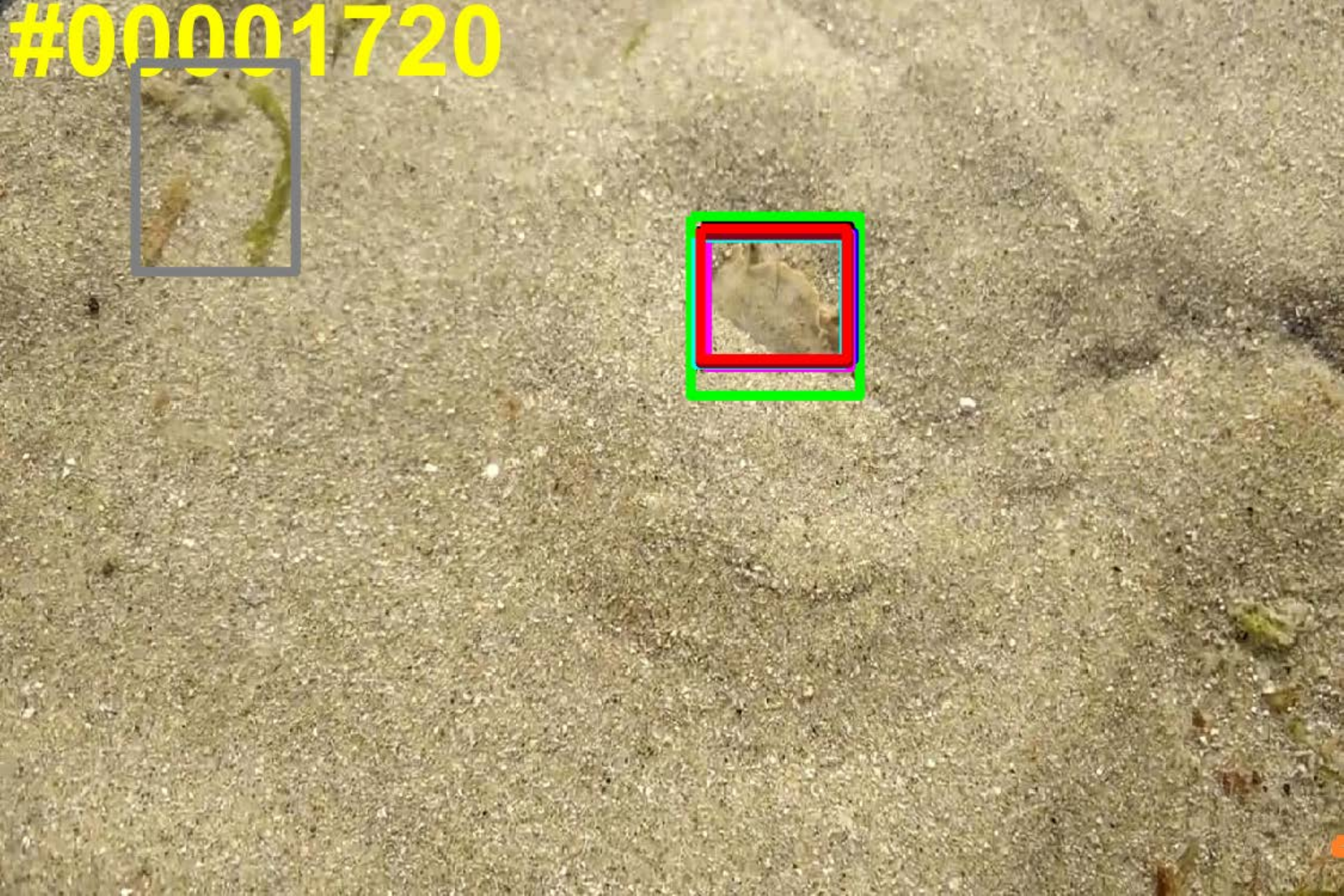}
		\end{minipage}
	    \begin{minipage}[c]{0.9\linewidth}
            \includegraphics[width=0.193\linewidth]{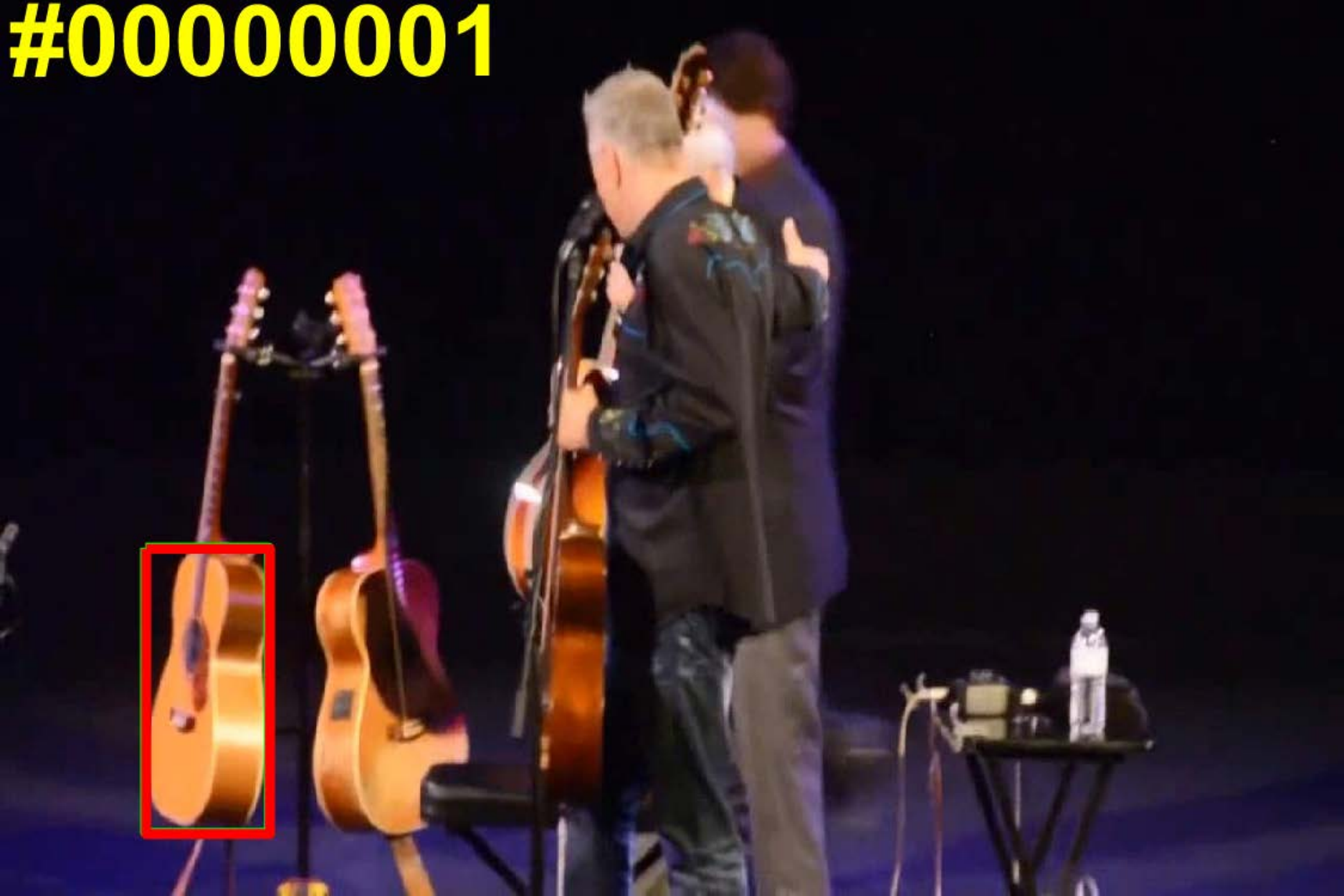}\vspace{1.5mm}
            \includegraphics[width=0.193\linewidth]{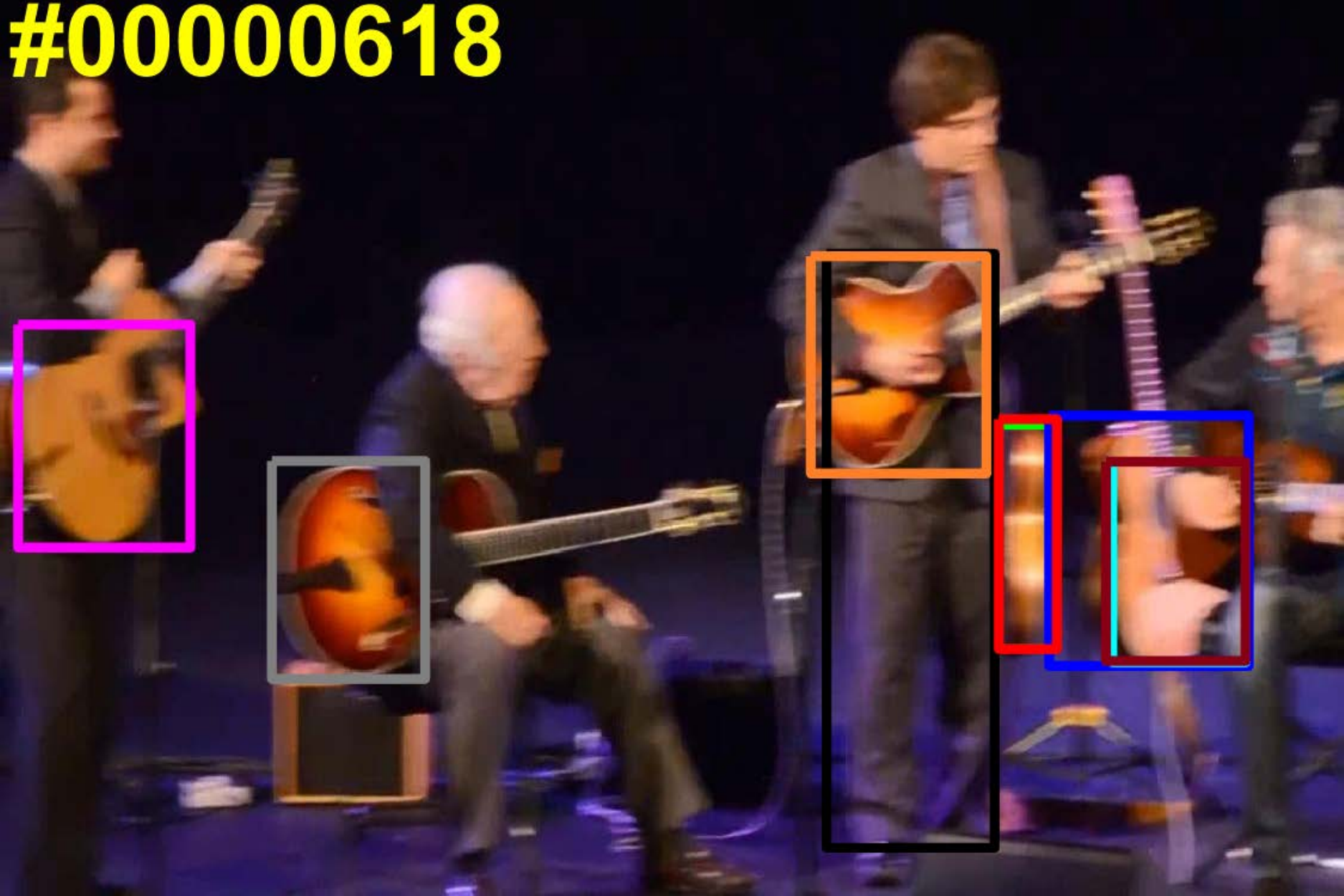}
            \includegraphics[width=0.193\linewidth]{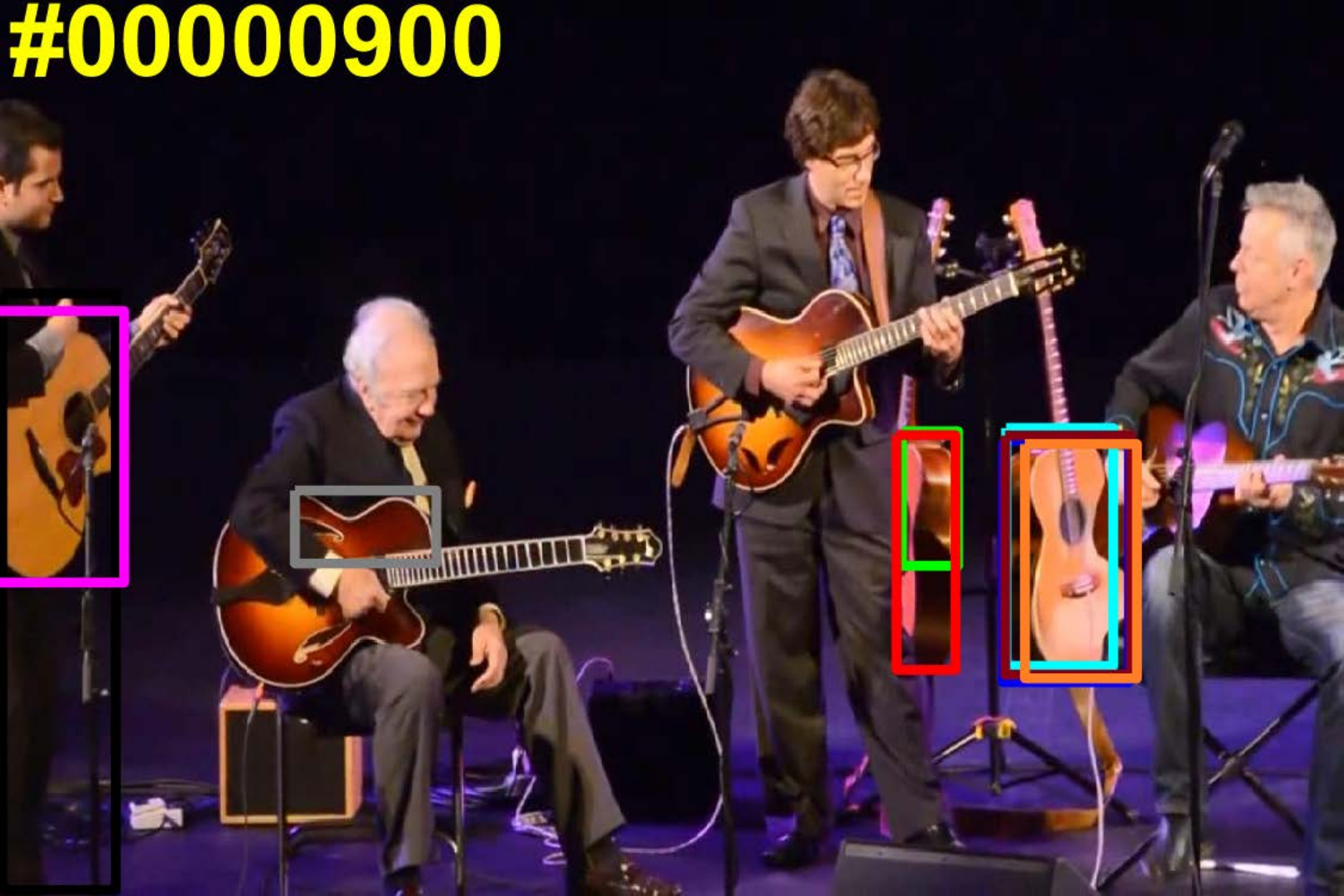}
            \includegraphics[width=0.193\linewidth]{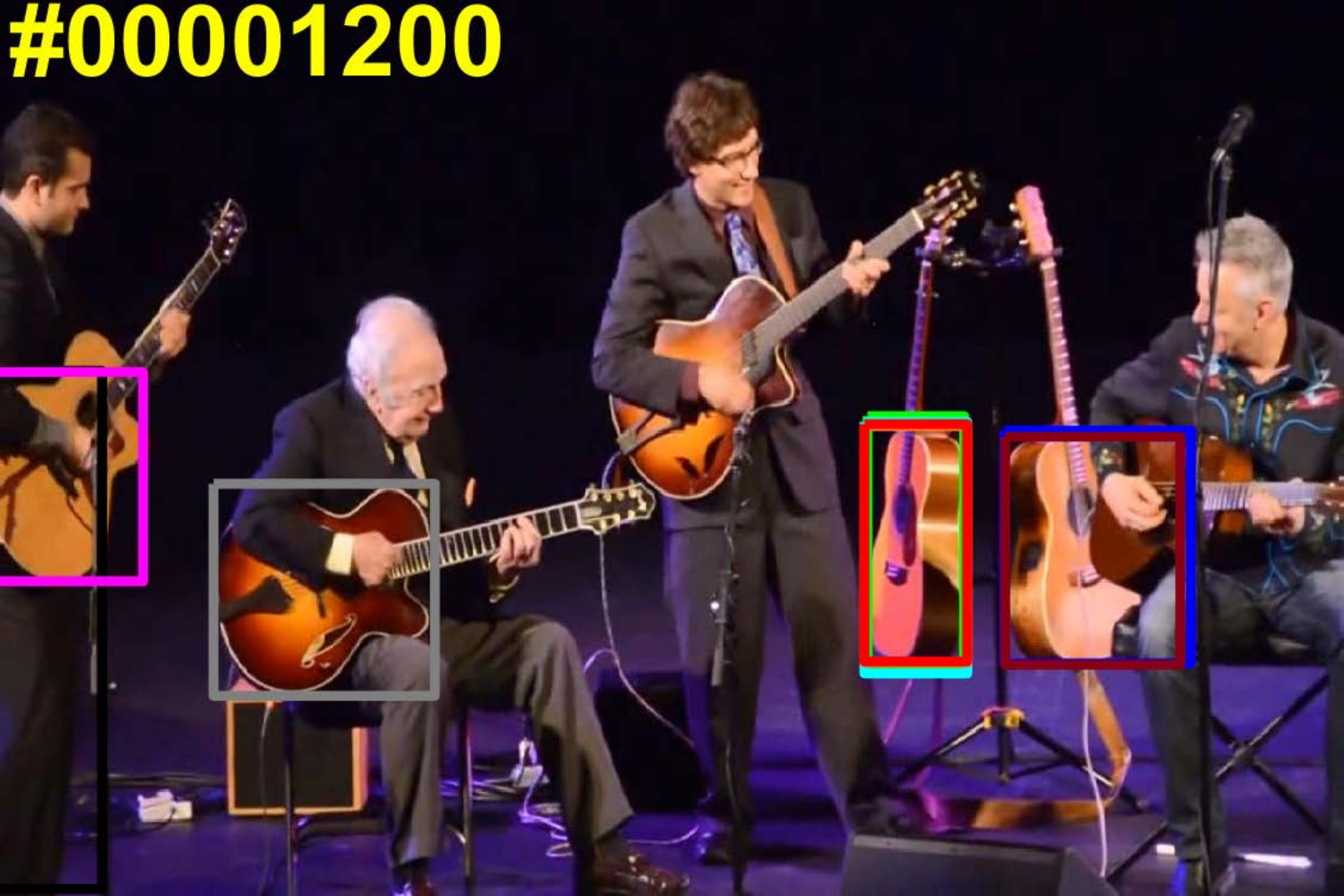}
            \includegraphics[width=0.193\linewidth]{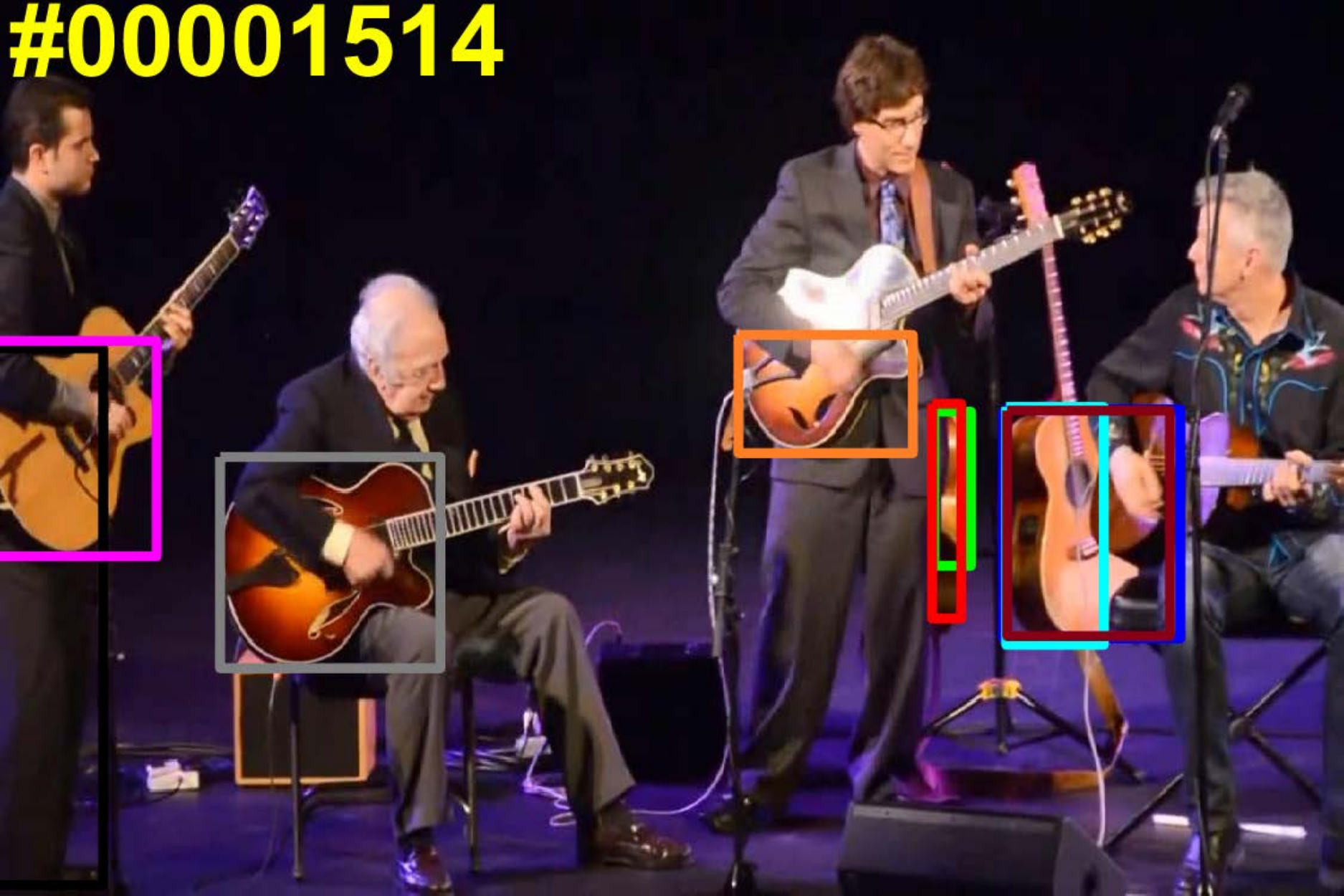}
		\end{minipage}
	    \begin{minipage}[c]{0.9\linewidth}
            \includegraphics[width=0.193\linewidth]{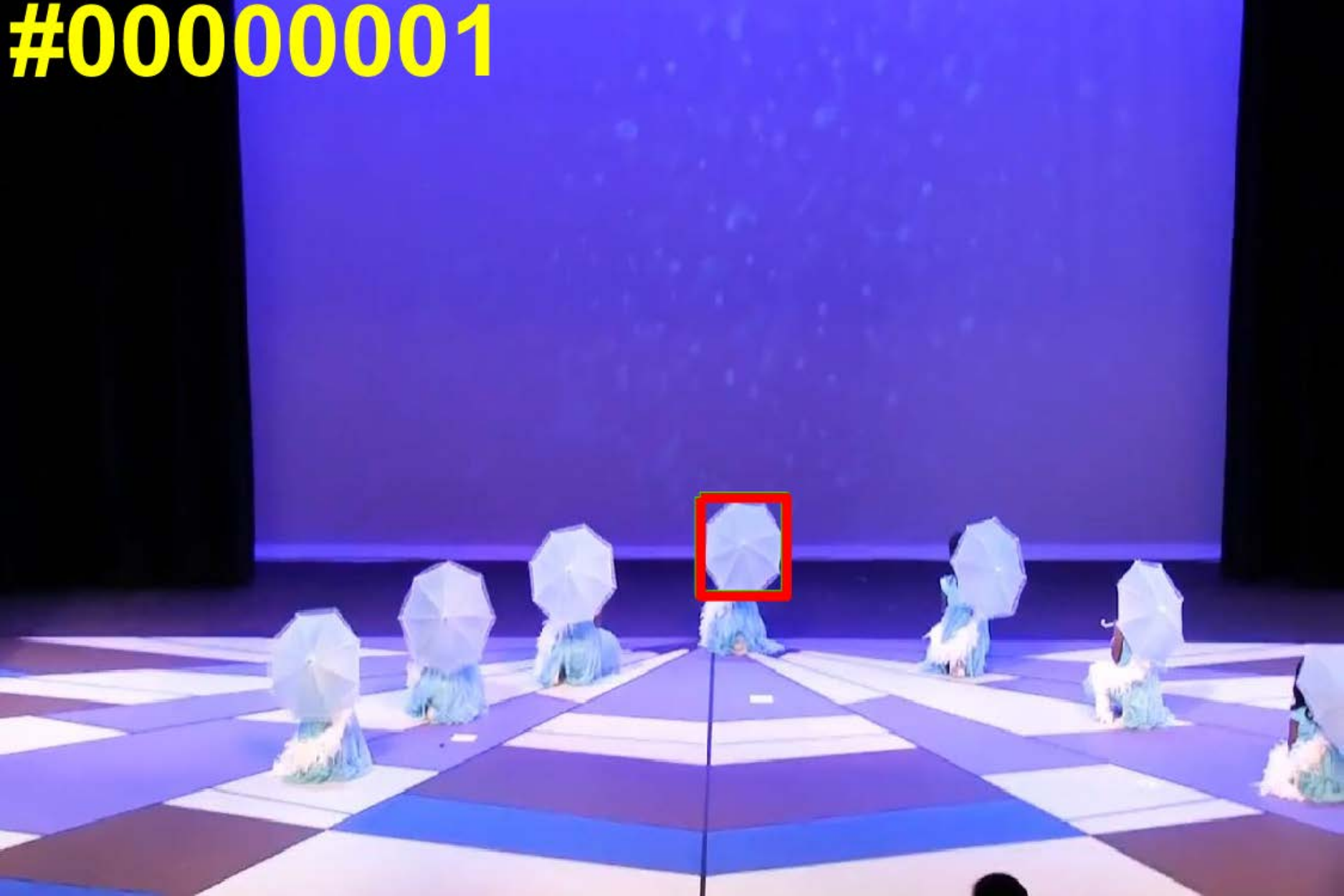}\vspace{1.5mm}
			\includegraphics[width=0.193\linewidth]{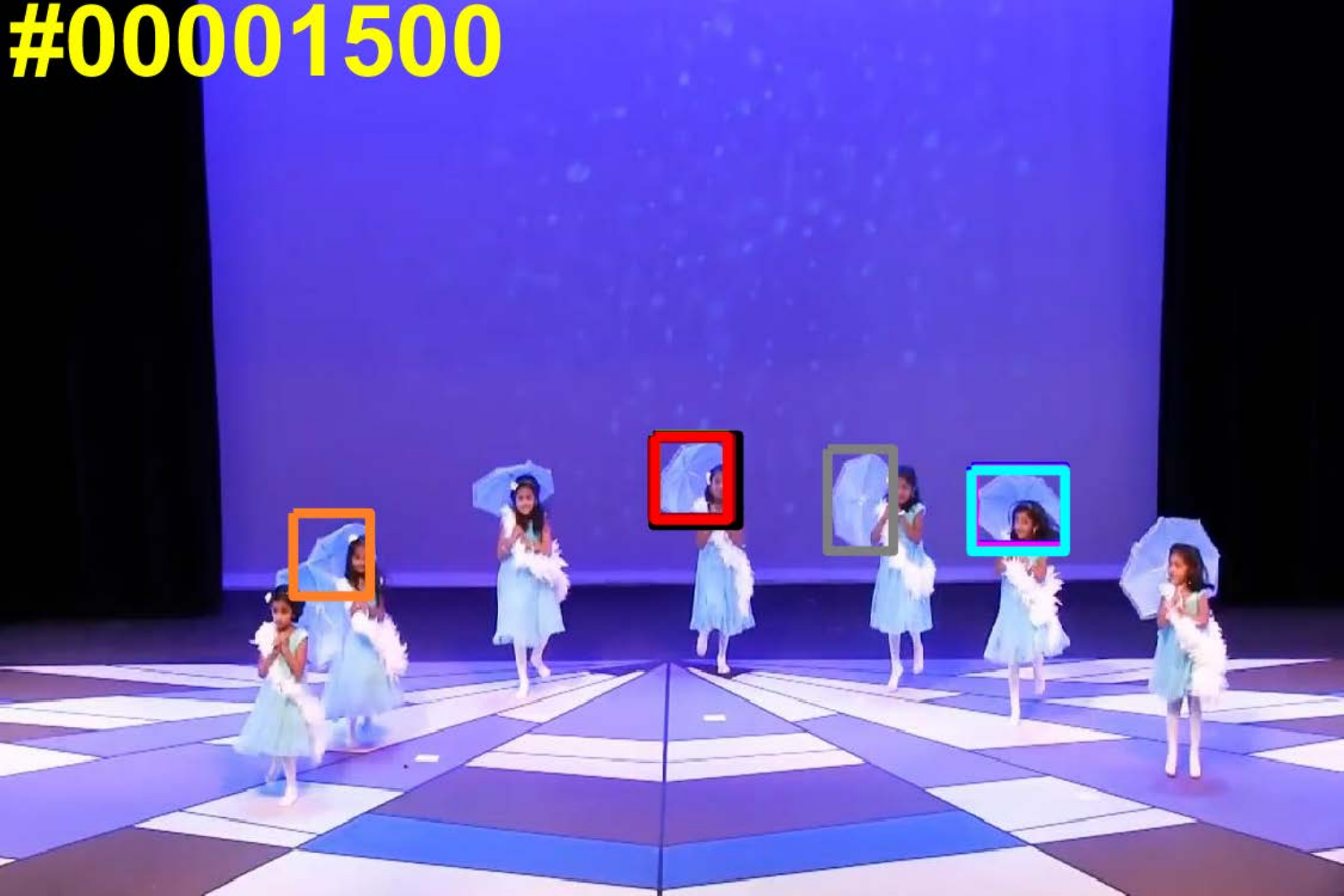}
            \includegraphics[width=0.193\linewidth]{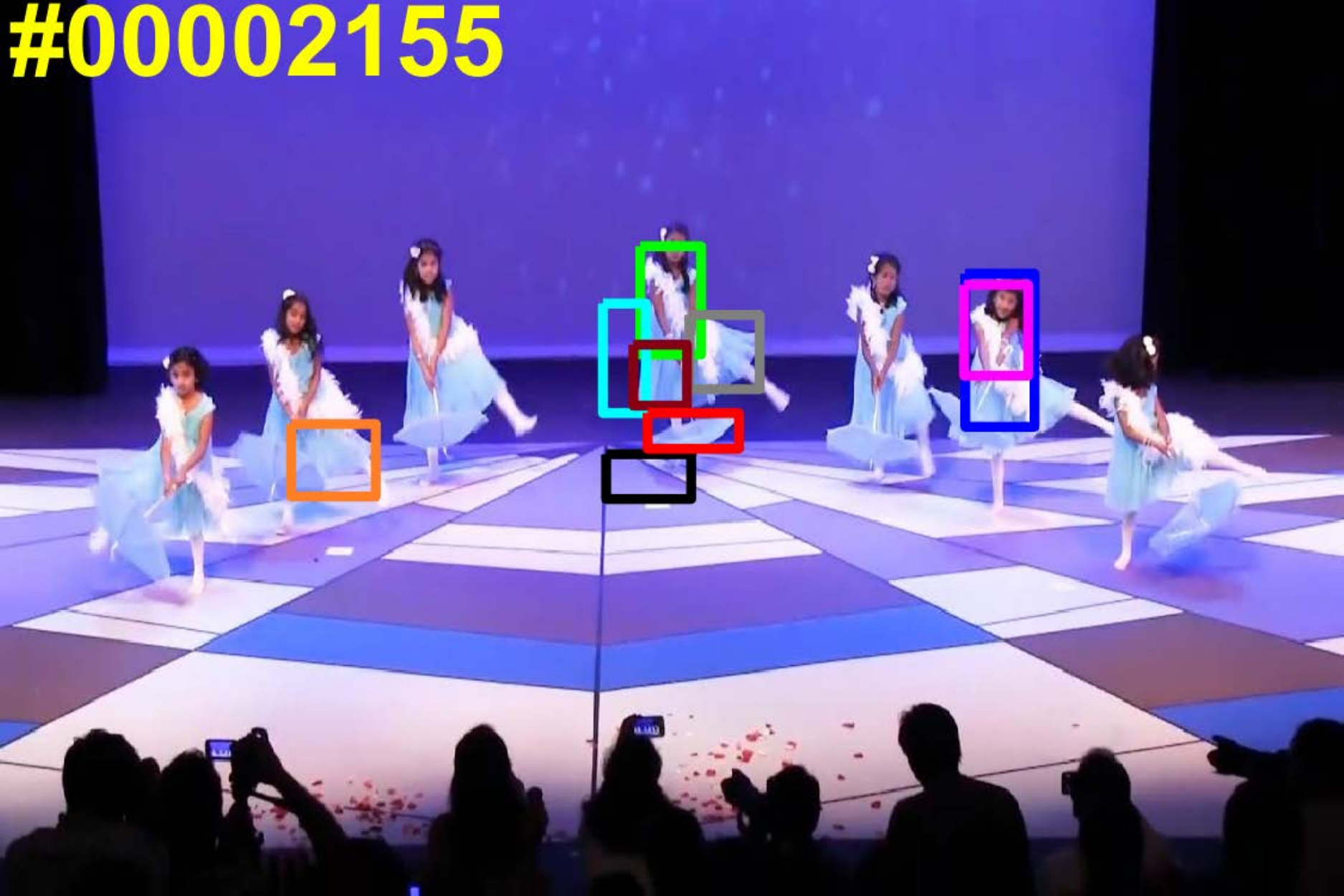}
            \includegraphics[width=0.193\linewidth]{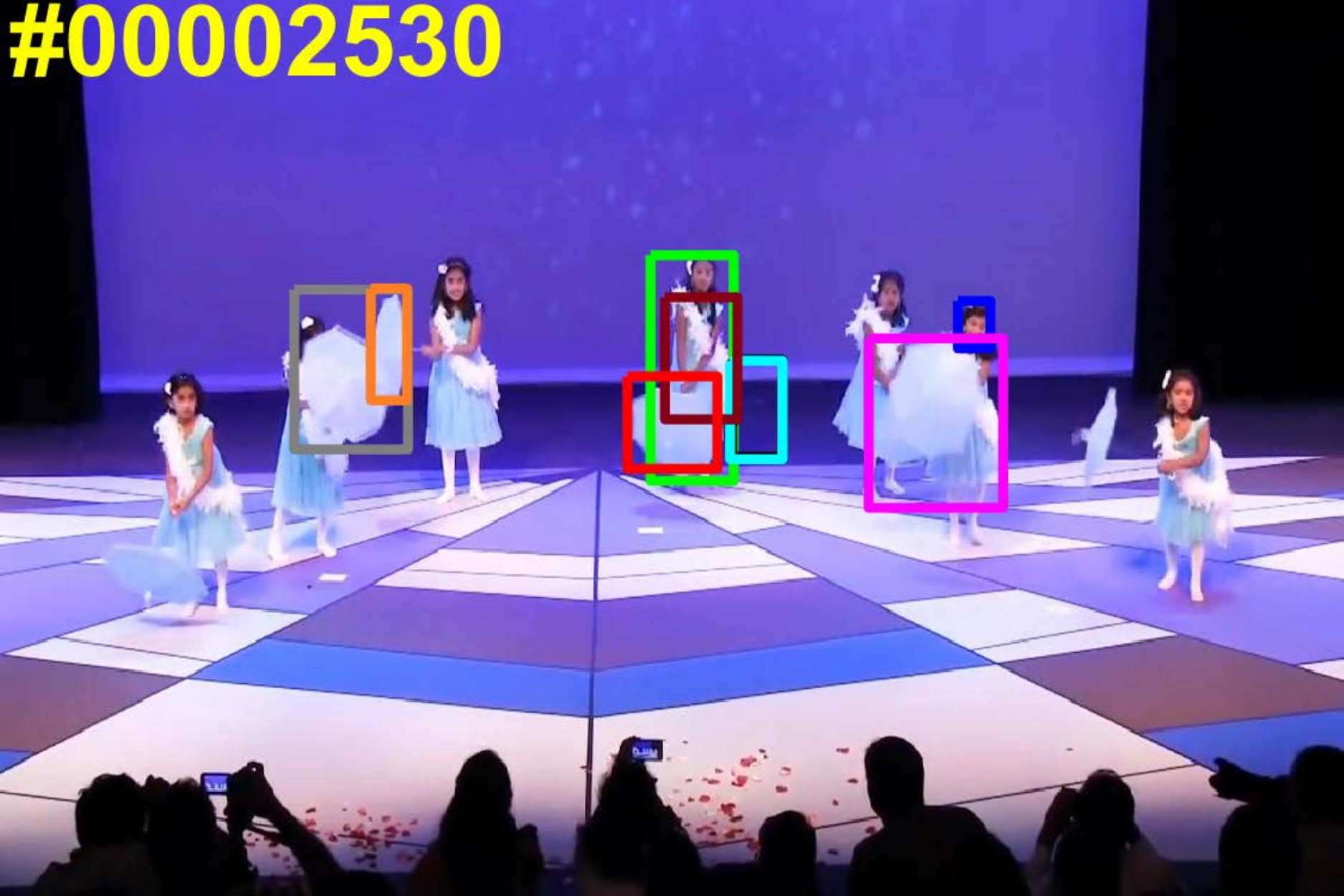}
            \includegraphics[width=0.193\linewidth]{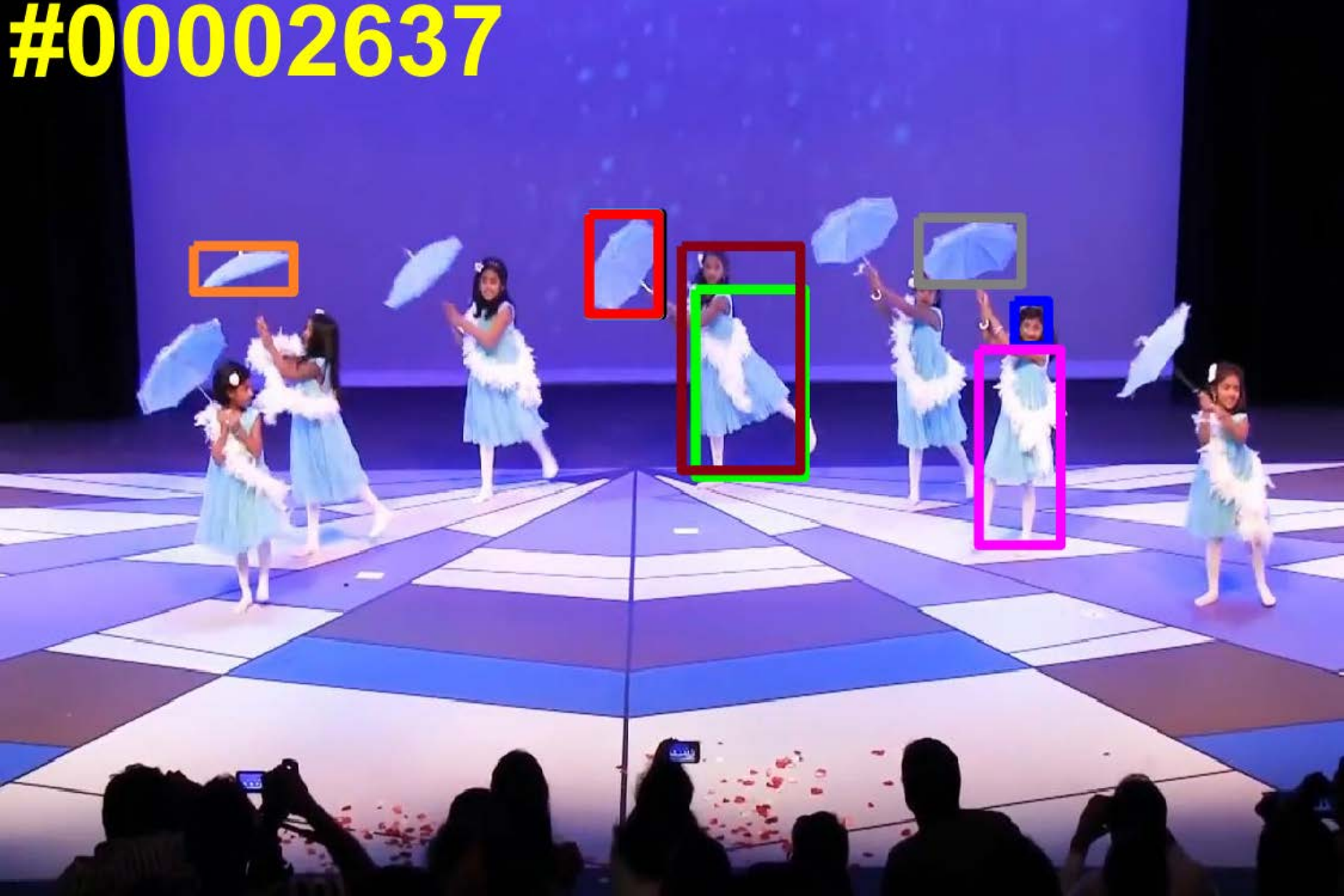}
		\end{minipage}
	    \begin{minipage}[c]{0.9\linewidth}
            \includegraphics[width=0.193\linewidth]{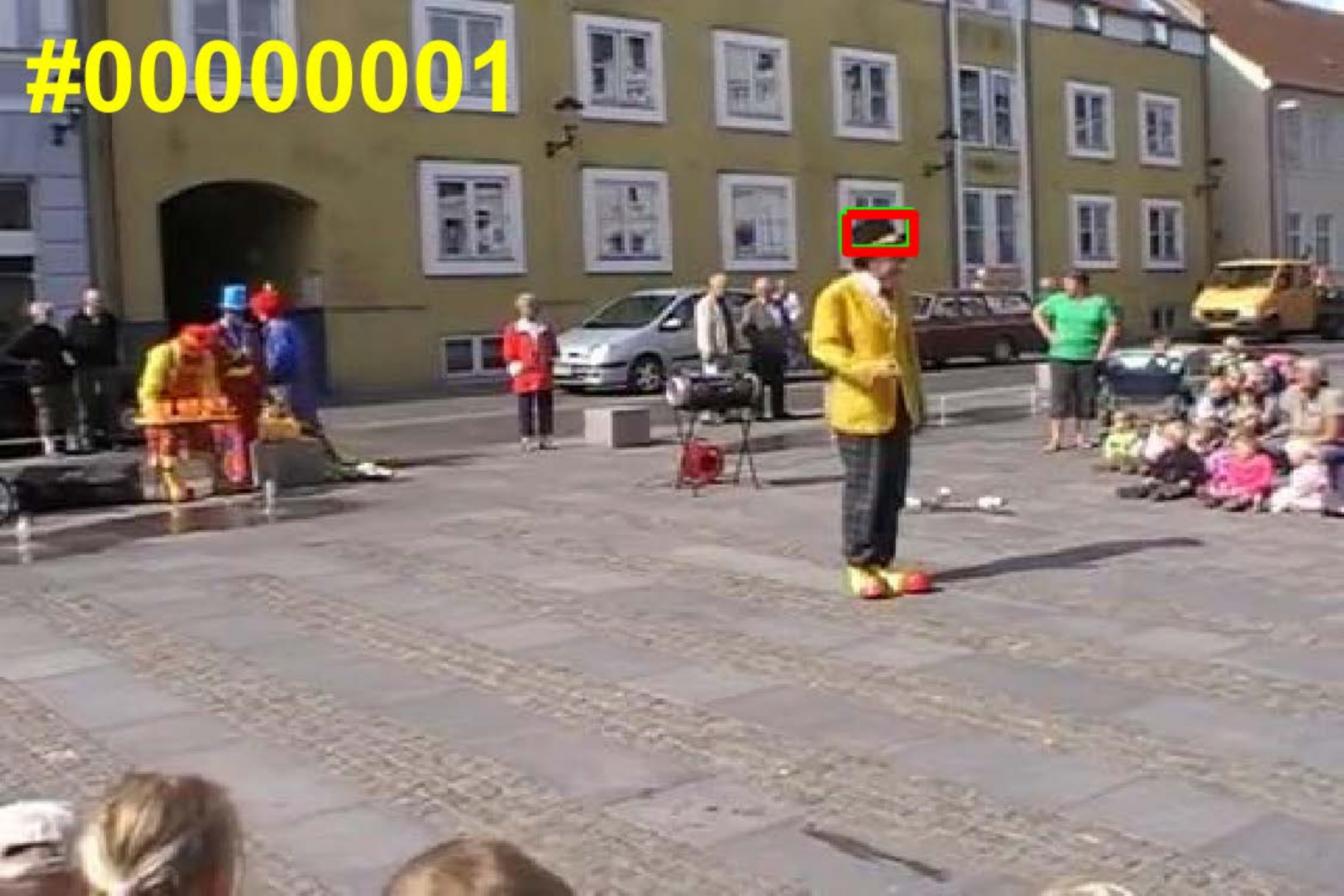}\vspace{1.5mm}
			\includegraphics[width=0.193\linewidth]{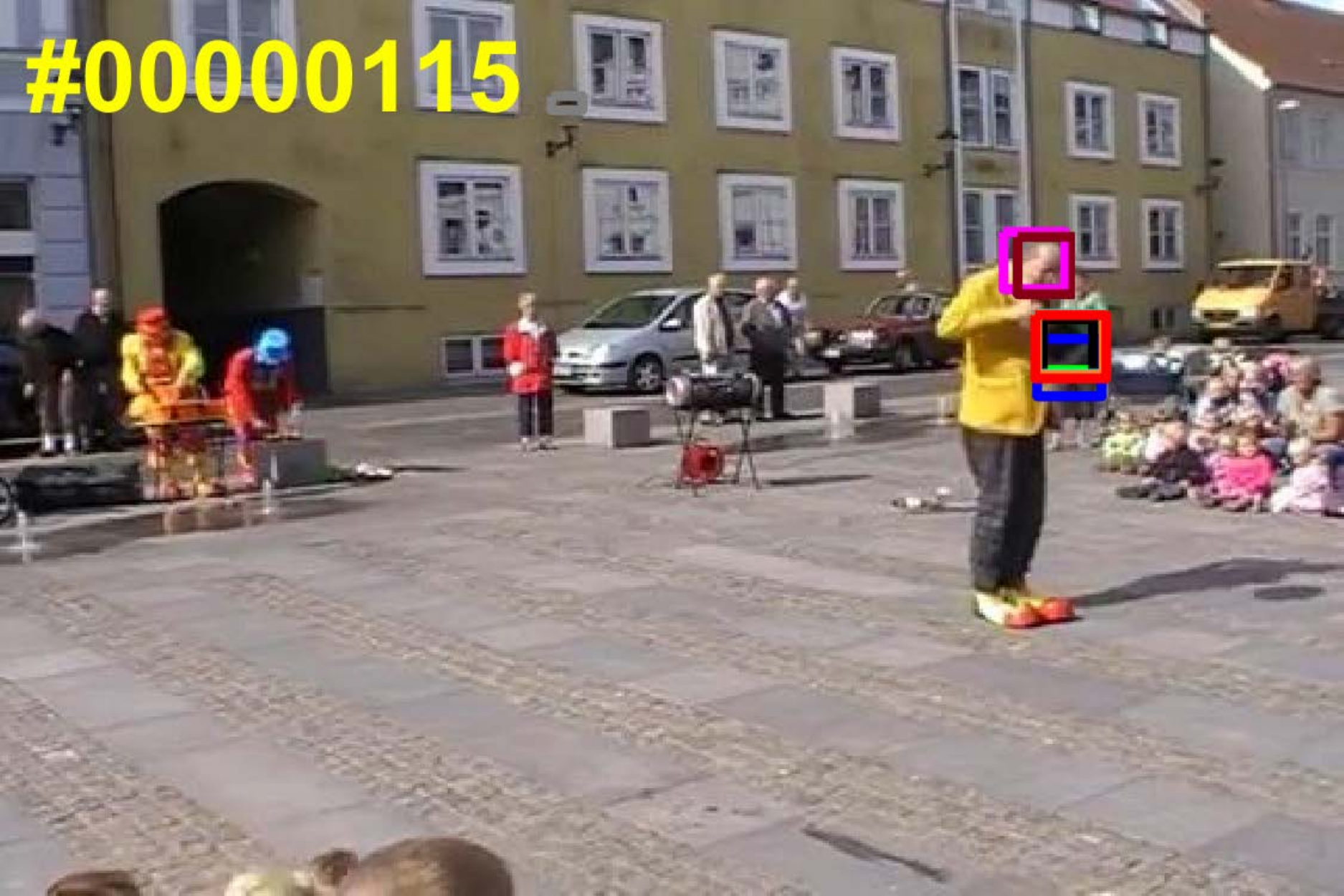}
            \includegraphics[width=0.193\linewidth]{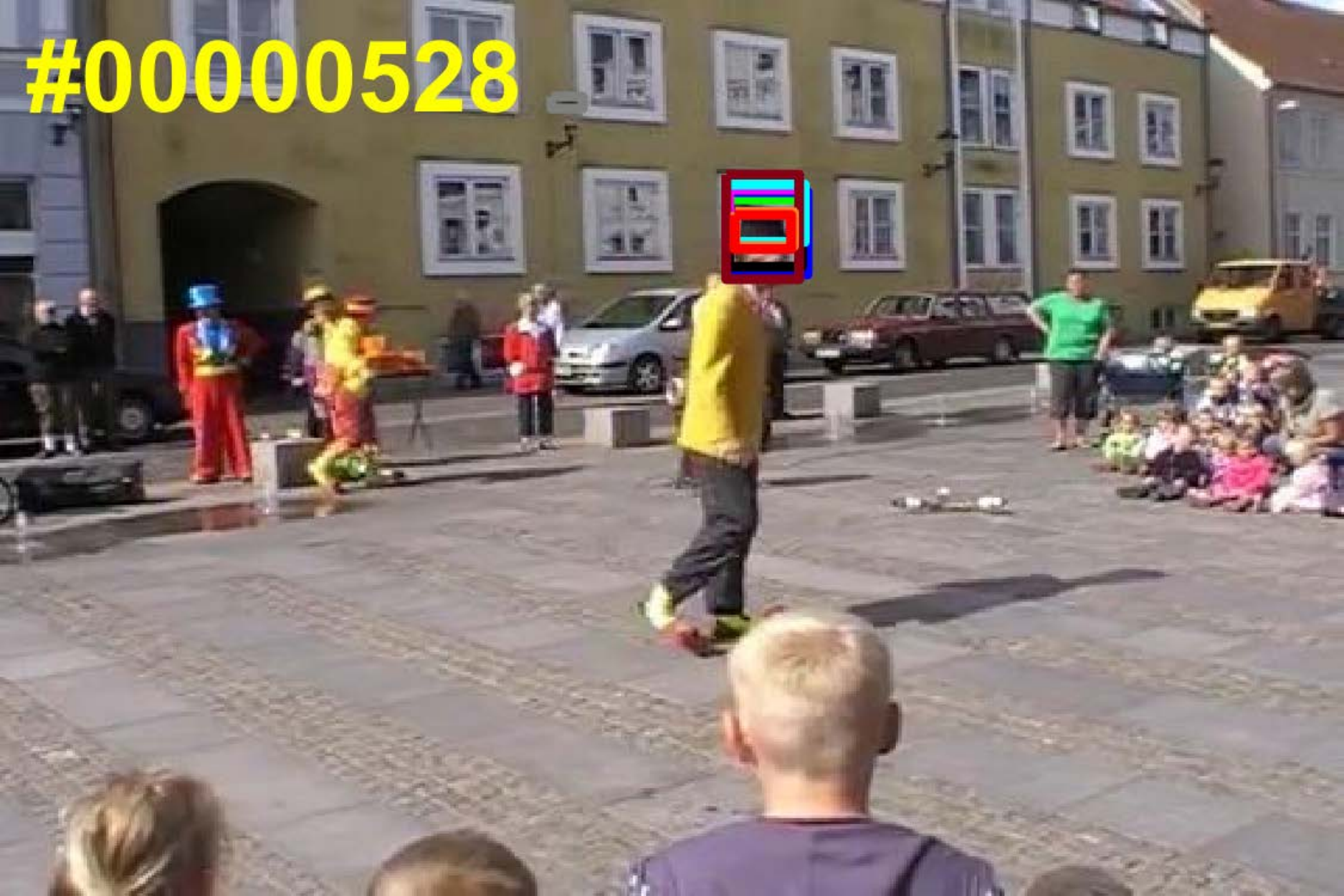}
            \includegraphics[width=0.193\linewidth]{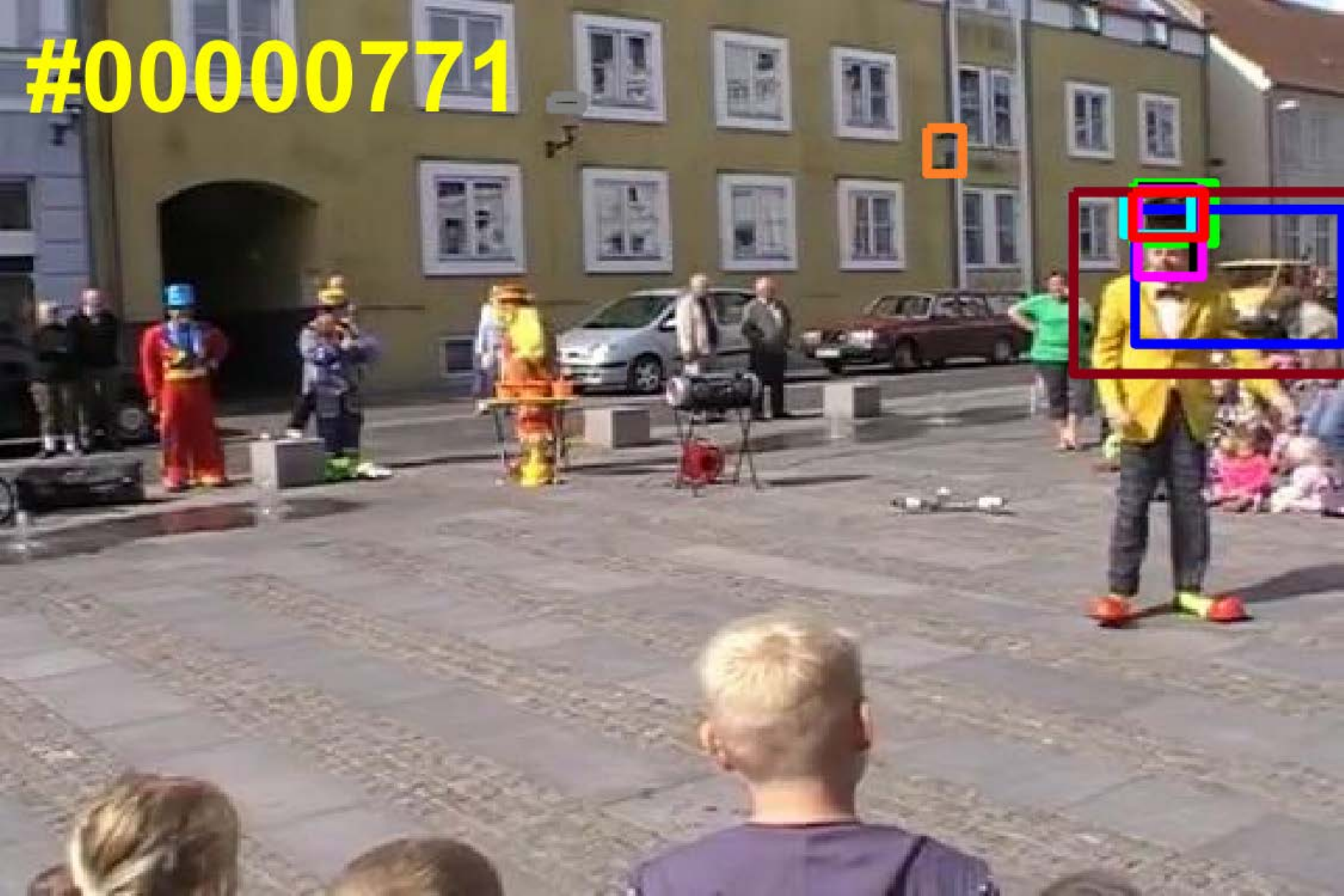}
            \includegraphics[width=0.193\linewidth]{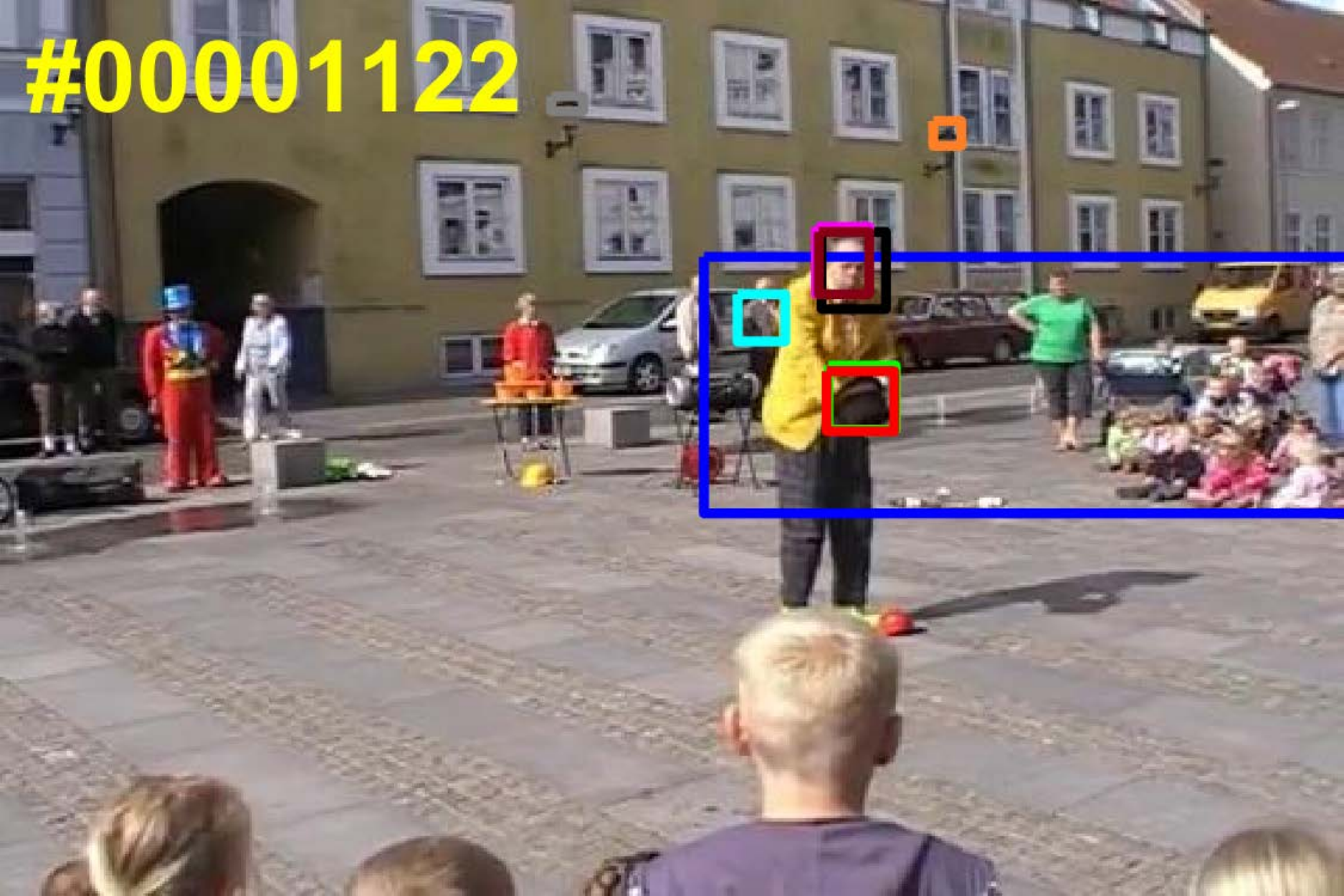}
		\end{minipage}
	    \begin{minipage}[c]{0.9\linewidth}
            \includegraphics[width=0.193\linewidth]{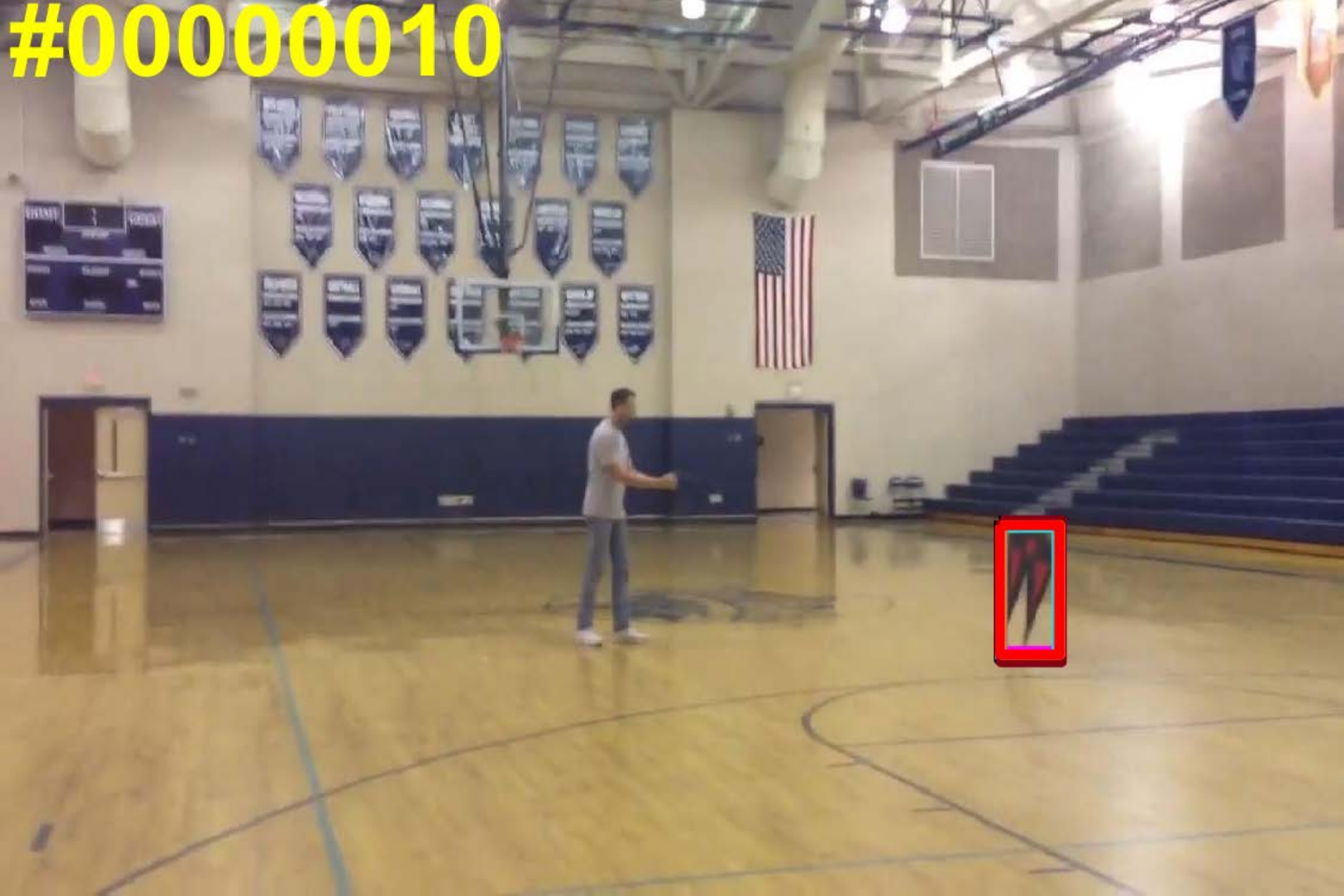}\vspace{1.5mm}
			\includegraphics[width=0.193\linewidth]{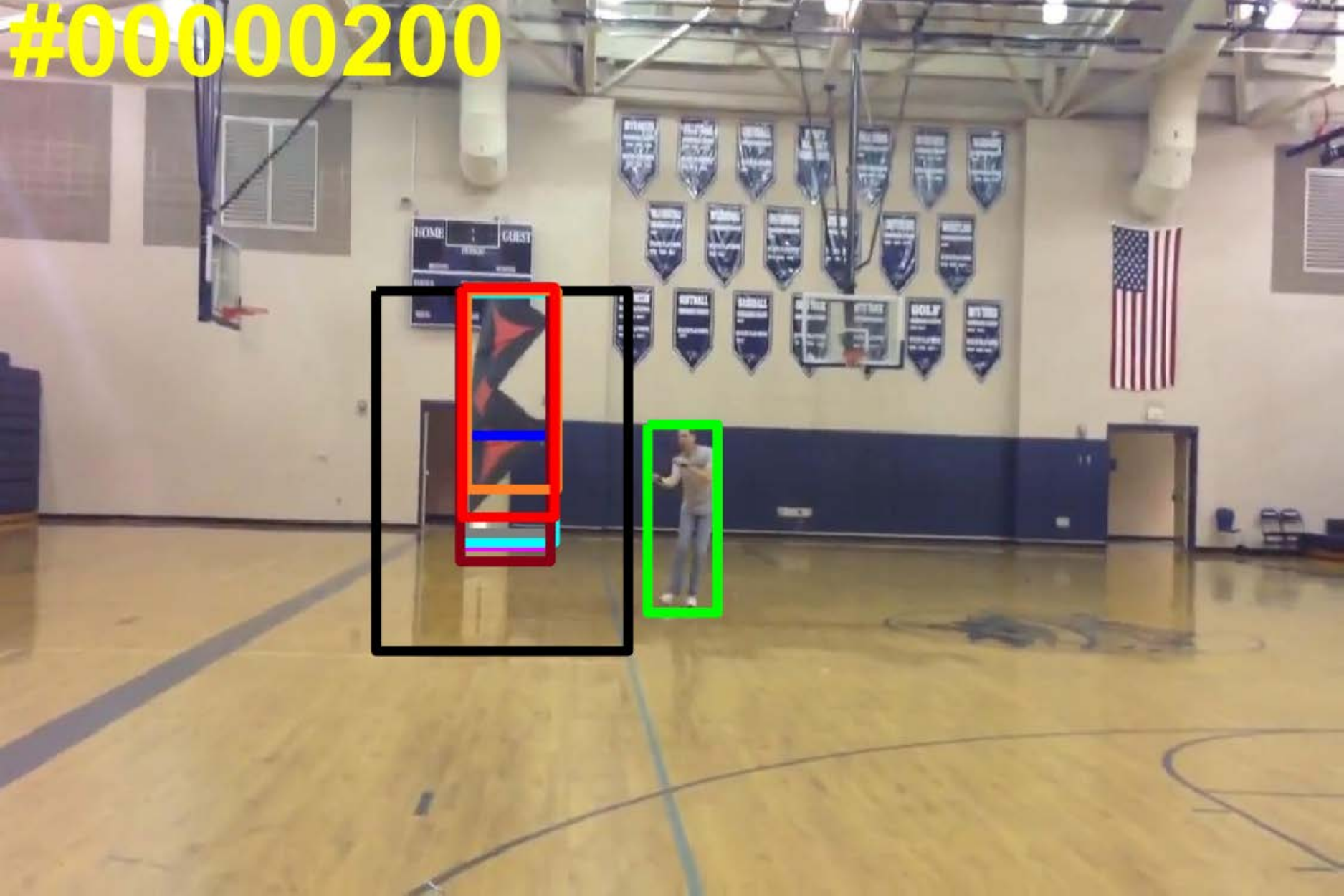}
            \includegraphics[width=0.193\linewidth]{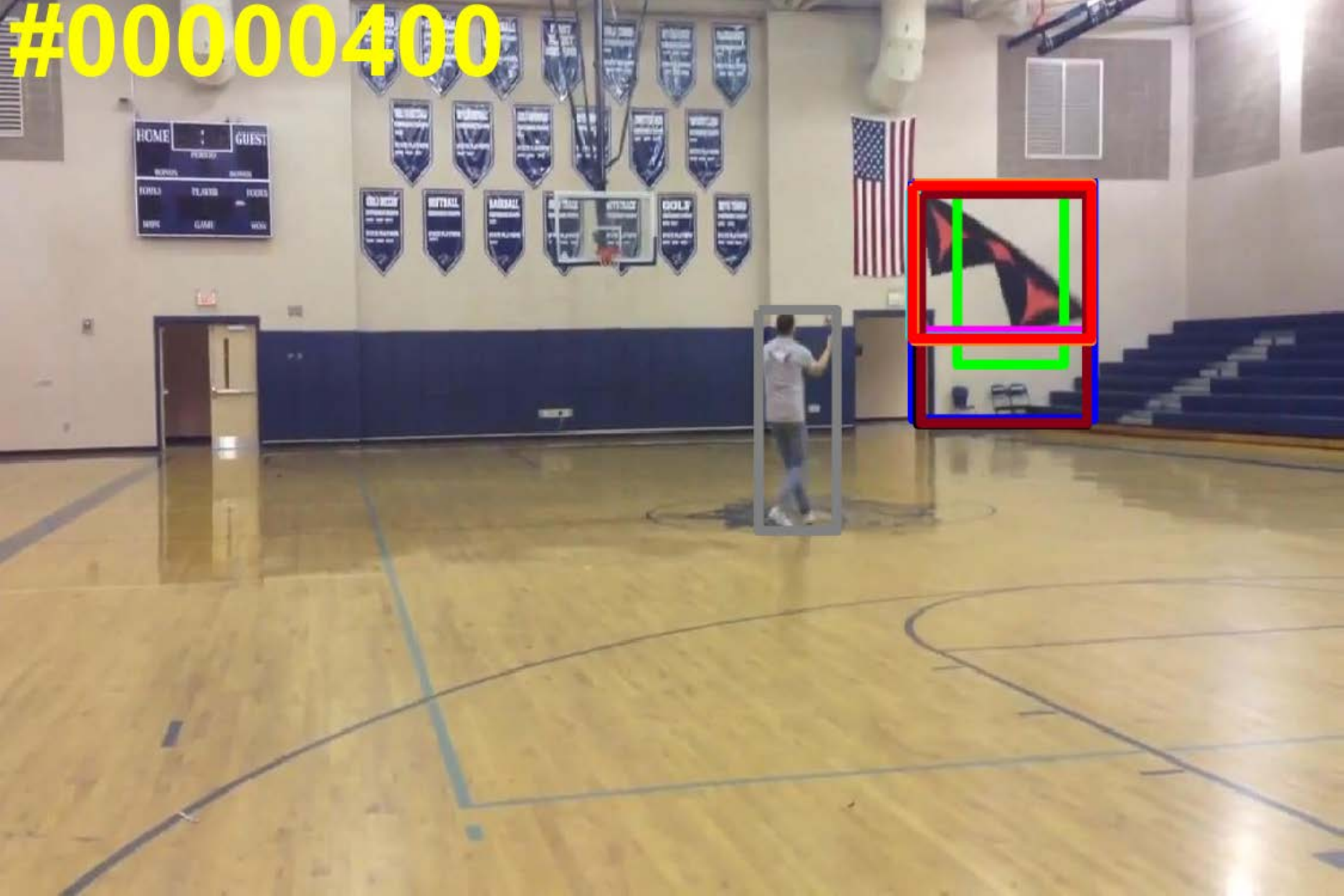}
            \includegraphics[width=0.193\linewidth]{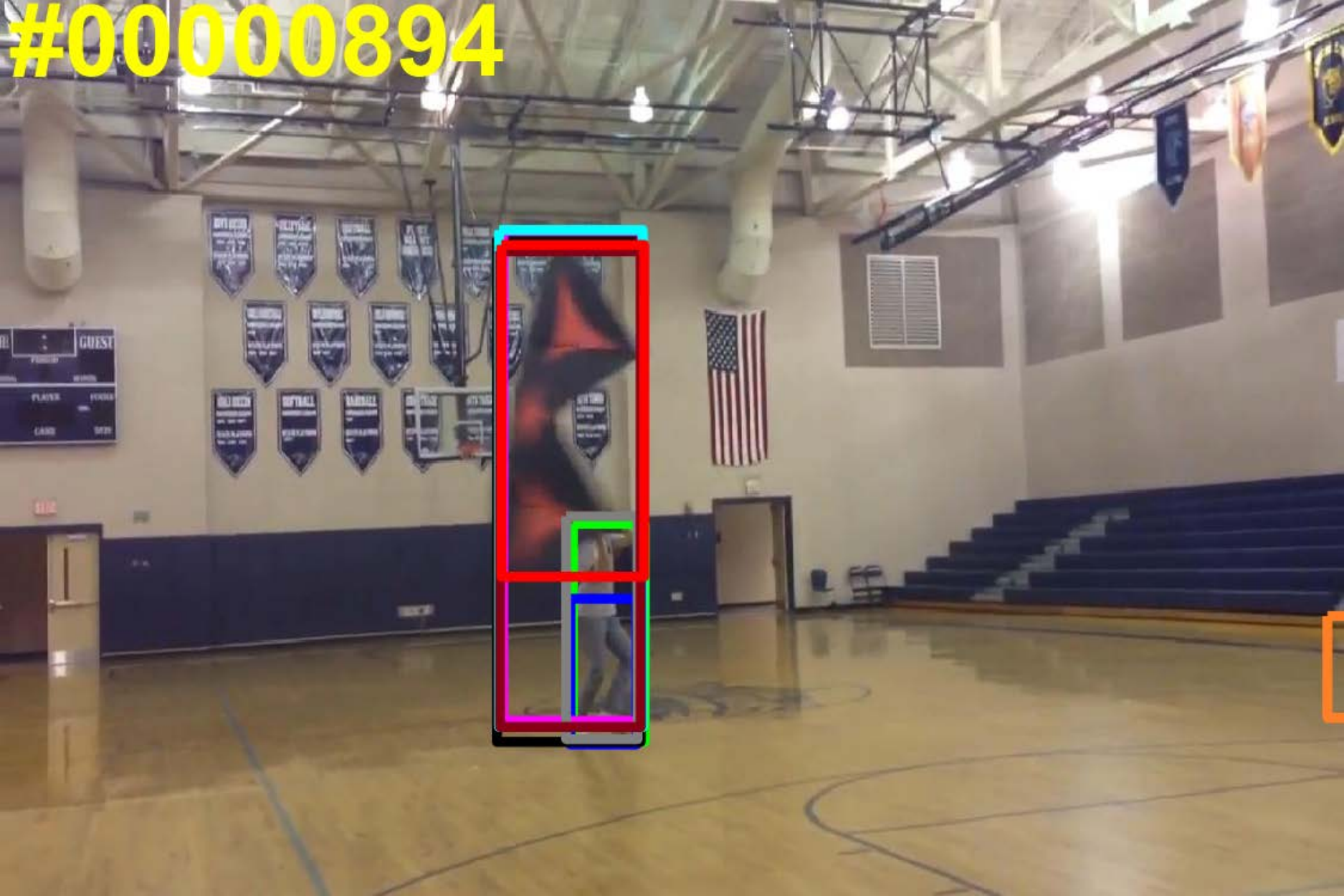}
            \includegraphics[width=0.193\linewidth]{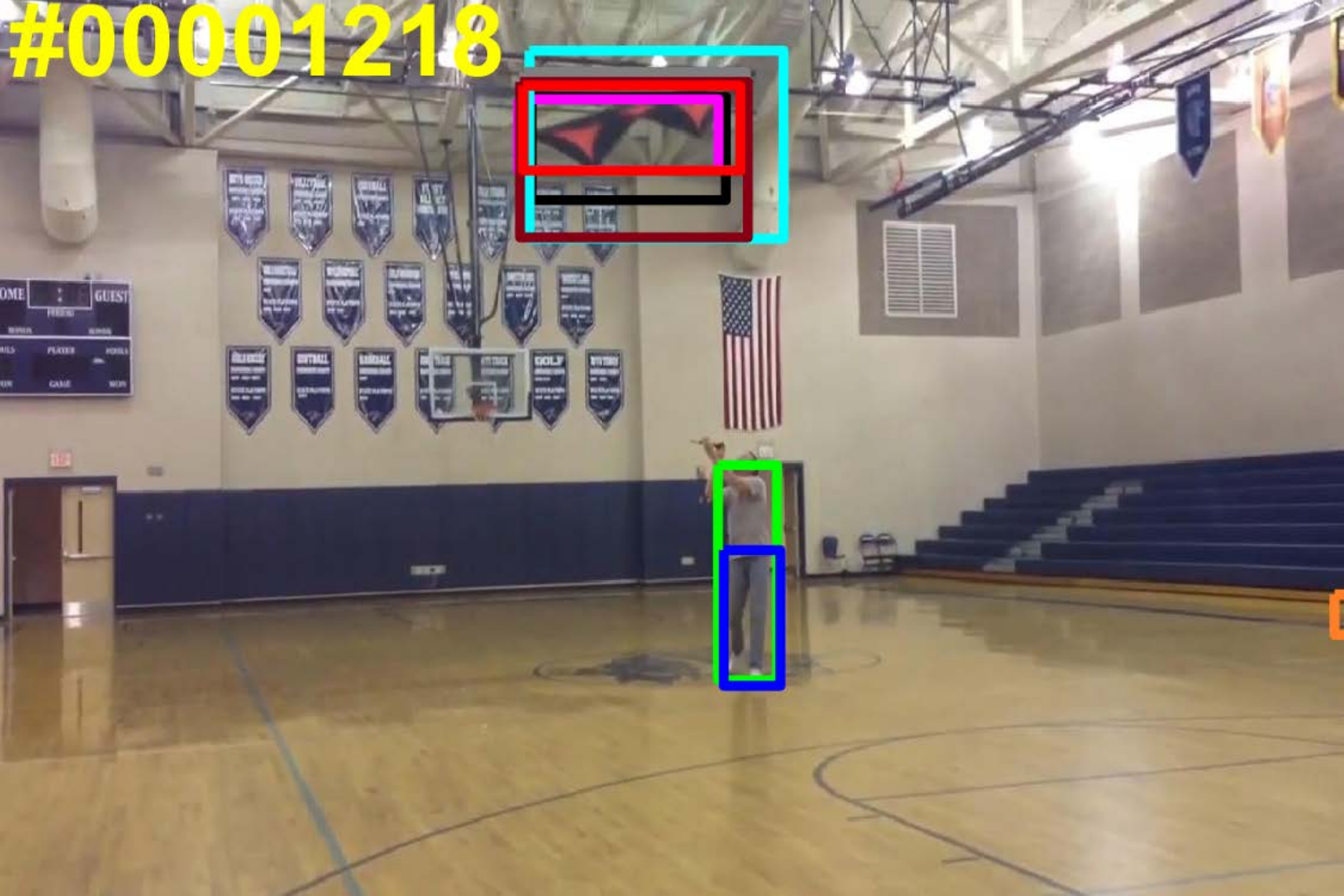}
		\end{minipage}
	    \begin{minipage}[c]{0.9\linewidth}
            \includegraphics[width=0.193\linewidth]{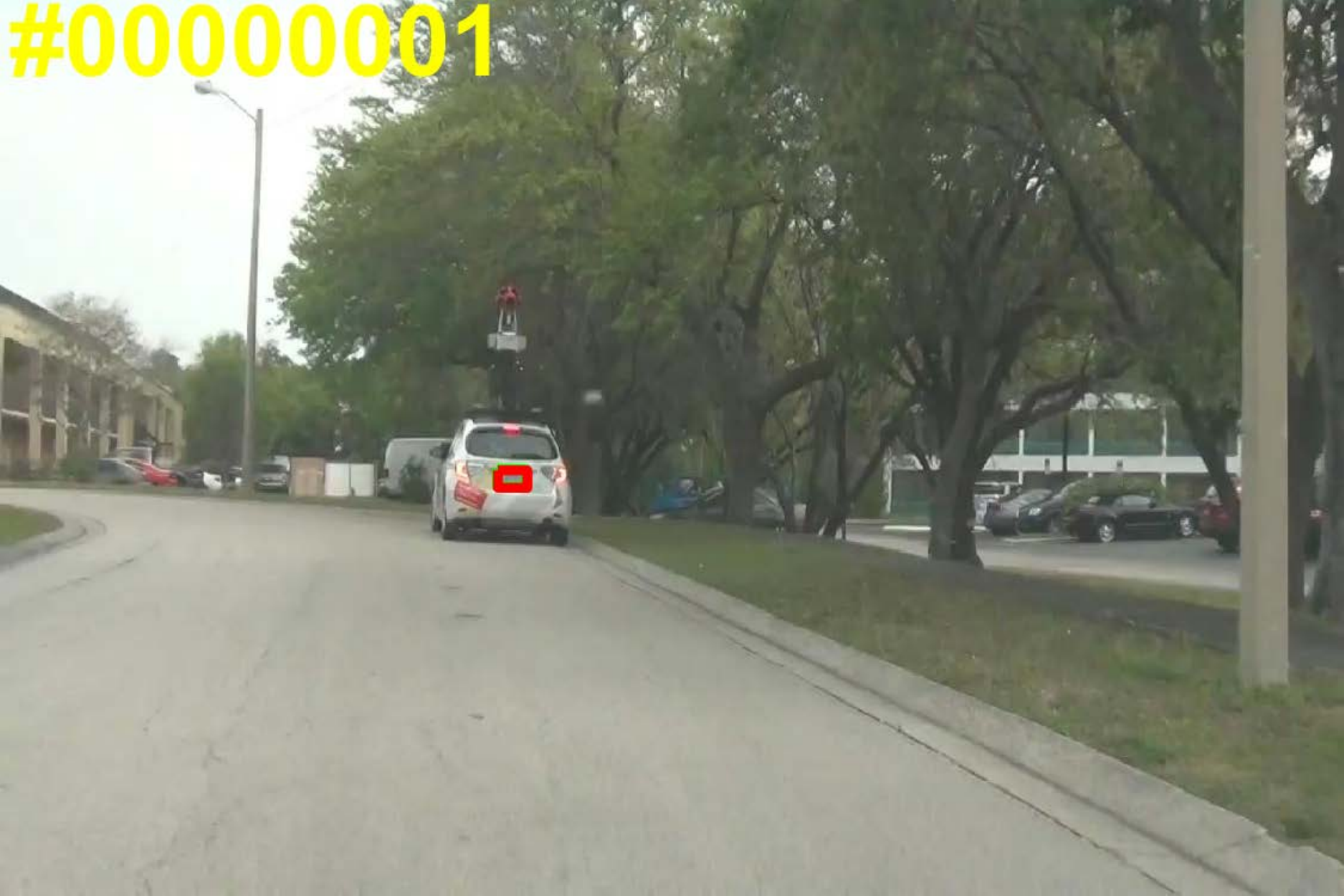}\vspace{1.5mm}
			\includegraphics[width=0.193\linewidth]{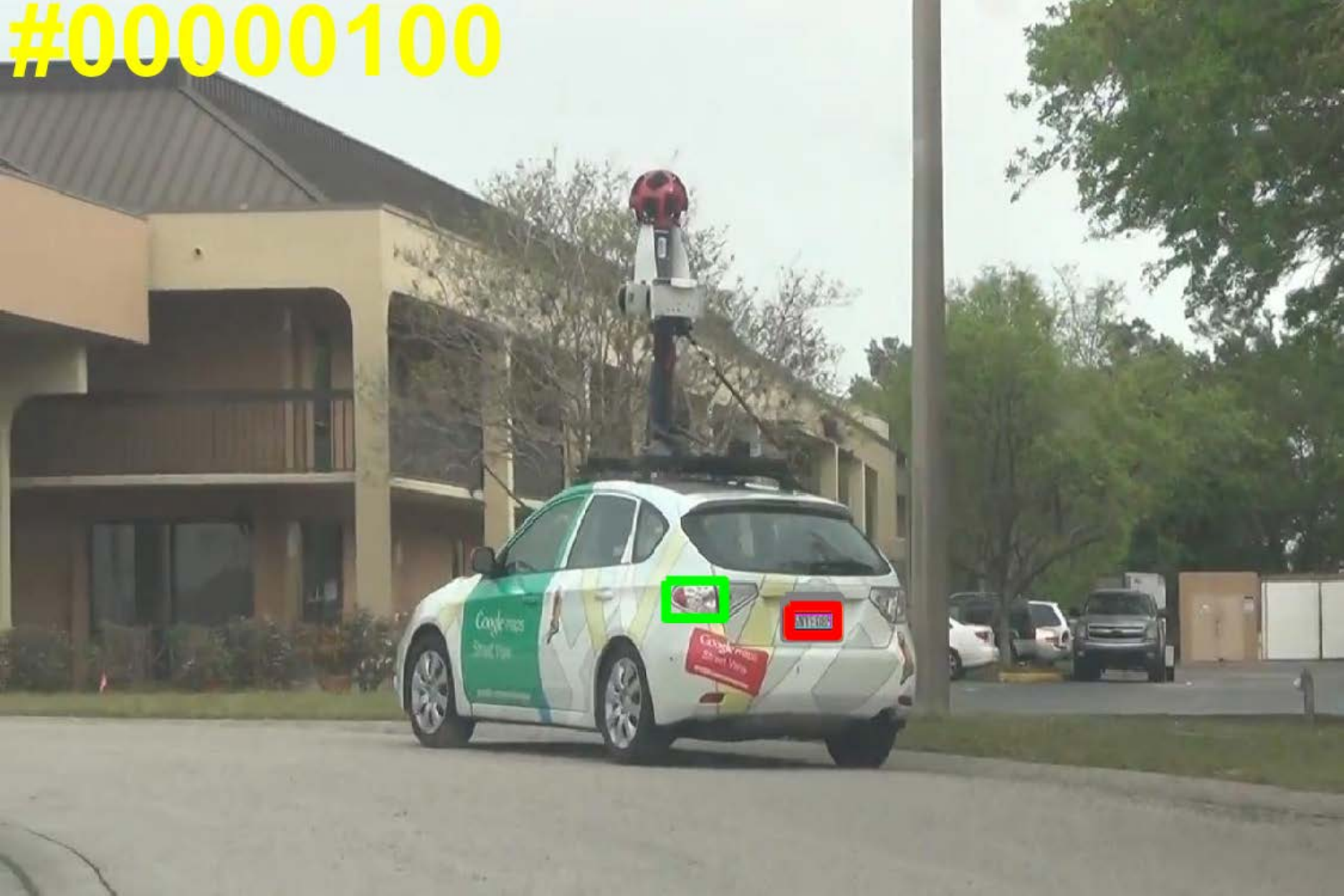}
            \includegraphics[width=0.193\linewidth]{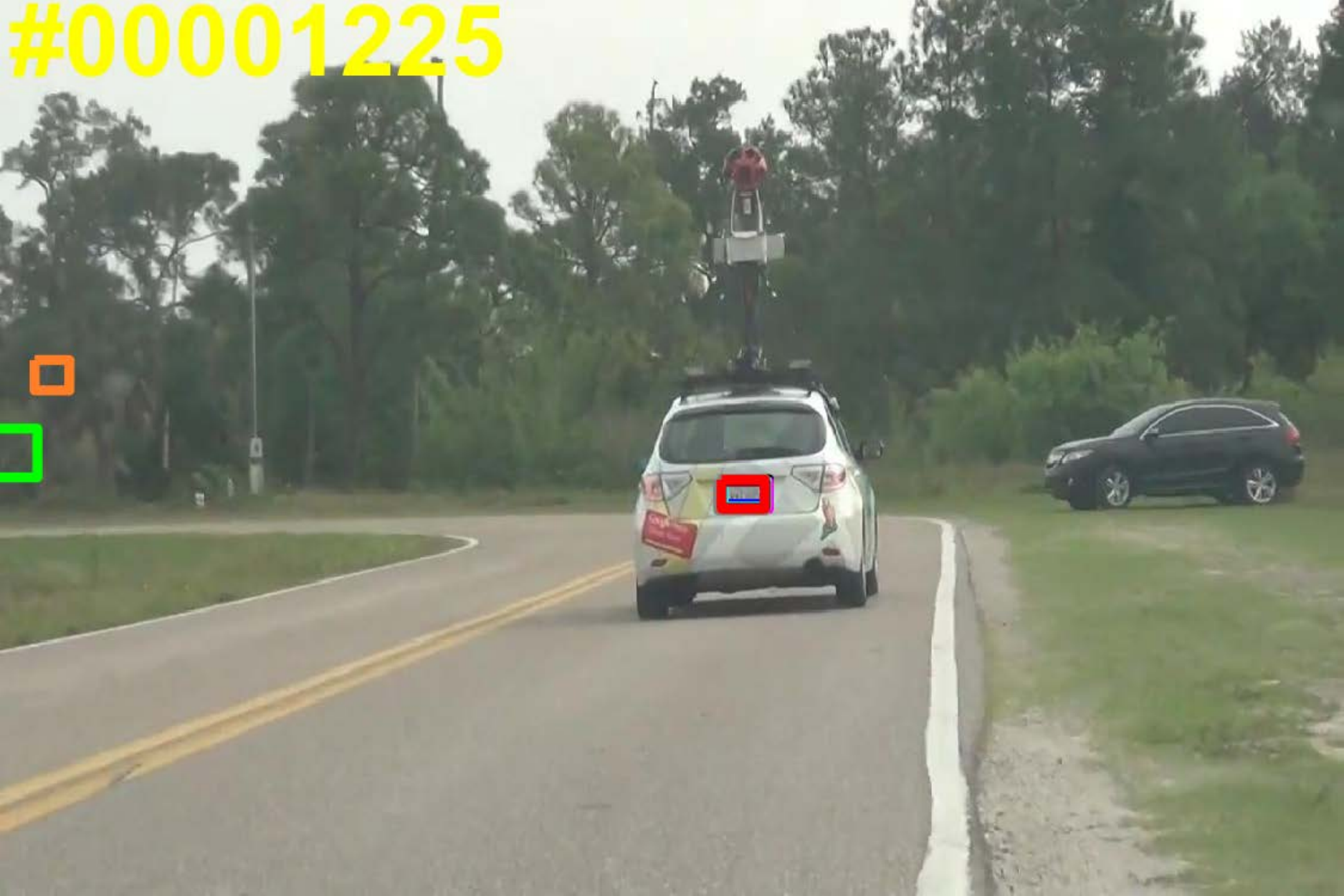}
            \includegraphics[width=0.193\linewidth]{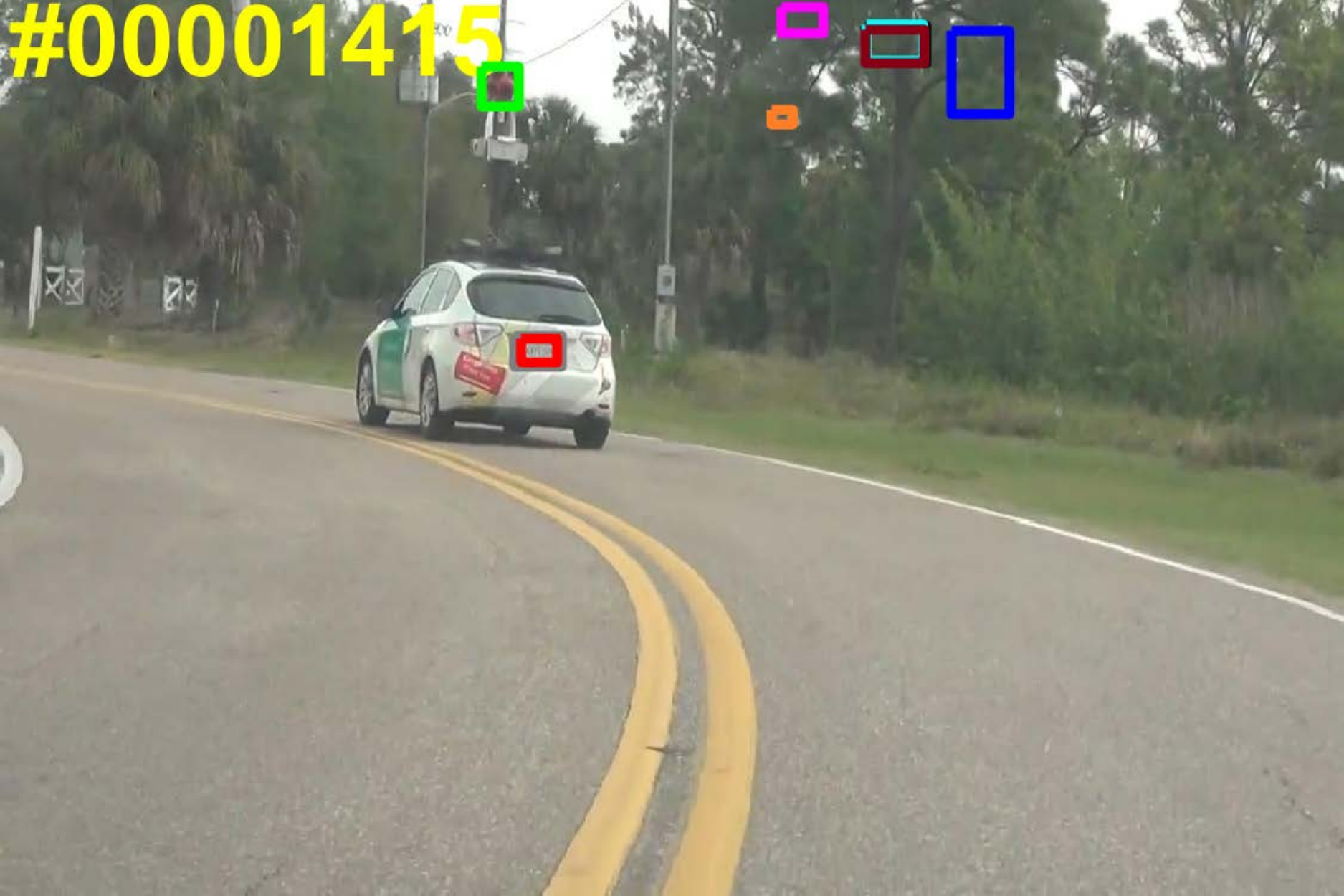}
            \includegraphics[width=0.193\linewidth]{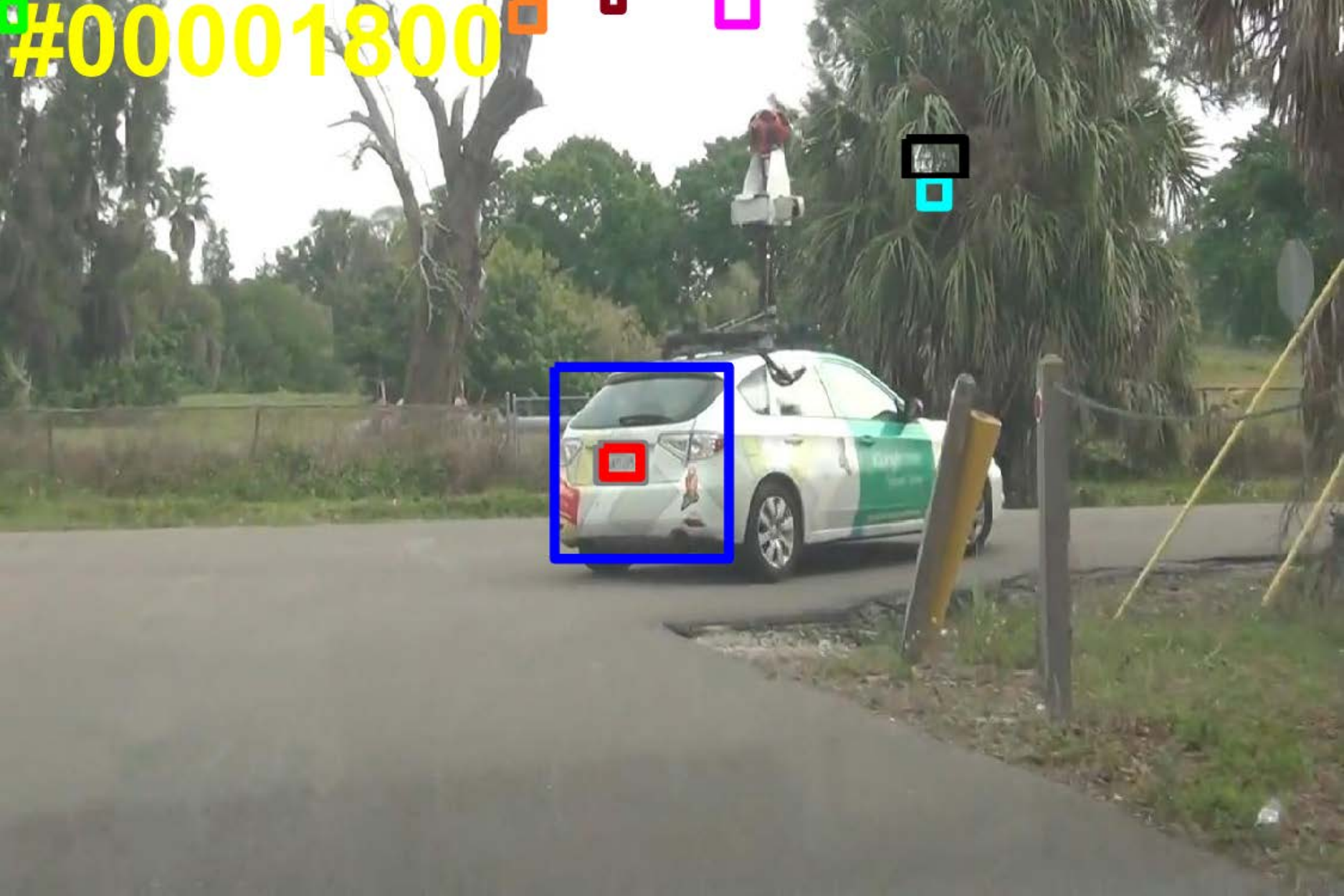}
		\end{minipage}

	    \begin{minipage}[c]{1\linewidth}
            \centering
            \includegraphics[width=0.9\linewidth]{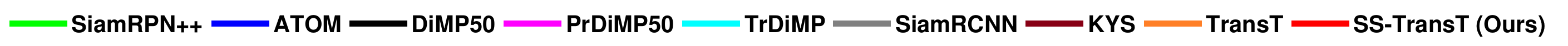}
		\end{minipage}
\end{center}
	\caption {\textbf{Visualization of tracking results on several challenging sequences.} The figure compares the results on seven challenging sequences (from top to bottom:\textit{bicycle-2}, \textit{crab-12} \textit{guitar-10}, \textit{umbrella-19}, \textit{hat-2}, \textit{kite-6}, and \textit{licenseplate-12} ) from the LaSOT dataset. It shows that our approach performs favorably against the state-of-the-art methods.}
	\label{fig:visual-results}
\end{figure*}

\vspace{2mm}
\noindent\textbf{Comparison against Unsupervised Methods.}
Figure~\ref{fig:overall-comparison} shows the performance of these methods.
Our approach achieves significant gains on all these datasets compared with the UDT, UDT-RPN++, and USOT methods.
We attribute the gains to the high-quality synthesized data generated by the proposed location-aware self-supervised learning mechanism with the specially customized transformations for effective training.
In contrast, the UDT and UDT-RPN++ algorithms exploiting the consistency constraint of forward and backward trajectories for training, cannot be applied to some challenging tracking scenarios where the consistency constraint does not hold.
In addition, replacing the backbone model of UDT as the SiamRPN++ model leads to performance drops on all these datasets.
This is because the consistency constraint of UDT pays less attention to the scale variation of the target during training, which makes the training of the bounding box regressor of the SiamRPN++ model less effective.
Compared with the supervised learning scheme, the proposed method achieves close performance on all these datasets, which shows the effectiveness under the unsupervised learning settings.

\subsection{Comparison against SOTA Methods}

In this part, we demonstrate the effect of the proposed approach in boosting existing tracking methods.
We integrate the proposed self-supervised data generation scheme into three representative tracking frameworks including the SiamRPN++~\cite{SiamRPN++}, DiMP~\cite{DiMP}, and TransT~\cite{TransT} models.
The corresponding integrated models are denoted as SS+SiamRPN++, SS+DiMP, and SS+TransT, respectively.
For a fair comparison, we train these models with the default training and testing settings reported by the original papers and just use the self-supervised data generation scheme to generate more training data.
In the following, we show the performance using both quantitative and qualitative evaluations.

\vspace{2mm}
\noindent\textbf{Quantitative evaluation.}
Table~\ref{tab:integration} shows the tracking results of the models with (SS+Su) and without (Su) using the synthesized training data on the OTB100, UAV123, NFS, TC128, and LaSOT datasets.
It shows that the proposed approach improves all the three kinds of tracking frameworks significantly on all the datasets.
For the OTB100, NFS, and TC128 datasets with various kinds of target variations, the proposed algorithm achieves average AUC score gains (over all these frameworks) of 1.1, 1.9, and 1.6, respectively.
The improvements on solving the diverse target variations can be attributed to the proposed Crop-Transform-Paste operation which accommodates different transformations to address various challenges in visual tracking.
In addition, our approach achieves average performance gains of 1.9 and 1.6 AUC scores on the UAV123, and LaSOT datasets.
The significant improvements on the test videos captured from the drone perspective and the long-term test videos demonstrate that the proposed Crop-Transform-Paste scheme generalizes well to different tracking scenarios.
This is because numerous kinds of variations, such as shifts, scale changes, rotation, occlusion, \etc, are similar under different scenarios and can be effectively simulated in the synthesized data using the proposed algorithm.

In addition to the above overall results, Figure~\ref{fig:plots-attri} shows the success plots over each tracking challenges on the LaSOT dataset of the methods including SiamRPN++~\cite{SiamRPN++}, ATOM~\cite{ATOM}, KYS~\cite{KYS}, DiMP50~\cite{DiMP}, PrDiMP50~\cite{PrDiMP}, SiamRCNN~\cite{SiamR-CNN}, TransT~\cite{TransT} and SS-TransT (trained using the proposed scheme).
Compared with TransT (baseline method), the SS-TransT model improves the performance by 1.7\%, 1.0\%,2.2\%, 1.0\%, 1.9\%, 1.8\%, 3.0\%, 2.3\%, 0.2\%. 1.3\%, 1.5\%, 1.6\%, 1.5\%, and 2.7\% over the attributes of aspect ratio change, background clutter, camera motion, deformation, fast motion, full occlusion, illumination variation, low resolution, motion blur, out-of-view, partial occlusion, rotation, scale variation, and viewpoint change, respectively.
These improvements on all these attributes demonstrate the effectiveness of the proposed Crop-Transform-Paste approach in providing high-diverse training data, which benefits the handling of different kinds of tracking challenging.
With the proposed approach, the SS-TransT model achieves favorable performance on all these attributes against the state-of-the-art methods.
It is worth noting that our approach improves the state-of-the-art performance on the large-scale dataset by absolute gains of 2.0\% and 1.4\% in terms of the normalized precision score and the AUC score, respectively.

\vspace{2mm}
\noindent\textbf{Qualitative analysis.}
Figure~\ref{fig:visual-results} presents the visualized results of the nine methods on  challenging sequences from the LaSOT dataset.
These sequences contain target objects with different categories and complex combinations of challenging attributes.
On the bicycle-2 and crab-12 sequences (row 1 and 2) with occlusion and deformation attributes, most trackers are influenced by the background objects that cover the target object and generate inaccurate box predictions.
In contrast, the SS-TransT tracker trained with the proposed scheme predicts the target boxes accurately.
On the guitar-10 and umbrella-19 sequences (row 3 and 4) with dramatic deformation and distractor, all trackers except the one trained in the proposed scheme are prone to drifting to background distractors.
On the hat-2, kite-6, and licenseplate-12 sequences (row 5, 6, and 7) involving the fast motion and out-of-plane rotation attributes, several trackers drift to the background due to fast motion, while others generate inaccurate target boxes influenced by the rotation attribute.
The proposed learning scheme helps the SS-TransT tracker handle all the above tracking challenges well and achieve accurate tracking performance.
We attribute this to the proposed self-supervised learning scheme, which generates synthesized training samples using challenge-oriented transformations.

\section{Conclusions}
In this paper, we address the issue of learning deep-tracking models under limited annotations.
Motivated by the success of self-supervised learning, we propose a Crop-Transform-Paste mechanism to perform self-supervised learning for training deep trackers with only one annotation per video.
To ensure the effectiveness and efficiency of the proposed scheme, we study the effects of different data transformations for visual tracking and designed challenge-oriented transformations for better dealing with tracking challenges.
Extensive evaluations on seven public datasets: OTB-2015, UAV123, NFS, TC128, as well as large-scale TrackingNet, LaSOT, and Got-10K demonstrate the effectiveness of the proposed self-supervised learning scheme in learning trackers under limited annotations, dealing with tracking challenges, and boosting the performance of the state-of-the-art methods.

\bibliographystyle{IEEEtran}
\bibliography{Tracking}

\begin{thebibliography}{10}
\providecommand{\url}[1]{#1}
\csname url@samestyle\endcsname
\providecommand{\newblock}{\relax}
\providecommand{\bibinfo}[2]{#2}
\providecommand{\BIBentrySTDinterwordspacing}{\spaceskip=0pt\relax}
\providecommand{\BIBentryALTinterwordstretchfactor}{4}
\providecommand{\BIBentryALTinterwordspacing}{\spaceskip=\fontdimen2\font plus
\BIBentryALTinterwordstretchfactor\fontdimen3\font minus
  \fontdimen4\font\relax}
\providecommand{\BIBforeignlanguage}[2]{{%
\expandafter\ifx\csname l@#1\endcsname\relax
\typeout{** WARNING: IEEEtran.bst: No hyphenation pattern has been}%
\typeout{** loaded for the language `#1'. Using the pattern for}%
\typeout{** the default language instead.}%
\else
\language=\csname l@#1\endcsname
\fi
#2}}
\providecommand{\BIBdecl}{\relax}
\BIBdecl

\bibitem{SIAMESEFC}
L.~Bertinetto, J.~Valmadre, J.~F. Henriques, A.~Vedaldi, and P.~H. Torr,
  ``Fully-convolutional siamese networks for object tracking,'' in
  \emph{European Conference on Computer Vision Workshops}, 2016.

\bibitem{LSTMSiameseT}
F.~Zhao, T.~Zhang, Y.~Wu, M.~Tang, and J.~Wang, ``Antidecay lstm for siamese
  tracking with adversarial learning,'' \emph{IEEE Transactions on Neural
  Networks and Learning Systems}, vol.~32, no.~10, pp. 4475--4489, 2021.

\bibitem{LARACF}
R.~Liu, Q.~Chen, Y.~Yao, X.~Fan, and Z.~Luo, ``Location-aware and
  regularization-adaptive correlation filters for robust visual tracking,''
  \emph{IEEE Transactions on Neural Networks and Learning Systems}, vol.~32,
  no.~6, pp. 2430--2442, 2021.

\bibitem{CCRRDT}
S.~Ge, C.~Zhang, S.~Li, D.~Zeng, and D.~Tao, ``Cascaded correlation refinement
  for robust deep tracking,'' \emph{IEEE Transactions on Neural Networks and
  Learning Systems}, vol.~32, no.~3, pp. 1276--1288, 2021.

\bibitem{SiamRPN++}
B.~Li, W.~Wu, Q.~Wang, F.~Zhang, J.~Xing, and J.~Yan, ``Siamrpn++: Evolution of
  siamese visual tracking with very deep networks,'' in \emph{IEEE Conference
  on Computer Vision and Pattern Recognition}, 2018.

\bibitem{SSSDO}
X.~Zhan, X.~Pan, B.~Dai, Z.~Liu, D.~Lin, and C.~C. Loy, ``Self-supervised scene
  de-occlusion,'' in \emph{IEEE Conference on Computer Vision and Pattern
  Recognition}, 2020.

\bibitem{sslpir}
I.~Misra and L.~v.~d. Maaten, ``Self-supervised learning of pretext-invariant
  representations,'' in \emph{IEEE Conference on Computer Vision and Pattern
  Recognition}, 2020.

\bibitem{SSDE}
F.~Tan, H.~Zhu, Z.~Cui, S.~Zhu, M.~Pollefeys, and P.~Tan, ``Self-supervised
  human depth estimation from monocular videos,'' in \emph{IEEE Conference on
  Computer Vision and Pattern Recognition}, 2020.

\bibitem{SimCLR}
T.~Chen, S.~Kornblith, M.~Norouzi, and G.~Hinton, ``A simple framework for
  contrastive learning of visual representations,'' in \emph{International
  Conference on Machine Learning}, 2020.

\bibitem{moco}
K.~He, H.~Fan, Y.~Wu, S.~Xie, and R.~Girshick, ``Momentum contrast for
  unsupervised visual representation learning,'' in \emph{IEEE Conference on
  Computer Vision and Pattern Recognition}, 2020.

\bibitem{DiMP}
G.~Bhat, M.~Danelljan, L.~Van~Gool, and R.~Timofte, ``Learning discriminative
  model prediction for tracking,'' in \emph{IEEE International Conference on
  Computer Vision}, 2019.

\bibitem{TransT}
X.~Chen, B.~Yan, J.~Zhu, D.~Wang, X.~Yang, and H.~Lu, ``Transformer tracking,''
  in \emph{IEEE Conference on Computer Vision and Pattern Recognition}, 2021.

\bibitem{SiamRPN}
B.~Li, J.~Yan, W.~Wu, Z.~Zhu, and X.~Hu, ``High performance visual tracking
  with siamese region proposal network,'' in \emph{IEEE Conference on Computer
  Vision and Pattern Recognition}, 2018.

\bibitem{DaSiamRPN}
Z.~Zhu, Q.~Wang, B.~Li, W.~Wu, J.~Yan, and W.~Hu, ``Distractor-aware siamese
  networks for visual object tracking,'' in \emph{European Conference on
  Computer Vision}, 2018.

\bibitem{TADT}
X.~Li, C.~Ma, B.~Wu, Z.~He, and M.-H. Yang, ``Target-aware deep tracking,'' in
  \emph{IEEE Conference on Computer Vision and Pattern Recognition}, 2019.

\bibitem{STMT}
Z.~Zhou, X.~Li, T.~Zhang, H.~Wang, and Z.~He, ``Object tracking via
  spatial-temporal memory network,'' \emph{IEEE Transactions on Circuits and
  Systems for Video Technology}, 2021.

\bibitem{MTTIT}
Q.~Liu, X.~Li, Z.~He, N.~Fan, D.~Yuan, W.~Liu, and Y.~Liang, ``Multi-task
  driven feature models for thermal infrared tracking,'' in \emph{Proceedings
  of the AAAI Conference on Artificial Intelligence}, 2020.

\bibitem{TrackingNet}
M.~Muller, A.~Bibi, S.~Giancola, S.~Alsubaihi, and B.~Ghanem, ``Trackingnet: A
  large-scale dataset and benchmark for object tracking in the wild,'' in
  \emph{European Conference on Computer Vision}, 2018.

\bibitem{GOT10k}
L.~Huang, X.~Zhao, and K.~Huang, ``Got-10k: A large high-diversity benchmark
  for generic object tracking in the wild,'' \emph{IEEE Transactions on Pattern
  Analysis and Machine Intelligence}, 2019.

\bibitem{UDT}
N.~Wang, W.~Zhou, Y.~Song, C.~Ma, W.~Liu, and H.~Li, ``Unsupervised deep
  representation learning for real-time tracking,'' \emph{International Journal
  of Computer Vision}, vol. 129, no.~2, pp. 400--418, 2021.

\bibitem{selfSDCT}
D.~Yuan, X.~Chang, P.-Y. Huang, Q.~Liu, and Z.~He, ``Self-supervised deep
  correlation tracking,'' \emph{IEEE Transactions on Image Processing}, 2020.

\bibitem{USOT}
J.~Zheng, C.~Ma, H.~Peng, and X.~Yang, ``Learning to track objects from
  unlabeled videos,'' in \emph{IEEE International Conference on Computer
  Vision}, 2021.

\bibitem{BYOL}
J.-B. Grill, F.~Strub, F.~Altch\'{e}, C.~Tallec, P.~Richemond, E.~Buchatskaya,
  C.~Doersch, B.~Avila~Pires, Z.~Guo, M.~Gheshlaghi~Azar, B.~Piot,
  k.~kavukcuoglu, R.~Munos, and M.~Valko, ``Bootstrap your own latent - a new
  approach to self-supervised learning,'' in \emph{Annual Conference on Neural
  Information Processing Systems}, 2020.

\bibitem{RandomCrop}
A.~Krizhevsky, I.~Sutskever, and G.~E. Hinton, ``Imagenet classification with
  deep convolutional neural networks,'' in \emph{Annual Conference on Neural
  Information Processing Systems}, 2012.

\bibitem{ColorJ}
C.~Szegedy, W.~Liu, Y.~Jia, P.~Sermanet, S.~Reed, D.~Anguelov, D.~Erhan,
  V.~Vanhoucke, and A.~Rabinovich, ``Going deeper with convolutions,'' in
  \emph{IEEE Conference on Computer Vision and Pattern Recognition}, 2015.

\bibitem{yolov1}
J.~Redmon, S.~Divvala, R.~Girshick, and A.~Farhadi, ``You only look once:
  Unified, real-time object detection,'' in \emph{IEEE Conference on Computer
  Vision and Pattern Recognition}, 2016.

\bibitem{CopyPasteNew}
G.~Ghiasi, Y.~Cui, A.~Srinivas, R.~Qian, T.-Y. Lin, E.~D. Cubuk, Q.~V. Le, and
  B.~Zoph, ``Simple copy-paste is a strong data augmentation method for
  instance segmentation,'' in \emph{IEEE Conference on Computer Vision and
  Pattern Recognition}, 2021.

\bibitem{mixup}
H.~Zhang, M.~Cisse, Y.~N. Dauphin, and D.~Lopez-Paz, ``mixup: Beyond empirical
  risk minimization,'' in \emph{International Conference on Learning
  Representations}, 2018.

\bibitem{CutMix}
S.~Yun, D.~Han, S.~J. Oh, S.~Chun, J.~Choe, and Y.~Yoo, ``Cutmix:
  Regularization strategy to train strong classifiers with localizable
  features,'' in \emph{IEEE International Conference on Computer Vision}, 2019.

\bibitem{CopyPaste}
N.~Dvornik, J.~Mairal, and C.~Schmid, ``Modeling visual context is key to
  augmenting object detection datasets,'' in \emph{European Conference on
  Computer Vision}.\hskip 1em plus 0.5em minus 0.4em\relax Springer, 2018.

\bibitem{SiamBAN}
Z.~Chen, B.~Zhong, G.~Li, S.~Zhang, and R.~Ji, ``Siamese box adaptive network
  for visual tracking,'' in \emph{IEEE Conference on Computer Vision and
  Pattern Recognition}, 2020.

\bibitem{VID}
O.~Russakovsky, J.~Deng, H.~Su, J.~Krause, S.~Satheesh, S.~Ma, Z.~Huang,
  A.~Karpathy, A.~Khosla, M.~Bernstein \emph{et~al.}, ``Imagenet large scale
  visual recognition challenge,'' \emph{International Journal of Computer
  Vision}, 2015.

\bibitem{COCO}
T.-Y. Lin, M.~Maire, S.~Belongie, J.~Hays, P.~Perona, D.~Ramanan,
  P.~Doll{\'a}r, and C.~L. Zitnick, ``Microsoft coco: Common objects in
  context,'' in \emph{European Conference on Computer Vision}, 2014.

\bibitem{youtubeBB}
E.~Real, J.~Shlens, S.~Mazzocchi, X.~Pan, and V.~Vanhoucke,
  ``Youtube-boundingboxes: A large high-precision human-annotated data set for
  object detection in video,'' in \emph{IEEE Conference on Computer Vision and
  Pattern Recognition}, 2017.

\bibitem{OTB2015}
Y.~Wu, J.~Lim, and M.-H. Yang, ``Object tracking benchmark,'' \emph{IEEE
  Transactions on Pattern Analysis and Machine Intelligence}, vol.~37, no.~9,
  pp. 1834--1848, 2015.

\bibitem{NFS}
H.~Kiani~Galoogahi, A.~Fagg, C.~Huang, D.~Ramanan, and S.~Lucey, ``Need for
  speed: A benchmark for higher frame rate object tracking,'' in \emph{IEEE
  International Conference on Computer Vision}, 2017.

\bibitem{UAV123}
M.~Mueller, N.~Smith, and B.~Ghanem, ``A benchmark and simulator for uav
  tracking,'' in \emph{European Conference on Computer Vision}, 2016.

\bibitem{OTB2013}
Y.~Wu, J.~Lim, and M.-H. Yang, ``Online object tracking: A benchmark,'' in
  \emph{IEEE Conference on Computer Vision and Pattern Recognition}, 2013.

\bibitem{Mosaic}
A.~Bochkovskiy, C.-Y. Wang, and H.-Y.~M. Liao, ``Yolov4: Optimal speed and
  accuracy of object detection,'' \emph{arXiv preprint arXiv:2004.10934}, 2020.

\bibitem{TC128}
P.~Liang, E.~Blasch, and H.~Ling, ``Encoding color information for visual
  tracking: Algorithms and benchmark,'' \emph{IEEE Transactions on Image
  Processing}, vol.~24, no.~12, pp. 5630--5644, 2015.

\bibitem{LaSOT}
H.~Fan, L.~Lin, F.~Yang, P.~Chu, G.~Deng, S.~Yu, H.~Bai, Y.~Xu, C.~Liao, and
  H.~Ling, ``Lasot: A high-quality benchmark for large-scale single object
  tracking,'' in \emph{IEEE Conference on Computer Vision and Pattern
  Recognition}, 2019.

\bibitem{ATOM}
M.~Danelljan, G.~Bhat, F.~S. Khan, and M.~Felsberg, ``Atom: Accurate tracking
  by overlap maximization,'' in \emph{IEEE Conference on Computer Vision and
  Pattern Recognition}, 2019.

\bibitem{KYS}
G.~Bhat, M.~Danelljan, L.~Van~Gool, and R.~Timofte, ``Know your surroundings:
  Exploiting scene information for object tracking,'' in \emph{European
  Conference on Computer Vision}, 2020.

\bibitem{PrDiMP}
M.~Danelljan, L.~V. Gool, and R.~Timofte, ``Probabilistic regression for visual
  tracking,'' in \emph{IEEE Conference on Computer Vision and Pattern
  Recognition}, 2020.

\bibitem{SiamR-CNN}
P.~Voigtlaender, J.~Luiten, P.~H. Torr, and B.~Leibe, ``Siam r-cnn: Visual
  tracking by re-detection,'' in \emph{IEEE Conference on Computer Vision and
  Pattern Recognition}, 2020.

\end{thebibliography}
\vspace{-0.25in}
\end{document}